\documentclass[journal,comsoc]{IEEEtran}
\usepackage[T1]{fontenc}
\ifCLASSINFOpdf
\else
\fi
\usepackage{amsmath}
\usepackage[cmintegrals]{newtxmath}
\usepackage{graphicx}
\usepackage{caption}
\usepackage{rotating}
\usepackage{xcolor, soul}
\usepackage{lineno,hyperref} 
\usepackage{amsfonts} 
\usepackage{caption}  
\usepackage{multirow}
\usepackage[margin=1cm]{geometry}
\usepackage{array,longtable}
\usepackage{cite}
\usepackage{comment}
\usepackage{threeparttable}
\usepackage{color} 
\begin{document}
\title{Resource Allocation and Service Provisioning in Multi-Agent Cloud Robotics: A Comprehensive Survey}                                                                                     
\author{Mahbuba~Afrin,~\IEEEmembership{Student~Member,~IEEE,} 
 		Jiong~Jin,~\IEEEmembership{Member,~IEEE,} 
		Akhlaqur~Rahman,~\IEEEmembership{Member,~IEEE,}       
        Ashfaqur~Rahman,~\IEEEmembership{Senior~Member,~IEEE,}
        Jiafu~Wan,~\IEEEmembership{Member,~IEEE,}
        and~Ekram~Hossain,~\IEEEmembership{Fellow,~IEEE}
\thanks{Mahbuba Afrin is with the School of Software and Electrical Engineering, Swinburne University of Technology, Melbourne, VIC 3122, Australia and Data61, CSIRO, Sandy Bay, TAS 7005, Australia (e-mail: mafrin@swin.edu.au, mahbuba.afrin@data61.csiro.au).}
\thanks{Jiong Jin is with the School of Software and Electrical Engineering, Swinburne University of Technology, Melbourne, VIC 3122, Australia (e-mail: jiongjin@swin.edu.au).}
\thanks{Akhlaqur Rahman is with the School of Industrial Automation and Electrical Engineering, Engineering Institute of Technology (EIT), Melbourne, VIC 3000, Australia (e-mail: akhlaqur.rahman@eit.edu.au).}
\thanks{Ashfaqur Rahman is with Data61, CSIRO, Sandy Bay, TAS 7005, Australia (e-mail: ashfaqur.rahman@data61.csiro.au).}\thanks{Jiafu Wan is with the School of Mechanical and Automotive Engineering, South China University of Technology, Guangzhou 510640, China (e-mail: mejwan@scut.edu.cn).}\thanks{Ekram Hossain is with the Department of Electrical and Computer Engineering, University of Manitoba, Winnipeg, R3T 5V6, Canada (e-mail: Ekram.Hossain@umanitoba.ca).}}
%
\markboth{}%
{Afrin \MakeLowercase{\textit{et al.}}: Resource Allocation and Service Provisioning in Multi-Agent Cloud Robotics: A Comprehensive Survey}
\maketitle
\begin{abstract}
Robotic applications nowadays are widely adopted to enhance operational automation and performance of real-world Cyber-Physical Systems (CPSs) including Industry 4.0, agriculture, healthcare, and disaster management. These applications are composed of latency-sensitive, data-heavy, and compute-intensive tasks. The robots, however, are constrained in the computational power and storage capacity. The concept of multi-agent cloud robotics enables robot-to-robot cooperation and creates a complementary environment for the robots in executing large-scale applications with the capability to utilize the edge and cloud resources. However, in such a collaborative environment, the optimal resource allocation for robotic tasks is challenging to achieve. Heterogeneous energy consumption rates and application of execution costs associated with the robots and computing instances make it even more complex. In addition, the data transmission delay between local robots, edge nodes, and cloud data centres adversely affects the real-time interactions and impedes service performance guarantee. Taking all these issues into account, this paper comprehensively surveys the state-of-the-art on resource allocation and service provisioning in multi-agent cloud robotics. The paper presents the application domains of multi-agent cloud robotics through explicit comparison with the contemporary computing paradigms and identifies the specific research challenges. A complete taxonomy on resource allocation is presented for the first time, together with the discussion of resource pooling, computation offloading, and task scheduling for efficient service provisioning. Furthermore, we highlight the research gaps from the learned lessons, and present future directions deemed beneficial to further advance this emerging field.
\end{abstract}
%
%
\IEEEpeerreviewmaketitle
\section{Introduction}\label{intro}
\IEEEPARstart{I}{}n recent years, a significant emphasis is given on building smart Cyber-Physical Systems (CPSs) in industry, transport, healthcare and agriculture to perform complex engineering operations with limited human involvement, reduced cost and improved performance \cite{rajkumar2010cyber}. The basic component of these CPSs are the robots. A robot is an autonomous entity that can perceive the external environment, make intelligent decisions, and trigger physical actions. Although a robot is equipped with computing resources to conduct small-scale data processing, it cannot carry out large-scale computations on its own \cite{survey_cloudRObotics_architecture}. To deal with this limitation of standalone robots, the concept of multi-robot system has been emerged. In multi-robot systems, a group of robots (connected via a wired or wireless network) work collaboratively to achieve a common goal \cite{Survey_curretSatus_CloudRobotics}. For example, in a smart factory, some robots handle the inventory while others check the safety and quality; and collectively they contribute to the production management. Nevertheless, the performance of multi-robot system is subject to the heterogeneous energy capacity of robots. In case of widespread deployment, the maintenance of persistent communication among the robots becomes difficult due to their mobility. The lack of storage within the robots causes further disruptions in exchanging and preserving the large volume of data during robot-to-robot interactions \cite{survey_cloudRObotics_architecture}. To overcome these limitations, the concept of cloud computing has been extended to multi-robot system, which is termed as cloud robotics \cite{survey_cloudRObotics_architecture,Survey_curretSatus_CloudRobotics}. In this paradigm, cloud offers the computing resources such as virtual machines or containers and engages resources from both local robots and remote data centres for scalable and extensive data processing \cite{rahman2019energy}.
\begin{figure*}[!t]
\centering
\includegraphics[height=3in,width=6in]{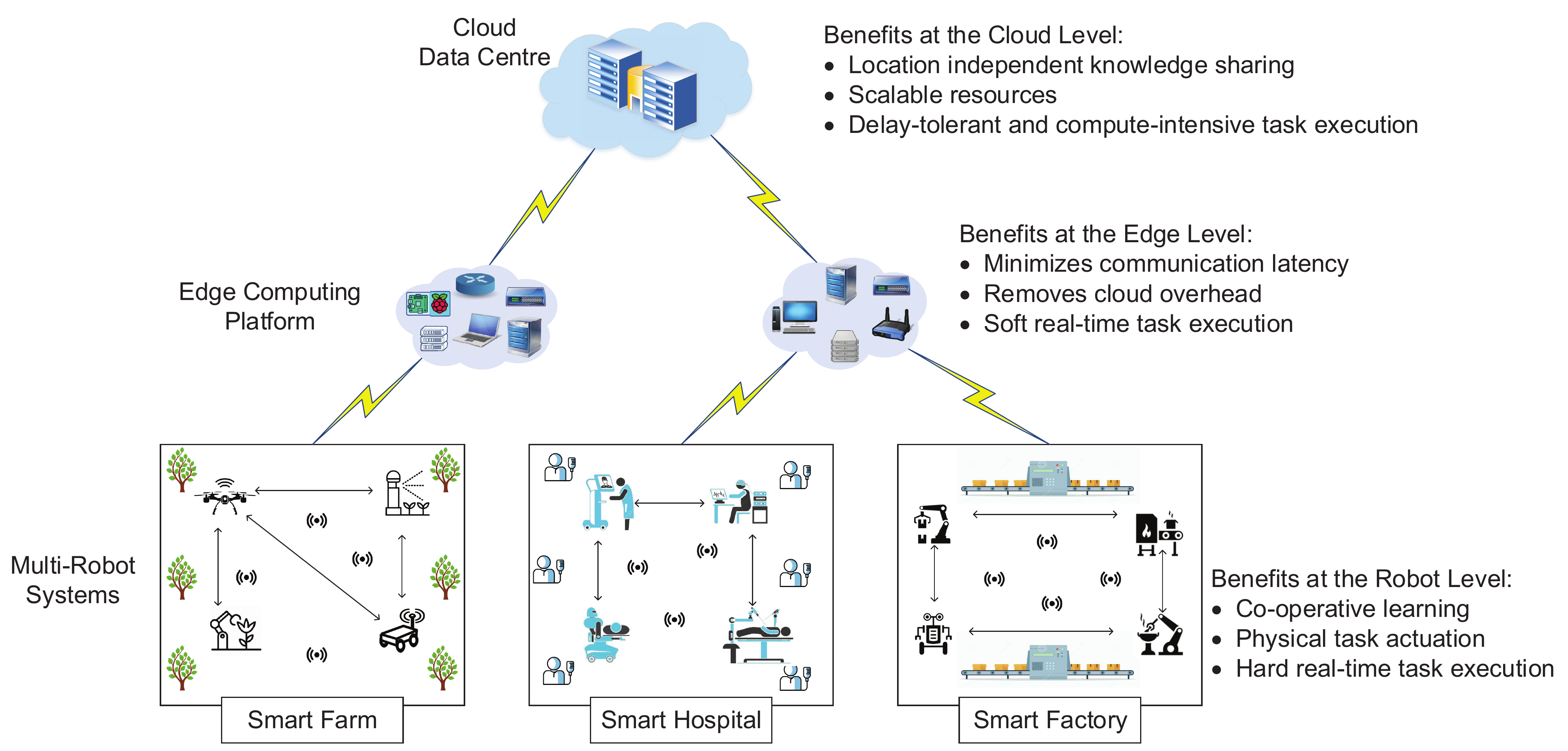} 
\caption{A multi-agent cloud robotic system and its benefits.}
\label{fig:multi-agent_cloudRobotics}
\end{figure*}
\par However, the robotic systems and the cloud data centres are normally multi-hop distance apart and that is responsible for longer communication time and data transfer delay. As a result, cloud robotics often becomes less suitable for latency-sensitive operations. Edge computing can play a vital role in addressing this limitation by bringing the infrastructure, platform, and software services closer to the data sources through cloudlet \cite{SPE_QCASH} and fog nodes \cite{mouradian2017comprehensive,mahmud2019quality,Mahmud_Acm_fog_2020}. The integration of edge computing \cite{edge_computing} with existing cloud robotics \cite{afrin2019multi} paves the way of creating a new computing paradigm for executing robotic applications and their associated tasks on the physical resources at different communication hops from robot-to-cloud. Since this new paradigm simultaneously harnesses the computing capabilities at device level (robot), edge level (fog node, cloudlet) and remote level (cloud data centres), in this paper, it is termed as multi-agent cloud robotics. Figure \ref{fig:multi-agent_cloudRobotics} depicts the organization and benefits of a multi-agent cloud robotic system. In this system, tasks with hard real-time constraints such as robotic surgery in smart hospital \cite{ma2015robot}, soft real-time tasks with less restrictive deadline such as pesticide spraying in smart farm \cite{musat2018advanced}, and parcel distribution in smart factory \cite{smartfactory,shaik2020enabling} can be performed. The edge computing platform performs the soft real-time tasks including field image processing and path planning. On the other hand, the latency-tolerant and high storage requiring tasks such as eHealth report distribution and weather forecasting are managed by the cloud data centre.

In the context of CPSs, multi-agent cloud robotics can bring various benefits as listed below.
\begin{itemize}
\item{By exploiting computing resources at different infrastructure levels, it provides options to the CPSs for executing diverse robotic applications.}
\item{It offers an abstraction to classify the tasks of robotic applications as per the characteristics of computing infrastructures.}
\item{It provides distributed resources to process big data generated in the CPSs with reduced communication delay.}
\item{It enhances the competency of robots in sharing knowledge using local and edge network rather than depending on the distant cloud.}
\item{It significantly reduces the bandwidth requirements (for sending data to cloud) in CPSs dealing with multiple devices, by bringing the resources closer to the robots.}
\item{With multi-agent cloud robotics, the dependency of the CPSs to the cloud data centre as well as the load on cloud data centre (to handle multiple CPSs) is sharply decreased.}  
\end{itemize} 
\par Based on the characteristics, a multi-agent cloud robotic system appears similar to the contemporary computing paradigms, namely, Mobile Cloud Computing (MCC) \cite{shiraz2012review} and Multi-access Edge Computing (MEC) \cite{taleb2017multi}, where the edge computation is extensively harnessed. However, there exist explicit differences among the functionalities of these paradigms. They are discussed as follows.
\subsection{Multi-Agent Cloud Robotics and Related Computing Paradigms}
MCC facilitates resource and energy-constrained smart phones by enabling them to offload compute-intensive mobile applications to cloud data centres for execution. Through an intermediate layer of cloudlets between the smart phones and cloud data centres, MCC manages a 3-tier execution platform for the offloaded applications. MCC supports both local and global distribution of computing and application services through private resources. These services are accessed using LAN or WiFi technology \cite{othman2013survey,dinh2013survey}. On the other hand, MEC manages a virtual server on the cellular base stations to assist faster networking, location tracking and content delivery services for the smart phone and Internet of Things (IoT) users \cite{taleb2017multi,porambage2018survey}. It usually offers a multi-tier computing platform for the applications, where the services are reachable via one-hop proximal distant access points using LTE, 3GPP or 5G network technology \cite{markakis2017efficient}. Generally, the MEC infrastructures are provided by telecom operators. Unlike MCC and MEC, a multi-agent cloud robotic system follows an $n$-tier resource orientation, where the services are offered by harnessing local (robots), proximal (edge resources), or global (cloud data centres) computing agents. Here, the primary data producers are the sensor clusters, IoT devices and robots and the services are accessed by wireless access technologies including WiFi, LTE, Zigbee, Bluetooth, LoRa technology including 3GPP and 5G \cite{survey_cloudRObotics_architecture,sarker2019offloading,beigi_smartcity_cloudRobotics}.\par Moreover, compared to MCC and MEC, multi-agent cloud robotics explicitly supports artificial intelligence so that robots can make autonomous decision. Being an intelligent entity, robots are able to learn from the environment and alter their predefined mobility pattern dynamically. Conversely, MCC and MEC do not support autonomous mobility as the end devices generally operate through human intervention. These distinctive architectural and operational features make a multi-agent cloud robotic system more complicated than MCC and MEC.
\par Since the multi-agent cloud robotics is a complex synthesis of robotics, tele-operation, edge computing and core cloud technologies, the performance of such an environment depends on proper management and co-ordination among heterogeneous resources. Therefore, efficient resource allocation and service provisioning policies are required to fully exploit the benefits of multi-agent cloud robotics. Generally speaking, resource allocation refers to the assignment of requested applications to the competent resources, whereas service provisioning sets the base for resource allocation through resource pooling and computation offloading, and assists in task scheduling. The development of efficient resource allocation and service provisioning policies often becomes challenging due to the diverse requirements unique to multi-agent cloud robotics.
\begin{table*}[t]
\centering 
\caption{Comparison with related computing paradigms of multi-agent cloud robotics}\label{Tab:summary_rel_concept} 
\scriptsize
\begin{tabular}{|p{3.6cm}|p{3.7cm}|p{4.5cm}|p{4.9cm}|}\hline 
\textbf{Dimension} & \textbf{Mobile Cloud Computing} & \textbf{Multi-Access Edge Computing} & \textbf{Multi-Agent Cloud Robotics} \\\hline 
Geo-distribution& Local/ Global&	Local/ Proximal&	Local/ Proximal/ Global\\\hline
Resource orientation& 3-tier& 2/3-tier & $n$-tier \\\hline 
Infrastructure provider& Private entities & Telecom operators& Private entities/ Cloud provider\\\hline
Application type& Lightweight & Lightweight & Lightweight/ Heavyweight \\\hline
Data producer & Mobile devices & IoT devices/ Cellular gateways& Multiple sensors/ IoT devices/ Robots \\\hline
General data type &	Batch &	Simple stream &	Complex stream \\\hline
Service access & LAN/ WiFi & LTE/ 3GPP/ 5G &	WiFi/ LTE/ Zigbee/ Bluetooth/ LoRa/ 3GPP/ 5G \\\hline
Autonomous learning support&No&	No&	Yes\\\hline
Autonomous mobility&No&	No&	Yes\\\hline
\end{tabular}  
\end{table*} 
\subsection{Challenges of Resource Allocation and Service Provisioning in Multi-Agent Cloud Robotics}
The domain specific requirements and challenges for resource allocation and service provisioning in multi-agent cloud robotics are listed below:
\begin{itemize}
\item{\textit{Real-time learning and autonomous action:} Robotic applications require advanced artificial intelligence and machine learning techniques for situation-aware real-time and automated action. Although edge infrastructure in multi-agent cloud robotics helps meeting real-time service requirements, it lacks support for installing numerous resource-enriched computing devices with large amount of processing cores and GPUs due to cost, structural and maintenance constraints. Consequently, they limit the scope of executing compute-intensive artificial intelligence and machine learning techniques. Therefore, the resource allocation and service provisioning in multi-agent cloud robotics urges rigorous classification and selection of compatible resources for robotic applications which is not mandatory for MCC and MEC dealing with predefined set of events and actions.}
\item{\textit{Complex data stream processing:} Multi-agent cloud robotics requires modular robotic applications with diverse programming models including map-reduce and distributed data flow to process complex data streams generated by robot-embedded multi-purpose sensors \cite{birk2009networking}. Due to resource and communication constraints, the local and proximal resources of multi-agent cloud robotics often fail to participate in such intense data processing that increases the burden on global resources. Therefore, resource allocation and service provisioning policies for a multi-agent cloud robotics system need to determine where to process data streams in real time. This significantly differs from MCC and MEC, which generally process simple batch or stream data generated by conventional sensors of smart phones and IoT devices.}
\item{\textit{Dynamic robotic cooperation:} There are some cases such as earthquake disaster management and submarine cable maintenance when multiple heterogeneous robots need to work collaboratively for attaining a common goal. Moreover, while operating in such adverse working environments with limited communication support, the real-time robot-to-robot and robot-to-edge or cloud interactions get hampered significantly \cite{beigi_smartcity_cloudRobotics}. To meet these requirements and constraints, the robotic cooperation often depends on distributed but coordinated algorithms such as federated learning and distributed learning, which adds further resource management overhead to multi-agent cloud robotics. The resource and service provisioning policies should address this issue deliberately that is not compulsory in MCC and MEC due to their simplified working environments.}
\item{\textit{Cross-infrastructure interoperability:} In multi-agent cloud robotics, heterogeneous and distributed computing resources complement the robots to carry out their responsibilities. These resources are owned by different service providers. For example, multi-agent cloud robotic systems can use the Google or Amazon cloud infrastructure and their edge infrastructure can be set by the cellular service providers. The resource sharing, privacy, and fault tolerance functions of these infrastructure are also managed with separate black box software systems. Therefore, the resource allocation and service provisioning policies in multi-agent cloud robotics require support for cross-infrastructure and policy-driven interoperability, which is not essential for MCC and MEC because of their association with homogeneous resource providers.}
\item{\textit{Synchronized decision making:} Since multi-agent cloud robotics involves computing infrastructure from different communication layers to execute robotic applications, an explicit synchronization of decision making entities belonging to these infrastructure is highly required. However, the attainment of such synchronization becomes challenging due to the intelligence interference among robots, edge, and cloud resources. Therefore, resource allocation and service provisioning policies should fairly distribute the resource management responsibilities among multiple decision making entities, which is not obligatory for MCC and MEC relying on a single decision maker.}
\item{\textit{On-demand computation and communication trade-off:} The offloading facility in multi-agent cloud robotics enables the robots having low energy or low processing power to access the edge and cloud infrastructure and execute the compute-intensive robotic applications. To conduct this activity, a seamless interaction among local robots, edge and remote resources are required which consumes additional communication time. It becomes more complicated because of the uncertain mobility of robots. Furthermore, the heterogeneity of resources in multi-agent cloud robotics with respect to processing cores, networking standards and energy consumption rates intensifies the appropriate resource selection problem for offloading. Therefore, resource allocation and service provisioning policies in multi-agent cloud robotics require on-demand computation and communication trade-off \cite{AK_MotionandConnectivity,AK_SmartCity}. On the other hand, in MCC and MEC, the computation and communication trade-off often depend on predefined rules because of the static mobility pattern of the users.}
\item{\textit{System-specific policy:} In multi-agent cloud robotics, the Quality of Service (QoS) requirements of robotic systems vary from one to another. For example, the service delivery deadline for a robotic application in smart hospital is more stringent than that of in smart farm. Similarly, for some robotic systems, higher accuracy is important such as robotic surgery, whereas for others the minimization of cost is vital such as smart factory maintenance. A generalized resource allocation and service provisioning policy is not feasible for multi-agent cloud robotics like MCC and MEC. Therefore, system-specific policies are required for efficient resource management, which is quite difficult due to the dynamics of robotic systems.}
\item{\textit{Energy-delay optimization:} Although computing resources from every layer of the hierarchical architecture participate in executing robotic applications, still the realization of such an environment is constrained by the energy limitations of the resources. Moreover, it is challenging to finish the application execution with the residual energy of the resources. Specifically, in a multi-agent cloud robotic system, robots can perform simultaneously as a service provider and a user. As a result, the energy limitations of robots significantly hamper the execution of applications compared to MCC and MEC. Therefore, selection of appropriate resources and time-efficient resource allocation are required to deliver services within the energy constraints of the resources.}
\item{\textit{Comprehensive business model:} Cloud computing has a widely accepted business model for subscription-oriented services. However, such business model is not feasible for edge infrastructure as it mainly deals with event-driven and localized demand. On the other hand, robotic systems often require a set of reserved resources for processing their complex stream of data. Due to such variations of service requirements, it is complicated to develop a comprehensive business model for multi-agent cloud robotics that significantly disrupts the cost and budget-satisfying resource allocation, which is generally easier to develop in MCC and MEC.}
\item{\textit{Security and safety:} In a multi-agent cloud robotics system, robots are able to share the resources with other robots, edge resources as well as cloud data centre. Consequently, this system can easily become a target for security threats at every communication layer (local, edge, and cloud). The number of points of data exposure is comparatively higher in multi-agent cloud robotics than in MCC and MEC. Even in a local network, if any robot becomes vulnerable to privacy and security threats, it can easily manipulate others' data in the system because of robot itself being an intelligent entity. This type of risk is relatively lower in MCC and MEC. In addition, due to the random mobility of the robots, they are required to communicate over heterogeneous networks that gives rise to another security issue. Therefore, for efficient resource allocation and service provisioning, the selection of reliable resources is crucial. Apart from the security vulnerabilities, uncertain events such as link failures, edge node shout down, power cut and interference can obstruct the robots to access the infrastructure services. In such cases, proactive and reactive fault tolerance techniques are helpful to make the system robust; however, it is difficult to develop these techniques due to the dynamics of multi-agent cloud robotic systems.}
\end{itemize}
\par Resource allocation and service provisioning have been well studied in different computing paradigms including MCC and MEC \cite{shiraz2012review,othman2013survey,dinh2013survey,taleb2017multi,porambage2018survey}. However, the adopted algorithms, techniques and recommendations of these paradigms cannot directly be applied to multi-agent cloud robotics due to the above-mentioned issues.
\subsection{Motivation of this Article and Our Contribution}
A notable number of research works have been conducted to ensure efficient resource allocation and service provisioning in multi-agent cloud robotics overcoming the existing challenges. Depending on network condition, application requirements and resource availability, different robot-edge-cloud interaction techniques such as peer to peer, proxy and clone-based communication have been developed \cite{survey_cloudRObotics_architecture,Survey_curretSatus_CloudRobotics}. Although these works have significant impact in enhancing the resource allocation and service provisioning on multi-agent cloud robotics, there are a very few efforts in the literature for categorizing them in a systemic manner. To address this issue, we have conducted an extensive survey on existing resource allocation and service provisioning policies for multi-agent cloud robotics. To the best of our knowledge, this is the first literature review on resource allocation and service provisioning problem in multi-agent cloud robotics. This work can be a notable inclusion to the existing literature, helping readers understand the state-of-the-art of multi-agent cloud robotics with its promising use cases for Industry 4.0 applications, healthcare, smart agriculture, disaster management and other robotic applications. The importance of optimal resource allocation, resource pooling, computation offloading, and task scheduling are well studied in this survey for efficient service provisioning to the multi-agent robotic systems. A summary of lessons learned from the existing works in literature are also offered to identify the research gaps to address the challenges of resource allocation and service provisioning. Moreover, a holistic framework for efficient resource allocation and service provisioning is also provided and we believe that it could add a value to the existing literature and open directions for future research. The major contributions of this paper are: 
\begin{itemize}
\item A taxonomy of resource allocation considering the resource type, performance metrics, application structure, service model and allocation mechanism is presented in this survey.
\item Existing approaches on resource pooling, computation offloading and task scheduling in this field are also well explored for efficient service provisioning.
\item The lessons learned from the literature review are summarized and the gaps in efficient resource allocation and service provisioning for robotic systems in multi-agent cloud robotics are identified.
\item Moreover, future research directions with a holistic framework premised on them will assist researchers to improve the state of multi-agent cloud robotics.
\end{itemize}  
\begin{table}[!t]
\centering 
\caption{A list of abbreviations used in this survey}\label{Tab:abbreviations} 
\scriptsize
\begin{tabular}{|p{1.5 cm}|p{6.5cm}|}
\hline 
BoT & Bag-of-Tasks \\\hline
CAN & Controller Area Network \\\hline
CPSs & Cyber-Physical Systems\\\hline
CRASP & Cross-infrastructure Resource Allocator and Service Provisioner \\\hline
DAG & Directed Acyclic Graph \\\hline
DL & Deep Learning \\\hline
DSN & Data Stream Network \\\hline
IaaS & Infrastructure-as-a-Service (IaaS) \\\hline
ILP & Integer Linear Programming \\\hline
IoT & Internet of Things \\\hline
MCC & Mobile Cloud Computing\\\hline
MEC & Multi-access Edge Computing \\\hline
ML & Machine Learning \\\hline
NaaS & Networking-as-a-Service \\\hline 
NSGA-II & Non-dominated Sorting Genetic Algorithm-II \\\hline
PaaS & Platform-as-a-Service \\\hline
QoE & Quality of Experience \\\hline
QoS & Quality of Service \\\hline
RaaS & Robot-as-a-Service \\\hline
ROS & Robot Operating System \\\hline
SaaS & Software-as-a-Service \\\hline
SLA & Service Level Agreement \\\hline
SLAM & Simultaneous Localization and Mapping\\\hline
SOA & Service-oriented Architecture \\\hline
UAVs & Unmanned Aerial Vehicles\\\hline
uIP & Micro Internet Protocol \\\hline
VMs & Virtual Machines \\\hline
\end{tabular}  
\end{table}
\subsection{Article Organization}
The rest of the article is organized as follows. The list of abbreviations used in this survey is provided in Table \ref{Tab:abbreviations}. Before going into an in-depth discussion on resource allocation and service provisioning, in Section \ref{overview}, we summarize the recent practices of multi-agent cloud robotics both in academia and industry along with current and future applications of this paradigm. In this section, the contributions of our survey are also compared with related surveys. In Section \ref{resource_allocation}, a comprehensive review with the help of a taxonomy is presented particularly focusing on the research of resource allocation in multi-agent cloud robotics. Different techniques in resource pooling, computation offloading, and task scheduling are studied in Section \ref{service_provisioning}. From the lessons learned in this survey paper, the research gaps for efficient resource allocation and service provisioning are investigated in Section \ref{lessons}. A prospective holistic framework for efficient resource allocation and service provisioning along with directions for future research are provided in Section \ref{future}. Finally, concluding remarks are drawn in Section \ref{conclusion}.
\section{Overview of Multi-Agent Cloud Robotics Advancements} \label{overview}
In this section, the current and future applications and recent practices of multi-agent cloud robotics in industry are summarized. We also conduct a comparative study of our work with existing surveys.
\subsection{Current and Future Applications of Multi-Agent Cloud Robotics}
Multi-agent cloud robotics has significantly improved the performance of various CPSs including smart factory, remote healthcare, smart farm, and disaster management by offering real-time execution platform for different robotic applications including simultaneous localization and mapping (SLAM), robotic vision, path planning, navigation, grasping and surgical assistance \cite{Survey_curretSatus_CloudRobotics}.  
\subsubsection{Industry 4.0}
Industry 4.0 refers to the fourth industrial revolution that assists in digitization of industrial manufacturing with the help of IoT, big data analytics, and cloud computing. In this environment, a group of robots and human work collaboratively throughout the entire industrial value chains \cite{wee2015industry}\cite{industry4.0_survey}. As industrial robots are the key drivers for such applications, multi-agent cloud robotics inherently plays a significant role in developing these applications \cite{Aissam2019}. Smart factory is one of the prominent applications of Industry 4.0 \cite{wang2016ubiquitous,wang2016towards_smartfactory,smartfactory,wang2016implementing,wan2019reconfigurable}. 
\par Many works have concentrated on multi-agent cloud robotics for smart factory. The feasibility of multi-agent cloud robotics in gathering and processing sensory data for the navigation of autonomous vehicles in industrial environment is discussed in \cite{peake2015cloud ,cardarelli2015cloud,wan2019reconfigurable,cardarelli2017cooperative}. This concept is also exploited for ubiquitous product management \cite{wang2017cloud,wang2017integrated,wang2017ubiquitous,wang2018cloud}, customer maintenance \cite{Aissam2019,wang2016ubiquitous} and material handling \cite{wan_Context-aware-cloud-robotics} in smart factory.
\par In smart factory, cloud and edge based industrial robots have to deal with heterogeneous tasks and dynamic environment like taking variety of orders from consumers, dealing with different situations and making delivery of each order \cite{Aissam2019,wang2016ubiquitous}. The dynamic scheduling of the tasks to the components in smart factory according to workload is one of the key research challenges \cite{wan2018artificial}. For the task execution and resource sharing among cloud and edge-aided industrial robots, computational load scheduling is also equally important \cite{Smart_Manufacturing}. To meet these aspects, development of efficient resource allocation strategies in multi-agent cloud robotics is a fundamental demand for Industry 4.0 applications.
\subsubsection{Health Care}
The capabilities of multi-agent cloud robotics have also been leveraged to enable remote health monitoring \cite{ma2015robot} and elderly care support \cite{bonaccorsi2015design,bonaccorsi2016cloud,fiorini2017enabling}, and assist people with disabilities in mobility \cite{kamei2012cloud,kamei2017cloud} and communication \cite{ng2015cloud,manzi2017design}.
\par To monitor the health status of a patient in real time, a healthcare system based on robotics and cloud computing is designed in \cite{ma2015robot} which can also be controlled remotely by the doctors or carer. The authors of \cite{ng2015cloud} present a robotic telepresence platform by means of cloud technology, supporting mobility impaired peoples. Their platform assists the robot with autonomous navigation capabilities and teleoperation among robot and users. The developed solution in \cite{bonaccorsi2015design} bestows health-care management services to senior citizens and improves their living standards. It supports user with indoor localization service, care reminding service, environmental monitoring service in form of robotic service. Location based and personalized assistive services to seniors are also delivered by \cite{bonaccorsi2016cloud}. For chronic disease management, a domiciliary reminder service for personalised medical support using hybrid cloud robotics framework is afforded in \cite{fiorini2017enabling}. Another service is provisioned in \cite{manzi2017design} to support human-robot interaction and environmental sensing.
\par With the help of cloud and edge computing, service robots perform different services to overcome the limitations of healthcare \cite{zhang2015health,pham2018delivering,chung2019chatbot,wan2020cognitive}; however, proper allocation of resources is required to confer guaranteed QoS for 
time-critical task execution in multi-agent cloud robotic systems. 
\subsubsection{Agriculture}
Smart farming is a viable solution to develop the traditional agricultural system, where real-time data gathering, processing and analysis, as well as automation technologies on the farming procedures are applied \cite{musat2018advanced,bauer2018design}. Recently, the concept of multi-agent cloud robotics has been extended for smart farming with the help of smart, distributed and collaborated sensors, IoT, GPS, cloud computing, aerial and ground robots \cite{duckett2018agricultural,pivoto2018scientific} and offer autonomous field monitoring, pesticides spraying, weeding as well harvesting \cite{ kulbacki2018survey,valecce2019interplay,o2019edge,danton2020development}.
\par A service model using an agricultural expert cloud for ubiquitous agricultural environments is developed in \cite{cho2012agricultural}, which yields necessary information and services for cultivating any crops on any ecosystem. Based on cloud computing, another service-oriented smart farming architecture is introduced in \cite{apostol2015towards} that bestows local farm services, real-time services and cloud services including weather and map services to multiple farming areas. To collect data from deployed sensors in a smart farm including cameras, drones and soil sensors considering the bandwidth constraints, an IoT platform is presented in \cite{vasisht2017farmbeats}. Another integrated system to support collection, analysis, and prediction of agricultural environment for strawberry infection prediction is outlined in \cite{kim2018iot}.
\par Unmanned Aerial Vehicles (UAVs) play a significant role in smart farming system by providing imagery analysis and agricultural surveillance as well as on-demand communication \cite{nintanavongsa2017impact}. UAVs are applied in \cite{lottes2017uav} to perform vegetation detection, plant-tailored feature extraction and classification to estimate the distribution of crops and weeds in the field. Through a coalition of aerial and ground robots, how the services of multi-agent cloud robotics can be extended to smart farming is also demonstrated in \cite{kim2019unmanned}. However, agricultural production systems are prone to unpredictable environment (e.g., rain, temperature, humidity etc.) and unwanted events (e.g., animal diseases, pests) \cite{kamilaris2016agri}. Still the smart farms need to integrate robots and sensor data for delivering agro-services, which are not constrained by communication and unpredictable environments. The use of smart technologies including artificial intelligence and multi-agent cloud robotics are in their infancy stage in agriculture \cite{grieve2019challenges}, opening a way for further investigation.
%
\subsubsection{Disaster Management}
The services of multi-agent cloud robotics have been adopted widely for conducting unmanned search and rescue operations in hostile environments \cite{ermacora2013cloud,gregory2016application,jangid2016cloud,botta2017networking,Robots-as-a-Service-Search-and-Rescue} including alpine \cite{marconi2012sherpa} and fire-driven emergency scenarios \cite{afrin2019multi,afrin2018energy}.
\begin{table*}[!htb]
\centering 
\caption{Comparison with related survey papers}\label{Tab:summary_rel_survey} 
\scriptsize
\begin{tabular}{|p{3 cm}|p{2.6cm}|p{2.6cm}|p{2.6cm}|p{2 cm}|p{3.5 cm}|}
\hline 
\textbf{Works} & \textbf{Resource Allocation } & \textbf{Resource Discovery}  & \textbf{Computation Offloading}& \textbf{Task Scheduling}  & \textbf{Research Domain}  \\\hline
\cite{yousafzai2017cloud,manvi2014resource,toosi2014interconnected,luong2017resource, mann2015allocation} & \checkmark& $\wp$	  &  &	& Cloud computing\\\hline  
\cite{singh2016survey} & &\checkmark	  &  & $\wp$	& Cloud computing\\\hline
\cite{huang2013survey} &\checkmark &	  &  & $\wp$	& Cloud computing\\\hline
\cite{mastelic2015cloud,zhan2015cloud} &  &	  &  & \checkmark	& Cloud computing\\\hline
\cite{zhang2012auction} & \checkmark&	  &  &	& Wireless and mobile systems\\\hline
\cite{budzisz2014dynamic} & &	\checkmark  &  &	& Wireless access network\\\hline 
\cite{maallawi2014comprehensive} & &	 & \checkmark  &	& Wireless access and core network\\\hline 
\cite{rebecchi2014data} & &	 & \checkmark  &	& Cellular network\\\hline
\cite{shiraz2012review,othman2013survey,xu2013survey,abolfazli2013cloud,sanaei2013heterogeneity} & &$\wp$	 & \checkmark  &$\wp$	& Mobile cloud computing\\\hline  
\cite{mach2017mobile,mao2017survey,wang2017survey} & &	 & \checkmark  &	& Mobile edge computing\\\hline
\cite{taleb2017multi} & &	 & \checkmark  &	& Multi-access edge computing\\\hline
\cite{khamis2015multi,mosteo2010survey} & \checkmark&	  &  &	& Multi-robot system\\\hline 
\cite{survey_cloudRObotics_architecture,survey_CloudRoboticsandAutomation}& &   & $\wp$ 	 &    & Cloud robotics \\\hline 
\cite{Survey_curretSatus_CloudRobotics}& $\wp$ &   &  	 & $\wp$ & Cloud robotics \\\hline
\cite{saha_CloudRobotics_Survey}&$\wp$&   &  	$\wp$ &   & Cloud robotics \\\hline
\cite{Robotic_Cooperation}&  &   & $\wp$ 	 &    & Cloud robotics \\\hline
This survey & \checkmark & \checkmark	 & \checkmark  & \checkmark	& Multi-agent cloud robotics\\\hline
\end{tabular}  
\begin{tablenotes}
     \item[1] $\checkmark$ denotes broad discussion on the respective topic.
     \item[2] $\wp$ denotes partial discussion on the respective topic.
   \end{tablenotes}
\end{table*}
\par To enhance the collaboration of human rescuer and ground-aerial robots, in \cite{botta2017networking} cloud computing is augmented with robotics to provide fast, reliable and available resources for searching and rescuing operation. For the interaction between users and UAVs, another multi-agent cloud robotics architecture is presented in \cite{ermacora2013cloud} for emergency management and monitoring service. Moreover, for search and rescue operations in large-scale disaster, cloud resources are offered as infrastructure to robots in \cite{Robots-as-a-Service-Search-and-Rescue}. Multi-agent cloud robotics has also been enhanced for fire-driven emergency management service in \cite{afrin2019multi} and \cite{afrin2018energy} by incorporating edge resources.
\par In general, while providing disaster management services, a group of robots, edge and cloud resources work collaboratively towards completing a central mission. Thus, optimal allocation of tasks to the robots and offloading the tasks to the edge or cloud is required to achieve the objective successfully considering the environmental uncertainties.
\par Apart from the above mentioned applications, with the advancement of multi-agent cloud robotics, robots offer services in smart home system for making daily life more easier \cite{huiyong2013building,li2015ehopes,pan2016homecloud,do2018rish,yang2020ai}. Robotic services are also offered in security inspection, grocery shopping delivery, smart transportation, entertainment as well as education and many more applications \cite{survey_cloudRObotics_architecture}.
\subsection{Recent Industrial Practices of Multi-Agent Cloud Robotics}
Apart from the academia, several technology and business organizations are also focusing on the development of various solutions to integrate multi-robot systems, edge and cloud computing. For example, Google, Amazon, and Microsoft have already started extending cloud technologies to the robots. Google cloud robotics platform aims at harnessing artificial intelligence, cloud, and robotics to offer utility services to the customers. Amazon is patronizing the AWS RoboMaker that provides machine learning, monitoring and analytics support to the robots. Microsoft has developed Robot Operating System (ROS) to program the robots with higher capacity \cite{liu2019summary}. Additionally, the Honda RaaS platform aims at providing a wide range of robot and cloud-based services to support communication, robotic cooperation, and data sharing. Another company named CloudMinds works on production of robots with embedded deep learning facilities for real-time data collection and sharing using 5G. There exist other robotics companies including Fetch Robotics, InVia Robotics, Kuka, Plus One Robotics, FANUC Corporation, ABB Robotics that are currently working on warehouse automation \cite{Forbes,CloudMinds}.
\par Despite of such initiatives to standardize the concept of multi-agent cloud robotics, the efficient resource allocation and service provisioning for the robotic applications is still a major concern. Therefore, a good number of research works are currently being conducted to develop various resource allocation and service provisioning policies for multi-agent cloud robotics. This survey focuses on summarizing these works in a systematic manner. 
%
%
\subsection{Related Surveys}   
%
Several surveys in the literature highlight the importance of resource allocation. However, they do not review and address the challenges of resource allocation problem in multi-agent cloud robotics. For example, only a set of approaches to assign robots for task execution in multi-robot system are reviewed in \cite{khamis2015multi,mosteo2010survey}. In comparison, the resource allocation problem is well studied in cloud computing \cite{manvi2014resource,toosi2014interconnected,yousafzai2017cloud,luong2017resource}. The cloud resource provisioning and scheduling algorithms are summarized in \cite{singh2016survey,huang2013survey,mastelic2015cloud,zhan2015cloud}. Similarly, resource provisioning and auction-based radio resource allocation mechanisms for wireless and mobile systems are discussed in~\cite{zhang2012auction} and~\cite{budzisz2014dynamic}, respectively. Computation offloading is another widely explored technique in the field of wireless and mobile systems \cite{maallawi2014comprehensive,rebecchi2014data}. Existing offloading and service provisioning techniques in mobile cloud computing is comprehensively surveyed in \cite{shiraz2012review,
othman2013survey,xu2013survey,abolfazli2013cloud,sanaei2013heterogeneity}. Conversely, the computation offloading techniques in mobile edge computing and multi-access edge computing are briefly described in \cite{mach2017mobile,mao2017survey,wang2017survey} and \cite{taleb2017multi, rodrigues2019machine}, respectively.
\par On the other hand, in literature, there exist a notable number of surveys on cloud robotics. However, most of them conduct surveys on architectural perspectives \cite{survey_cloudRObotics_architecture,survey_CloudRoboticsandAutomation,Survey_curretSatus_CloudRobotics}. Additionally, the complexities and limitations of cloud robotics including the necessity of efficient resource and task allocation policies are partially discussed in \cite{saha_CloudRobotics_Survey}. The robotic cooperation techniques for critical CPSs such as disaster management and smart manufacturing are reviewed in \cite{Robotic_Cooperation} and \cite{Smart_Manufacturing}, respectively. Nevertheless, there does not exist any survey in the literature that summarizes the resource allocation and service provisioning techniques including resource pooling, computation offloading and task scheduling jointly for multi-agent cloud robotics. A comparative study of our survey with respect to the existing works is illustrated in Table \ref{Tab:summary_rel_survey}. In the following sections, different aspects of resource allocation and service provisioning in multi-agent cloud robotics are reviewed in detail.
\section{Resource Allocation in Multi-Agent Cloud Robotic Systems}\label{resource_allocation}
\begin{figure*}[ht]
\centering
\includegraphics[width=6in]{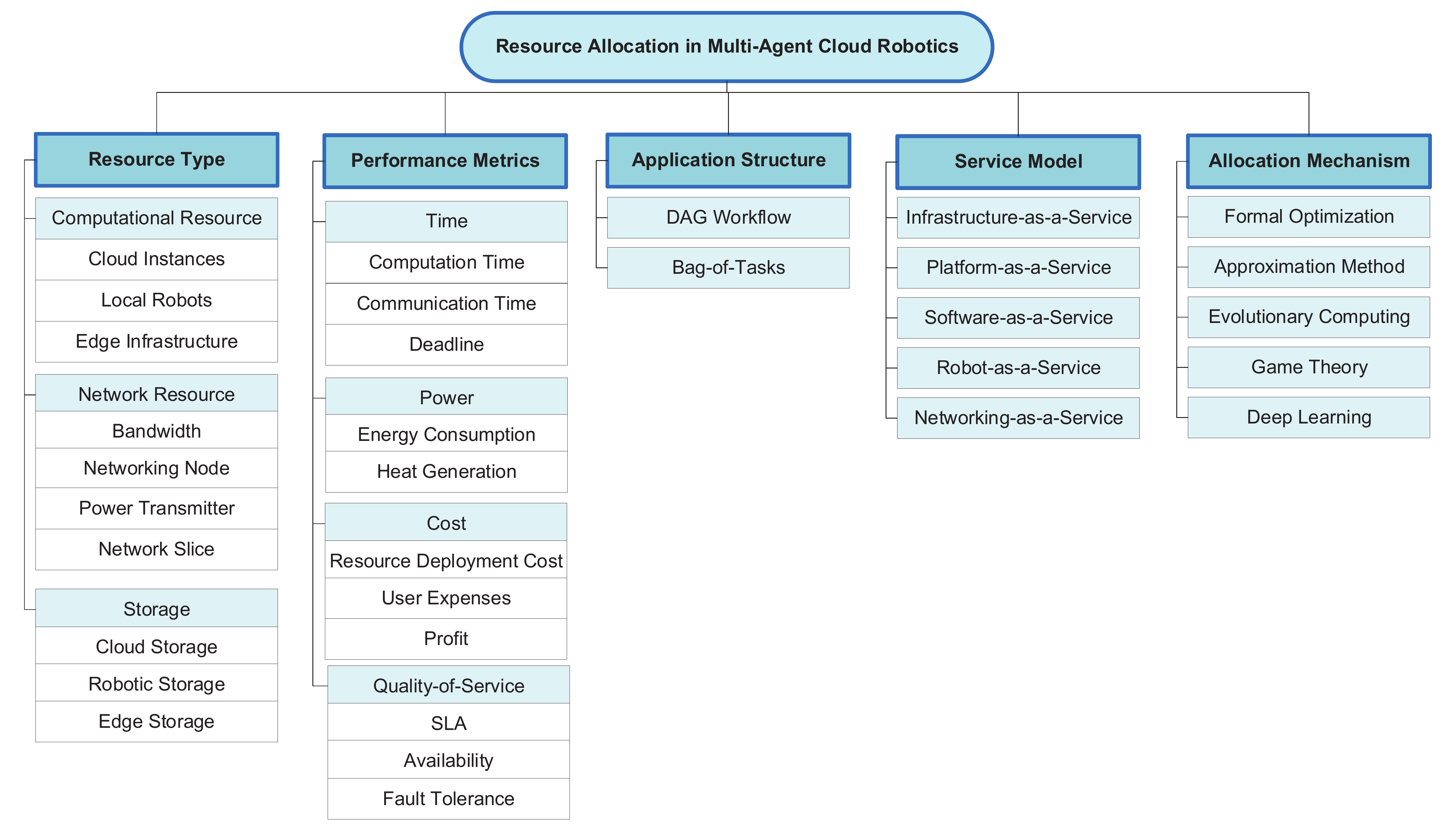} 
\caption{Taxonomy of resource allocation in multi-agent cloud robotics.}
\label{fig_taxonomy}
\end{figure*}

The robotic applications of real-world systems possess an inherent demand of faster processing. The optimal allocation of resources can play a vital role in meeting this requirement of the applications to a great extent. Usually, robotic applications encapsulate single or multiple robotic tasks. Basically, resource allocation defines the assignment of a resource to a task based on its availability and the QoS requirements of the task. However, these requirements of robotic tasks vary from one system to another. Furthermore, the resource availability in robot, edge and cloud instances changes with the course of time unless it is reserved. For example, the QoS requirements of tasks for a robotic surgery application is quite stringent than the tasks of a robotic parcel delivery application. Similarly, the robotic application that monitors environmental context occupies resources for a longer period compare to robot enabled autonomous irrigation application where the resource acquisition occurs only for a specific time. During such cases, the detailed exploitation of resource type, performance parameter, application structure, service model, allocation mechanism and evaluation method can guide towards the efficient resource allocation. In this work, these aspects of resource allocation are narrated thoroughly with the help of a taxonomy shown in Figure \ref{fig_taxonomy}.  
\subsection{Resource Type} 
As the robotic applications consist of latency sensitive, compute and data-intensive tasks, a wide variety of resources are required to execute these applications. Based on the necessities of the tasks, allocation of three types of resources, namely, computational resource, network resource and storage are discussed in the literature.
\subsubsection{Computational resource} The resources that are used for data processing and executing the manifold tasks of robotic applications are known as computational resource. Cloud instances, local robots and edge nodes are the basic computational resources.
\begin{itemize}
\item{{\em Cloud instances}:} Cloud instances are allocated to robotic applications for large-scale computation. Cloud service providers virtualize the computing servers and offer a variety of computing instances including virtual machines (VMs) and containers to the multi-agent robotic systems. The authors of \cite{sun2017cost,chen2018qos,liu2018reinforcement} focus on allocating cloud instances for executing robotic applications. Moreover, cloud instances are dynamically configurable according to the resource requirements of the applications. However, the integration of robotic systems with the cloud instances depends on the category of cloud infrastructure namely public, private, edge and hybrid cloud.
\begin{itemize}
\item{{\em Public cloud}:} In this of cloud infrastructure, the instances are accessed on-demand over the Internet \cite{Aissam2019}. It bestows the greatest level of efficiency in shared resources where the cloud users must pay as per the usage. In the literature, \cite{Rapyuta,beksi_object_recognition,infrastructure-for-robotic-applications,ramharuk2014cloud,Robot-As-a-Service,doriya2017development,Mohanarajah_2,toffetti2017cloud,yun2017towards,ali2018fastslam,PPAAS,limosani2016enabling, tian2017cloud,vick2016model,merle2017mobile}
deploy public cloud infrastructure for the robotic applications. However, the resources in public cloud often become vulnerable to security issues. Due to geographical distance, the data transfer also gets delayed while executing robotic tasks using public cloud.
\item{{\em Private cloud}:} In this type of cloud infrastructure, instances are accessed through a private network. It supports versatile and convenient end-to-end interaction, where the management of the resources are conducted within the local data centres. \cite{wen2016towards,koubaa2014service,C2TAM,turnbull2013cloud,vick2015robot,hong2018cloud} adopt private cloud for their systems. Private cloud provides more security than public cloud. However, the maintenance of private cloud requires dedicated administrator, which is not always cost efficient for the multi-robot systems. 
\item{{\em Hybrid cloud}:} It is a combination of public and private cloud infrastructure with an additional orchestration and automation support \cite{Aissam2019}. In this type of cloud infrastructure, the tasks of real-time or mission-critical applications are run on the private cloud instances while the public cloud instances deal with the delay tolerant tasks \cite{li2016toward}. Hybrid cloud for multi-agent cloud robotics is considered in \cite{DAvinCi,Cloudroid,furrer2012unr,miratabzadeh2016cloud,Robots-as-a-Service-Search-and-Rescue,chen2016hybrid}. Hybrid cloud can also encapsulate edge resources as mentioned in \cite{beigi_smartcity_cloudRobotics,li2016toward,ramharuk2014cloud}. Nevertheless, the goal of hybrid cloud is to ensure a scalable environment for networked robots where the instances can be extended either from public or private cloud infrastructure based on the task requirements. However, such management of instances in hybrid cloud becomes a challenging issue when there exist an explicit data-dependency among the robotic tasks.
\end{itemize}  
\par As there exist pros and cons of each type of cloud infrastructure, the resource allocation policies for multi-agent cloud robotics require to observe them deliberately while dealing with the robotic applications.
\item{{\em Local robots}:} The on-board CPU within the robots also supports the task execution of robotic applications. In multi-agent cloud robotic systems, local resources can be shared among the connected robots to complete the task execution in a collaborate manner. For example, the local robots are used for robotic task execution in \cite{lwowski2017task,pandey2015dynamic,li2018latency, li2017subtask,xie2019loosely}. However, the on-board computing components often become exhausted because of the size, shape, power supply, motion mode and working environments of the robots \cite{survey_cloudRObotics_architecture}. Resource re-configuration is also infeasible once the robots are built for a particular system. Despite of such constraints, it is still conducive to use the local robotic resources for the hard real-time robotic applications since they offer comparatively lesser data transfer delay.   
\item{{\em Edge infrastructure}:} The inter-communication delay in sending the computation request to cloud and receiving the response at the local robots resist the whole system in meeting the QoS requirements of robotic applications. Local robots have limited energy and computational capacity, which further hamper the execution of large-scale real-time applications. To address these limitations, the concept of edge computing is introduced between the robots and the cloud in multi-agent cloud robotics. Edge infrastructures such as cloudlet or edge cloud and fog nodes facilitate task execution in proximity of the robotic systems and consequently enhances the QoS of the applications.
\begin{itemize}
\item{{\em Cloudlet or edge cloud}:} Basically Cloudlet or edge cloud is an extension of cloud that can provide virtualized resources and assist multi-robot systems in executing the latency sensitive and compute-intensive tasks \cite{afrin2018energy}. Integrating both edge and cloud computing with the robotics, a platform is introduced in \cite{beigi_smartcity_cloudRobotics}. Moreover, the local robots and edge cloud are allocated for reducing communication delay in \cite{afrin2019multi,afrin2018energy,antevski2018enhancing}.
\item{{\em Fog node}:} Any personal computer, mobile device, smart edge device, car, sensors, traditional networking devices including set-top boxes, gateway routers, smart switches, proxy servers equipped with computational resources are generally used as potential fog nodes and they are deployed in a distributed manner across the edge \cite{mahmud2018fog,8894519}. Rather than forwarding the tasks of latency sensitive robotic applications to a centralized cloudlet or private cloud, they can be processed more efficiently at the edge network using fog nodes \cite{mahmud2018latency}. The idea of utilizing fog nodes in multi-agent cloud robotics is addressed in \cite{mohamed2018fog,fogRobotics,karjee2017distributed ,botta2019cloud,ichnowski2020fog}.
\end{itemize}
\end{itemize} 
\par To summarize, although cloud resources offer higher computational power, they include additional delay for data transfer which is not suitable for latency sensitive tasks. Although robots are feasible to support hard real-time tasks, they have energy constraints. As a viable solution, in multi-agent cloud robotics the concept of edge computing is augmented. However, edge resources are also restricted by the availability, energy consumption and deployment cost. Therefore, the selection of appropriate resource for executing all the tasks associated with a robotic application is challenging problem in multi-agent cloud robotics.
\subsubsection{Network resource} The physical and logical resources within a network that help connectivity, communication and data transmission are referred as network resource. In literature, the allocation of network resources are explored in \cite{wang2012game,wang2016pricing,el2016environment,HierarchicalAuction}. A mesh orchestration of networking nodes makes the sharing of network resources possible in multi-agent cloud robotic systems. To maintain guaranteed data flow over the shared network, bandwidth, network slices and power transmitters are allocated as the network resources.
\begin{itemize}
\item{{\em Bandwidth}:} The amount of data that can be transmitted in a fixed amount of time defines the bandwidth of a network. Optimal bandwidth allocation is very important in multi-agent cloud robotics for retrieving and sharing multi-sensor data, communicating over the network, and getting the robotic services from cloud and edge instances with QoS guarantee. In literature, \cite{wang2012game,wang2016pricing,el2016environment} focus on bandwidth allocation in multi-agent cloud robotics.
\item{{\em Networking node}:} The connection points for data transmission within a network such as routers, switches, multiple antennas, wireless access points and base stations are referred as networking node \cite{CloudResourceAllocation_Survey_2}. Even in \cite{HierarchicalAuction}, the robots are considered as the networking nodes. However, optimal placement of the nodes in the network is very crucial for sending and receiving data in a timely manner.
\item{{\em Wireless power}:} By harnessing the mobility of robots, various hazardous tasks can be done in remote environments. However, the main limitation of mobile robots is the short battery lifetime. Therefore, while conducting remote jobs, mobile robots are required to return to a base station for charging, manually battery replacement or tethering \cite{west2019debris}. However, these operations are time-consuming. As a sustainable solution, wireless power transfer has emerged to address the energy limitations of robots and provide a backup of on-board battery \cite{cheah_watson_lennox_2019}. In literature, there exist several techniques for wireless power transmission including radiative power transfer (microwave, laser) and non-radiative power transfer technologies are (inductive, magnetic resonance, capacitive and acoustic) \cite{shinohara2011power, chen2011contactless}. However, in multi-agent cloud robotics, the core challenge of transmitting wireless power is to mitigate the mutual interferences.
\item{{\em Network slice}:} A network slice is an end-to-end logical network that runs on a shared physical infrastructure to provide negotiated network services. It virtualizes the network infrastructure including terminal, access network, core network and make them deployable across multiple network providers \cite{samdanis2016network, kalor2018network}. In multi-agent cloud robotic systems, the utilization purpose of network infrastructure varies from one robotic system to another. For example, a multi-robot system may require high reliable data transfer, whereas the other one may require low latency with high data rate. Network slicing is an essential technology in multi-agent cloud robotics to handle such heterogeneous network service requirements of robotic systems. How the network slices will be composed and how much resource will be allocated for a particular network slice and how the network slices will be distributed across multiple robotic systems are the key research questions to solve in this domain. However, the efficient mechanisms for wireless power allocation and network slicing are still in their early phase of research and need to be explored more.
\end{itemize}
\subsubsection{Storage} This type of resources are used for storing data, information sharing and collective learning while executing any robotic application. In multi-agent cloud robotics, the cloud-based data servers, foglets, built-in memory of local robots offer the storage facilities. However, most existing works in multi-agent cloud robotics pay attention to computational and network resource allocation, while a few including \cite{chen2016smart} focus on allocating storage resource. Additionally, in multi-agent cloud robotics, it is very crucial to select the data and their storage location so that ease of data access can be ensured during real-time interactions.
\begin{itemize}
\item{{\em Cloud storage}:} Cloud infrastructure inherently serves an extensive storage space to the robots. Virtualization techniques and NoSQL databases play the vital roles in storing the information in cloud storages. These information are explicitly maintained, operated, and managed by the cloud service providers. Additionally, cloud storages support collective learning by providing application programming interfaces for analysing historical behaviours according to task requirements \cite{survey_cloudRObotics_architecture}. However, in multi-agent cloud robotic system while executing robotic applications, the exchange of information should be happening in real-time. In this case, an uninterrupted Internet connectivity is a must. Moreover, there are some sensitive applications where the data privacy is regarded as one of the QoS requirements. To deal with such applications, the selection of appropriate cloud infrastructures is very important.
\item{{\em Robotic storage}:} The built-in memory of the local robots can collectively support the additional storage resources for multi-agent cloud robotics. A group of connected robots usually share information among each other while executing the robotic applications in a collaborative manner. Robotic storage allows the robots to store necessary information locally so that the robots residing outside the communication range, can still participate in learning and data collection. However, even after the augmentation of robots, this type of storage offers a very limited capacity and does not support easy re-configuration. Therefore, they are considered well-suited for static environment. For dynamic environment, the cloud storage outperforms the local robotic storage.
\item{{\em Edge storage}:} Although cloud computing helps robots in sharing information, it adds high latency during the execution of hard real-time applications. In the form of fog servers and foglets, edge computing offers storage resources to multi-agent cloud robotic systems. They can easily store necessary information and update data according to the application requirements using the adaptive and augmented memory \cite{fogRobotics}. Additionally, in \cite{kattepur2016resource} and \cite{gudi2017fog}, the idea of using local robot as a fog storage server is introduced. Nevertheless, the multi-agent cloud robotic system needs to be concerned about the security and privacy preservation of data while utilizing edge storage. Therefore, recently much emphasize is given on developing efficient algorithms for secured edge and fog storage management in multi-agent cloud robotics.
\end{itemize}
\subsection{Performance Metrics} While allocating resources for robotic applications in multi-agent cloud robotics, different metrics including time, power, cost and QoS are exploited to monitor their performance and meet the service requirements. Various aspects of these parameters are discussed below.
\subsubsection{Time} It is one of the important metrics to measure the efficiency of resource allocation policies. In multi-agent cloud robotics, the minimization of computation time and communication time and the timely execution of tasks within deadline is desired for the improved performance.
\begin{itemize}
\item{{\em Computation time}:} The time required for executing a task on computational resources is defined as computation time. It mostly depends on the processing speed of assigned resource and size of the task. The authors of \cite{chen2018qos,pandey2015dynamic,li2018latency,li2017subtask,kattepur2017priori} design their resource allocation schemes to minimize the application completion time.
\item{{\em Communication time}:} It basically refers to the networking delay while exchanging data among different entities of multi-agent cloud robotic systems. Reduced communication time reflects the efficiency of network resources in assisting the application execution on the computational resources.  \cite{lwowski2017task,li2018latency,wang2016pricing,HierarchicalAuction} address the minimization of communication time.
\item{{\em Deadline}:} The maximum tolerable delay of a system in receiving application service delivery is specified as the deadline. Service delivery deadline plays a vital role in characterizing latency sensitive (real-time) and latency tolerant (non real-time) applications. It also functions as a decision parameter while allocating resources to meet the application's QoS requirement. Among the existing works, \cite{afrin2019multi,afrin2018energy} emphasise on the satisfaction of application deadline while allocating resources for the tasks of robotic application.
\end{itemize}
\par Additionally, there are some other time-based metrics such as resource discovery time, service access time, data sensing frequency of sensors which are also required to be considered as performance indicator while allocating resources in multi-agent cloud robotics.
\subsubsection{Power} The minimization of overall power usage is another important indicator of improving performance in multi-agent cloud robotics. Power usage is mainly focused on the energy consumption and heat generation aspects of the computing resources.    
\begin{itemize} 
\item{{\em Energy consumption}:} It is calculated in terms of computation time and per unit time energy requirements of the resources to execute a task. As robots are usually energy-constrained, low battery power can degrade the performances of multi-agent cloud robotic systems. At the same time, the energy consumption of cloud resources causes huge capital expenditure and operational cost. Hence, considering the energy parameter, appropriate resources are required to be allocated so that with less energy, desired performance of the system can be ensured. In the literature, very few works including \cite{afrin2019multi,afrin2018energy,chen2018qos,liu2018reinforcement} aims at minimizing the energy consumption of resources while assigning the applications.  
\item{{\em Heat generation}:} It refers to the thermal effect on computing resources while executing different tasks of robotic applications. As the heat emission increases with the number of resources, it increases the probability of hardware system failures and carbon footprint. Therefore, it is essential for service provider to improve the cooling system and generate less heat while handling the application tasks. In \cite{sun2017cost}, the reduction of both energy consumption and heat generation are emphasized to meet the power usage in multi-agent cloud robotic systems. 
\end{itemize} 
\par Apart from these power-related metrics, the detail analysis of residual battery lifetime of end devices and the exploitation of energy characteristics of communication medium are also required to design a power-efficient multi-agent cloud robotic system. However, such initiatives are subjected to extensive research.
%
\subsubsection{Cost} This performance parameter preserves the interest of both service providers and end users in multi-agent cloud robotics. Since service providers are responsible for facilitating computing services to numerous robotic systems, they always aim at making a comprehensive profit through extreme resource utilization \cite{cran_my}. Similarly, the users of robotic systems demand maximum Quality of Experience (QoE) without surpassing the budget. In such circumstance, cost-aware resource allocation mechanism preserves the economic benefits of both service provider and users of robotic systems. Resource deployment cost and user expenses are the main cost-related factors in multi-agent cloud robotics. 
\begin{itemize} 
\item{{\em Resource deployment cost}:} The associated cost for infrastructure placement in multi-agent cloud robotics defines the resource deployment cost. It includes the expenses for placing sensors, local robots as well as creating the VMs and containers in cloud. Networking cost is also encapsulated in resource deployment cost which is directly related to the charges of data transferring, bandwidth, and networking nodes. In the literature, most of the works emphasising the cost-related issues of multi-agent cloud robotics, aim at minimizing the resource deployment cost. For example, \cite{sun2017cost,li2017subtask,wang2012game,wang2016pricing} reduce the resource deployment cost while allocating resources for the robotic applications.
\item{{\em User expenses}:} The amount paid by the robotic systems and its users to occupy the resources for executing tasks of applications is known as user expenses. In multi-agent cloud robotics, service providers determine the per unit time usage cost of resources to execute a task. On the other hand, robotic systems urge to access the desired computing services within the budget constraint. Thus, the minimization of user expenses becomes an important element while allocating resources in multi-agent cloud robotics. There are several works in the literature including \cite{afrin2019multi,pandey2015dynamic} that investigate user expenses during resource allocation.  
\item{{\em Profit}:} The net business gain from the revenue and service execution cost is considered as the service provider's profit. The profit of service providers while assigning resources to the robotic tasks is discussed in \cite{liu2018reinforcement}. Moreover, the service provider's profit in multi-agent cloud robotics depends on the service charges and the business costs associated with a particular robotic service. These aspects are brought into attention in \cite{chen2016smart}. Furthermore, the pricing model for any computing service is determined by the operators and it depends on the resource availability. While allocating resources in multi-agent cloud robotics, the selection of pricing model also becomes an important concern as it helps to maximize profit within the budget constraints.
\end{itemize} 
\subsubsection{Quality-of-Service} The most significant performance parameter in multi-agent cloud robotics is QoS. It refers to the distribution of resources according to the requirements of the applications. QoS is driven according to the Service Level Agreement (SLA) between the service provider and robotic systems. The availability and fault tolerance of resources also play vital roles in defining the QoS. 
\begin{itemize}
\item{{\em Service Level Agreement (SLA)}:} An agreement between the cloud service provider and robotic system as per the usage of resources is denoted as SLA. In multi-agent cloud robotics, service provider is responsible to allocate resources according to the requirements of an application and avoid the SLA violations. The performance of a system in terms of SLA depends on ratio of the number of successfully executed tasks and the number of total assigned tasks \cite{CloudResourceAllocation_Survey_2}. In \cite{chen2018qos,wang2016pricing,el2016environment}, QoS is enhanced by minimizing the SLA violations.
\item{{\em Availability}:} The accessibility, usability, scalability of the resources while executing the applications defines the availability of resources in multi-agent cloud robotics. The service providers must ensure resource availability prior to allocating them to any robotic application. However, in \cite{xie2019loosely}, SLA and resource availability are simultaneously counted as the QoS parameters.
\item{{\em Fault tolerance}:} This property enables a computing paradigm to continue the execution of assigned tasks even after the failure of any computing and networking entities \cite{Fault_tolerance}. In multi-agent cloud robotics, uncertain anomalies such as link failures, power shortage, limited bandwidth and interference can hamper the simultaneous execution of the robotic application. Hence, it is very important to ensure fault tolerant during resource allocation. Additionally, security and reliability, maintainability can also drive the fault tolerance and collectively influence resource allocation.
\end{itemize} 
%
\begin{table*}[!htb]
\centering 
\caption{Summary of works concentrating on resource allocation in multi-agent cloud robotics}\label{Tab:summary_reosurce_allocation} 
\scriptsize
\begin{tabular}{|p{0.7 cm}|p{2.2cm}|p{2.4cm}|p{1.4cm}|p{1.3cm}|p{1.5 cm}|p{1.35cm}|p{2.8cm}|p{2cm}|}
\hline 
\textbf{Work} & \textbf{Resource Type} & \textbf{Performance Metrics} & \textbf{Application Structure}& \textbf{Service Model}&\textbf{Allocation Mechanism}& \textbf{Evaluation Method} & \textbf{Results}& \textbf{Use Case}\\\hline
\cite{li2017subtask} & Computational resource (cloud, local robot)&	Time (computation); Cost (resource deployment) & Workflow &PaaS	&Evolutionary computing (GA)& Simulations & Reduction of  computation time and cost up to 10\%.& Smart factory.\\\hline
\cite{xie2019loosely}&	Computational resource (cloud, local robot)& QoS (SLA, availability)&	Workflow &PaaS,RaaS & Evolutionary computing (GA)& Simulations, Testbed& Resources are available for up to 98\% tasks.& Generalized robotic applications.\\\hline
\cite{afrin2018energy} & Computational resource (local robot, edge cloud) &	Time (deadline); Power(energy consumption) & Workflow & IaaS &Evolutionary computing (NSGA-II)& Simulations& Reduction of time up to 15\% and energy consumption up to 10\%.& Smart factory.\\\hline
\cite{afrin2019multi} & Computational resource (local robot, edge cloud) & Time (deadline); Power(energy consumption); Cost (user expenses) &  Workflow &IaaS&	Evolutionary computing (NSGA-II)& Simulations& Up to 18\% improvement in optimizing time, energy and cost.&Smart factory.\\\hline
\cite{wang2012game}& Network (bandwidth) & Time(communication); Cost(resource deployment) & Bag-of-Tasks &SaaS& Game Theory (Stackelberg game)&Simulations&  No time out in dynamic buffering.& SLAM.\\\hline
\cite{wang2016pricing}&	Network (bandwidth)& Time (communication); Cost  (resource deployment)&  Bag-of-Tasks & SaaS&Game Theory (Stackelberg game)& Simulations, Testbed& Time and cost increase with number of node. & SLAM.\\\hline
\cite{HierarchicalAuction} &Network (bandwidth, node)& Time (communication)& 	Bag-of-Tasks	& SaaS&Game Theory (Auction)&Simulations, Testbed& Up to 35.9\% increase in bandwidth usage.& Navigation. \\\hline
\cite{lwowski2017task}& Computational resource (cloud, local robot)&	Time (communication)&  Workflow & SaaS&Game Theory (Auction)&Simulations, Testbed& Speed and scalability of the system increases. & SLAM.\\\hline
\cite{pandey2015dynamic}& Computational resource (cloud, local robot)	& Time (computation); Cost (User expenses)& Workflow &PaaS,SaaS& Approximation (Heuristic)& Simulations& Performs 77\% better than using only local resources.& Underwater disaster management. \\\hline
\cite{chen2018qos} & Computational resource (cloud)	& Time (computation); Power (energy consumption); QoS (SLA)	&  Workflow & RaaS,SaaS&Approximation (Heuristic)& Simulations& Saves 99.3\% time, reduces  23.8\% energy consumption and satisfy QoS.& SLAM, grasping, navigation.\\\hline
\cite{li2018latency} & Computational Resource (cloud, local robot)	& Time (computation, communication) & Workflow & SaaS& Optimization (ILP)& Simulations& Produces the optimal solution with minimal latency & Disaster management.\\\hline
\cite{sun2017cost} & Computational resource (cloud)& Power (energy, heat generation); Cost (resource deployment) 	& Bag-of-Tasks & IaaS & Approximation (Greedy)&Simulations& 22\% better near-optimal solution.& Generalized robotic applications.\\\hline
\cite{chen2016smart} &	Storage (cloud storage)	& Cost (resource deployment cost)&	 	Bag-of-Tasks &	IaaS& Approximation (Heuristic)& Simulations&  The cost of servers are minimized. & Generalized robotic applications.\\\hline
\cite{el2016environment} &	Network (bandwidth)& QoS (SLA) &Bag-of-Tasks &	SaaS&Approximation (Fuzzy)& Simulations& Reduction of bandwidth usage and successful execution of collaborative tasks. &Robotic teleoperation.\\\hline
\cite{liu2018reinforcement} & Computational resource (cloud)& Power (energy consumption); Cost (Resource deployment) &	Bag-of-Tasks &IaaS&	Deep learning (RL)& Simulations& RL scheme performs better than GA under the condition of limited cloud resources.&Generalized robotic applications.\\\hline  
\cite{li2020energy}& Network (Power)& Time (Communication);Power (energy consumption) &	Bag-of-Tasks &IaaS, NaaS& Approximation (Successive convex)& Simulations& Energy efficiency of UAV increases.& Trajectory design.\\\hline
\end{tabular}  
\end{table*}
\subsection{Application Structure} In multi-agent cloud robotics, a robotic application consists of single or multiple tasks. Based on the dependencies and requirements of different tasks, the robotic applications follow various application structures for execution. Directed Acyclic Graph (DAG) workflow and Bag-of-Tasks (BoT) are the mostly used models for robotic applications \cite{iosup2010grid}.  
\subsubsection{DAG workflow} A workflow contains a sequential series of tasks having data dependencies among them. In most cases, it is described as a DAG, where the nodes are tasks, and the edges denote the task dependencies. The execution order of tasks in a workflow can be either serialized or parallel. This type of structure depicts how the tasks are interrelated that eventually helps in allocating the resources optimally with better efficiency \cite{afrin2019multi}. \cite{afrin2018energy,xie2019loosely,chen2018qos,lwowski2017task,pandey2015dynamic,li2018latency,li2017subtask} 
adapt DAG workflow while allocating resources for the tasks of robotic applications.
\subsubsection{Bag-of-Tasks} Generally, Bag-of-Tasks (BoT) is applied to those applications where no data dependencies exists among the tasks. BoT applications are parallel applications comprised of independent but similar tasks \cite{iosup2010grid}. Identical tasks from different applications as well as from the same application can form BoT application model, which assists concurrent utilization of resources in distributed environments. In literature, \cite{chen2016smart,sun2017cost,liu2018reinforcement,wang2012game,wang2016pricing,el2016environment, HierarchicalAuction} use BoT for modelling and allocating resources for multiple applications. 
\par Nevertheless, while allocating the resources for the tasks of robotic applications, the multi-tenancy facility of resources is required to be taken into account. It helps in determining whether multiple resources will execute the same task or vice-versa. Additionally, the resources need to be allocated according to the delay sensitivity of applications even it requires to go beyond their architectural differences.
\subsection{Service Model}
There are five types of service models, namely, Infrastructure-as-a-Service (IaaS), Platform-as-a-Service (PaaS), Software-as-a-Service (SaaS), Robot-as-a-Service (RaaS) and Networking-as-a-Service (NaaS) that a multi-agent cloud robotic system can widely support. 
\subsubsection{IaaS} It refers to the pool of virtualized resources used for executing the tasks of robotic applications. Generally, virtual machine, containers, servers, storage are delivered as IaaS. The frameworks developed in \cite{chen2016hybrid,C2TAM,miratabzadeh2016cloud} yield IaaS provided by the cloud service provider to the robots. Moreover, there are some application specific researches for disaster management \cite{Robots-as-a-Service-Search-and-Rescue}, vision acquisition \cite{turnbull2013cloud}, where the integration of customized IaaS with robotic systems are considered explicitly. Additionally, in \cite{infrastructure-for-robotic-applications}, the confederation of cloud infrastructure is considered for supporting robotic systems. While providing IaaS for the robotic systems, the infrastructures are required to support diversity so that these resources can be beneficial to heterogeneous robots. In addition, the resource distribution among multiple robotic systems should be vibrant enough to fulfil the requirements. Furthermore, the infrastructure should be extensible in terms of adding new functionalities such as self-healing and fault tolerance.
\subsubsection{PaaS} It refers to the operating systems, application run-time environments, programming interfaces, databases, web servers that help in developing robotic applications. In literature,  
\cite{Mohanarajah_2,toffetti2017cloud,antevski2018enhancing} discuss PaaS for multi-agent cloud robotics. The PaaS highlighted in \cite{Mohanarajah_2,toffetti2017cloud} offer cloud aided map optimization and autonomous robot patrol for SLAM application. The feasibility of edge computing-enabled PaaS for robotic applications is investigated in \cite{antevski2018enhancing,beigi_smartcity_cloudRobotics,tanwani2019fog}. A platform is introduced in \cite{tanwani2020rilaas} to offer robot inference and learning as-a-Service for grasp planning and objection recognition.  
\subsubsection{SaaS} It refers to the on-demand robotic applications over the Internet such as object recognition, path planning, map building, speech translation, knowledge base etc. In the literature, \cite{wen2016towards,DAvinCi,ali2018fastslam,PPAAS,limosani2016enabling, vick2015robot,vick2016model} discuss SaaS for multi-agent cloud robotics. In \cite{wen2016towards}, robotic software packages are migrated to cloud so that multiple robots can simultaneously access them. The parallelization of robotic algorithms using cloud resources is discussed in \cite{DAvinCi}. Robotic path planning is delivered as a software in \cite{PPAAS}. Additionally, the motion planning and control of industrial robots using cloud resources are discussed in \cite{vick2015robot,vick2016model,limosani2016enabling,doriya2017development}. Cloud based computation are used in \cite{kehoe2013cloud} for 3D robot grasping. SaaS is also offered in \cite{beksi2014point} for robot vision tasks including detection, segmentation, and object classification.
\subsubsection{RaaS} It refers to robots that can be dynamically combined as-a-Service to execute specific applications \cite{Survey_RobotCloud}. The robots are distributed in different locations and can be accessed as service through multi-agent cloud robotic system for executing various robotic applications. RaaS is supplied in terms of computing, sensing, motion planning, navigation, and perception service. In literature, \cite{Robot-As-a-Service,furrer2012unr,koubaa2014service,tian2017cloud,merle2017mobile} particularly aims at serving RaaS. The idea of using a robot as an all-in-one service-oriented architecture (SOA) unit to simultaneously perform the responsibilities of service providers, service brokers, and service clients is revealed in \cite{Robot-As-a-Service}. \cite{furrer2012unr} enables robots and sensors to expose their functions to a virtualized resource pool and forms a cloud of robots. Robots computation facilities are offered as a cloud service to the end users in \cite{koubaa2014service}. Moreover, the on-board sensors of the robots offer sensing-as-a-Service by perceiving information from the environment and support the processing and actuation, based on the sensed data \cite{nakashima2015fourth}. Sensing capability of the robots is provided as a service in \cite{chung2009door} for door-opening control problem in the real environment. Visual sensing is also offered in \cite{bistry2010cloud,desouza2002vision} by supporting the collaborative execution of image processing tasks on the robots and cloud. Furthermore, a prototype is developed for robotic grasping application in \cite{tian2017cloud} to offer robotic and automation as-a-Service. Additionally, an access control policy for mobile robots is proposed in \cite{merle2017mobile}. 
\subsubsection{NaaS} To support the end devices having limited networking capacity and communication constraints, mobile robots such as drones or unmanned aerial vehicles (UAVs) provide Networking-as-a-Service (NaaS). Residing in proximity of the end devices, these robots act as either a repeater, or wireless range extender, or base station, or an access point. In literature, UAVs perform as wireless base stations in \cite{mozaffari2016efficient} that provide coverage for ground users. In \cite{nasir2019uav}, UAVs are deployed as flying base stations to support the coverage and throughput of wireless communication. UAV serves as a data ferry between the source base station and multiple destination receivers in \cite{8930577}. UAVs facilitate data transmission of IoT devices in \cite{fu2020joint,motlagh2017uav}. UAV can also play the role of aerial cloudlet to collect and process the computation tasks offloaded by ground users \cite{li2020energy}. Moreover, drones and UAVs are expected to be an important component of 5G or beyond 5G cellular architectures to make wireless broadcast or point-to-multipoint transmissions easier \cite{sekander2018multi}. However, while offering NaaS in multi-agent cloud robotics, several issues such as the energy consumption minimization of service robots, flying time minimization with optimal trajectory planning of UAVs need to be focused for efficient resource allocation and service provisioning.
\par In literature, there exist other research works that deal with multiple service models simultaneously to facilitate various robotic applications. For example, \cite{doriya2017development,yun2017towards,li2016toward,pillajo2015implementation} offer IaaS, PaaS and SaaS collectively to multi-agent cloud robotics. Additionally, In \cite{Rapyuta}, the architecture of Rapyuta platform is discussed that can be used by the developers to model and design robotic applications, and the applications running on this platform can be directly accessed as a software service. In most of the cases, IaaS, PaaS, and SaaS are jointly exploited for web applications. However, there exist significant differences between general web applications and robotic applications. Web applications are typically stateless, single processes that use a request-response model to talk to the client. On the other hand, robotic applications are state-driven, multi-processed, and require a bidirectional communication with the client \cite{Rapyuta}. Therefore, while provisioning services in multi-agent cloud robotics, the selection of appropriate service models as per the requirements of a particular robotic system is very important.
%
\subsection{Mechanism} The successful execution of robotic applications in multi-agent cloud robotics largely depends on the efficient selection of resource allocation mechanisms. The design of resource allocation mechanism varies in harmony with the system environment and application requirements. In the literature of multi-agent cloud robotics, the formal optimization, approximation method, evolutionary computing, game theory, heuristic and deep learning techniques are adopted widely for resource allocation.
\subsubsection{Formal optimization} It helps to solve any optimization problem with single objective or multi-objective under certain constraints. In multi-agent cloud robotics, depending on the key performance metrics of a particular system, the optimization method is selected. For example, in a smart factory, the main objective of resource allocation is to reduce the resource deployment cost. On the other hand, in a smart farm, the resource allocation objective is to complete the trajectory of the UAVs within their battery lifetime. Based on the characteristics of optimization function such as convex (minimization) or concave (maximization) and the relation among the constraints, the choice of optimization method (linear programming, integer programming, non-linear programming, combinatorial, stochastic) is made in multi-agent cloud robotic system. For example, \cite{li2018latency} linearizes a mixed-integer non-linear problem into an integer linear programming (ILP) model and reduces the computation and communication time during robotic application execution. Similarly, \cite{li2018latency} allocates resources with minimal latency. However, for complicated optimization problem, it is not always feasible to find the solution satisfying all the constraints, especially when the resource allocation is required to conduct in real-time.
\subsubsection{Approximation method} Approximation methods help to find a solution of any complex optimization problem within a reasonable time frame \cite{pearl1984heuristics}. In literature, heuristic, greedy, fuzzy, dynamic programming are used for solving resource allocation problem in multi-agent cloud robotics. For example, a heuristic approach that allocates computational tasks among the cloud and the local robots and minimizes execution time and cost is applied in \cite{pandey2015dynamic}. The approach is evaluated by simulations and the results show that it helps to select computing resources efficiently and reduces the time and cost by 77\% for using the local resources only. While allocating the computational resources, \cite{chen2018qos} also uses a heuristic algorithm and finds a near-optimal value for time and energy consumption. This algorithm is evaluated by simulations and the results depicts that it saves 99.3\% time, reduces 23.8\% energy consumption, and satisfy QoS. Greedy algorithm is another kind of heuristic that takes the best immediate or local solution of an optimization problem and subsequently improves it to find a global optimum. In the literature, \cite{sun2017cost} follows greedy algorithm as heuristic for computational resource allocation. The simulation results show that it provides 22\% better near-optimal solution. Furthermore, \cite{chen2016smart} uses max-heap as a heuristic for storage allocation in multi-agent cloud robotics and minimizes the cost of servers. Similarly, \cite{el2016environment} offers fuzzy logic-based heuristic for bandwidth allocation in multi-agent cloud robotics. Although heuristic approaches assist to find the optimal or near optimal solution in faster way, they are unable to provide the best optimal solution. Hence, it is only suitable in approximating the exact solution.
\subsubsection{Evolutionary computing} It is one of the popular methods for solving NP-hard optimization problem which is also applicable to multi-agent cloud robotics. For example, in an autonomous oil factory maintenance application, to support the on-demand mobility and the path planning of the robots as well as the selection of access points for sending data to cloud simultaneously become complicated. This joint optimization problem can be easily solved with the help of evolutionary computing. In literature, a genetic algorithm (GA) based strategy is formulated for product transportation in cloud-based manufacturing that optimizes the computational load of robots, overall cost and processing time concurrently \cite{li2017subtask}. The simulation results depict that this strategy reduces computation time and cost up to 10\%. Another GA based resource deployment method is used in \cite{xie2019loosely} that organizes multiple resources for exchanging messages and executes robotic applications meeting their QoS. This method is evaluated by both simulations and experiments on real testbed. The results show that the solution is able to satisfy the QoS requirements of 98\% tasks. As GA based approaches are based on random initial population, they often fail to reach the most optimal solution satisfying all objectives and constraints. Therefore, non-dominated sorting genetic algorithm-II (NSGA-II) \cite{NSGAII} is adopted in \cite{afrin2019multi,afrin2018energy} for optimizing multiple resource allocation objectives simultaneously. The energy consumption is optimized by \cite{afrin2018energy} while executing the tasks of robotic applications within its deadline. According to simulation results, this solution reduces the task execution time by 15\% and the energy consumption of resources by 10\%. In \cite{afrin2019multi}, this solution is further improved by considering application deadline, energy consumption of resources and the user expenses concurrently. Simulations results demonstrate up to 18\% improvement in optimizing time, energy, and cost. Both solutions initially select better population for evolution so that the stopping criteria is met in quicker time. Basically, evolutionary computing methods always follow a population-based trial and error approach to find a solution, which sometimes becomes time consuming and infeasible for real-time applications. 
\subsubsection{Game theory} This approach assists in finding an optimal outcome from a set of choices by analysing the costs and benefits of each independent participants within a system, especially in a competitive scenario \cite{wei2010game}. In multi-agent cloud robotics, sometimes it may happen that multiple applications compete to use the same set of available resources. For example, UAVs are required in a smart farm for crop health monitoring, pesticide control as well as livestock management applications which arises a competition among the applications to occupy the resources. Therefore, to solve this kind of issue, some works in the literature apply game theoretic approaches. For example, a bandwidth allocation strategy using Stackelberg game is described in \cite{wang2012game} and the performance of the strategy is evaluated through simulations. In addition, \cite{wang2016pricing} allocates bandwidth in multi-agent cloud robotics with a task-oriented pricing mechanism. Experimental results accumulated from testbed and simulations show that the time and cost increase with number of network nodes. For bandwidth allocation, auction mechanism is also investigated in \cite{HierarchicalAuction}. Both simulation and real testbed results depict that the mechanism is efficient to increase bandwidth usage by 35.9\%. K-means clustering, and auction-based game models are used in \cite{lwowski2017task} to assign robots and allocate resources for search and rescue operations. The evaluation study is conducted by means of simulations as well as testbed, and the results illustrate that the speed and scalability of the system increases. Generally, by identifying either Nash equilibrium or the best response, or the pareto efficiency, game theory approaches solve complex problems. However, these approaches assume that the agents or players make rational decisions at all the times, which is not obvious in many cases. In multi-agent cloud robotic systems, it is not always possible to interact with the players consistently as the communication is a constraint in such environment. Considering the system environments and application types, it is also challenging to define and solve a game theoretic model of resource allocation problem in real-time.
\subsubsection{Deep learning} The formal optimization and approximation method including heuristics, meta-heuristics such as evolutionary approaches are computationally expensive to be run on local robots or edge resources. In addition, game theoretic approaches are not always suitable due to the heterogeneity of resources and the communication overhead. To overcome these issues, machine learning (ML) based approaches are also explored in the literature. When the optimization problem becomes complex and the exploitation of contextual information is important to make resource allocation decisions, ML approach with deep learning (DL) can be a viable solution. However, for training and testing, a large amount of data is required in DL approach which is not always available in a dynamic environment with limited storage. Therefore, as a combination of ML and DL, deep reinforcement learning is becoming a promising technique for multi-agent cloud robotics to allocate resources for the robotic tasks with discrete and continuous state and action space. For example, in a smart factory, sometimes robots may require computation support, sometimes may require data offloading support and in some cases may require storage support. Depending on the dynamic environment, heterogeneous types of services are requested in the system. To deal with this uncertain demand, learning based resource application mechanism will be an attainable solution. Reinforcement learning based mechanism for computational resource allocation is used in \cite{liu2018reinforcement} that optimizes the energy consumption of resources and profit of service providers. The simulation results illustrate that RL scheme performs better than GA under the condition of limited cloud computing resources. Nevertheless, this approach becomes infeasible when the reward function or the objective function is ambiguously defined without assessing the system environment and other constraints.
\par In recent years multi-agent cloud robotics has become one of the major fields of interest for both academia and industries. Table \ref{Tab:summary_reosurce_allocation} highlights a brief summary of the reviewed papers in respect of resource allocation. Although many important aspects are identified in the existing literature, there exist some other issues that are required to be addressed for further improvement in this domain. Since multi-agent cloud robotic systems are comprised of heterogeneous devices with different capabilities, the allocation of resources needs to be efficient enough to satisfy the tasks requirements. For mission critical application such as fire-fighting or robotic surgery, low latency communication and reliable connectivity is required. Thus, reliable, and delay-aware resource allocation technique is necessary while selecting allocation mechanism. Since the resources are energy constrained, energy-efficient resource allocation policies are essential to meet the QoS in dynamic network environments. Additionally, the performance of most of the resource allocation policies are evaluated through simulations. Only a few works perform experiments on real-world testbeds. Moreover, it is evident that real-world robotic applications possess an inherent need for faster processing. Optimal allocation of resources and efficient service provisioning for the robotic systems will boost the successful execution of these applications. In the following section, different aspects of service provisioning such as appropriate resource pooling, proper task selection for offloading and scheduling the offloaded tasks are discussed in detail.
\begin{figure*}[!t]
\centering
\includegraphics[width=5in]{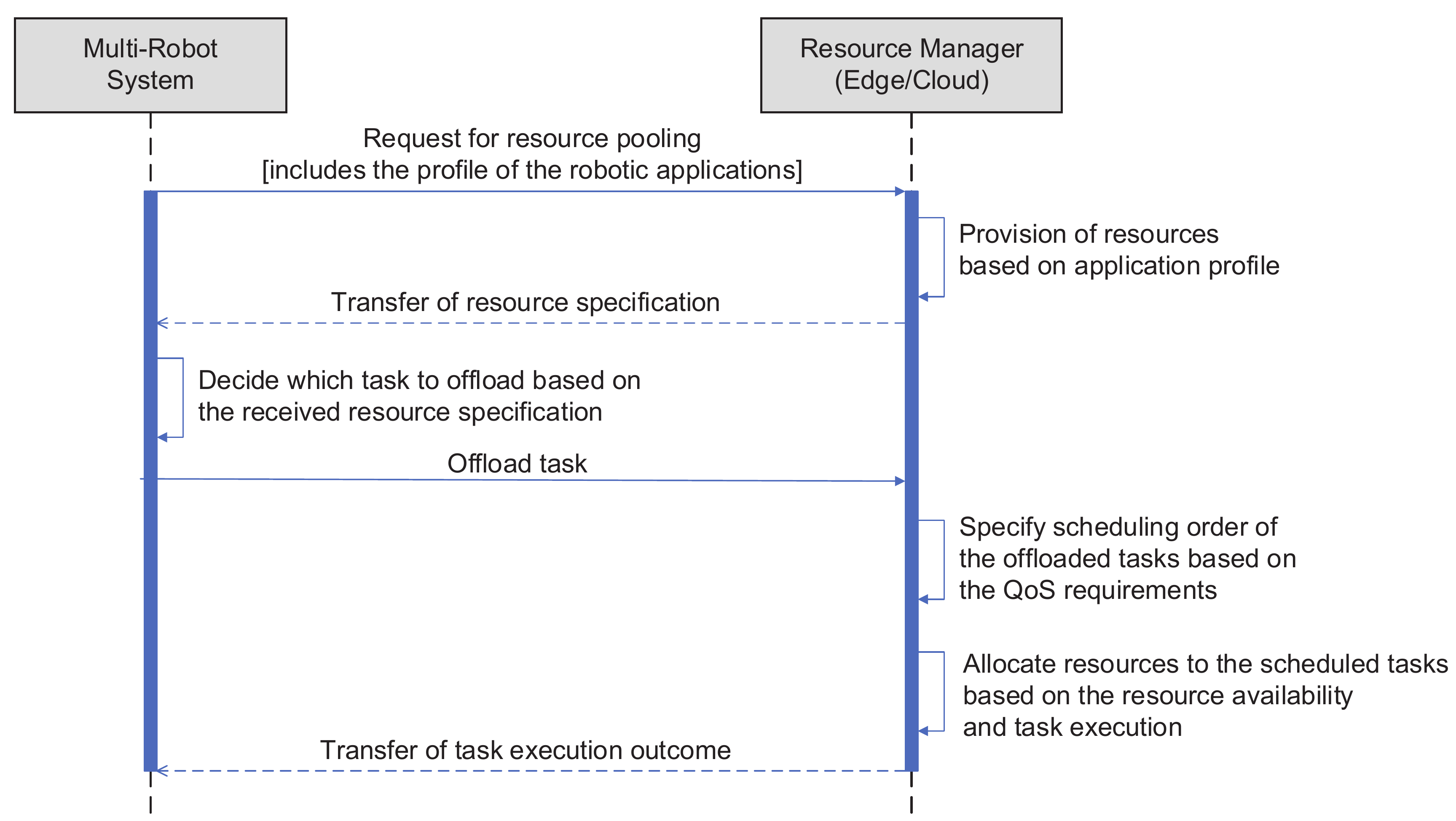} 
\caption{Interaction between multi-robot system and resource manager during resource allocation and service provisioning.}
\label{fig:interaction_resource_allocation_service_provisioning}
\end{figure*}
\section{Service Provisioning in Multi-Agent Cloud Robotic Systems} \label{service_provisioning}
To efficiently allocate resources for the tasks of robotic applications, a multi-robotic system and the resource manager of edge and cloud environments work collaboratively as shown in Fig. \ref{fig:interaction_resource_allocation_service_provisioning}. This process initiates from the multi-robotic system by requesting the resource manager to discover resources for the robotic applications. While making such requests, the multi-robotic system also forwards the application profile such as their resource requirements and dependencies to the resource manager. Based on the application profile, the resource manager provisions resources for the multi-robotic system. Resource pooling includes the activation or reservation of the resources and the creation of resource pool. However, after provisioning, the resource specifications and configurations are forwarded to the multi-robotic system. Based on the resource specification, the multi-robotic system decides which task of the robotic application is required to offload. A resource manager can receive a huge number of offloaded tasks from different multi-robotic systems. After receiving the offloaded tasks, the resource manager defines the scheduling order of the tasks based on their QoS requirement such as the deadline constraints. In parallel, the resource manager allocates resources to the tasks according to their availability. Depending on such interactions between multi-robot system and resource manager, the pooling of resources and the offloading and scheduling of the tasks are regarded as the pre-processing operations for the resource allocation. In this work, these pre-processing operations for resource allocation are collectively termed as service provisioning.
\subsection{Resource Pooling} Resource pooling defines the arrangement of the resources for task processing and storing data to provide services to the users. For realizing resource pooling, a service provider at first needs to locate and retrieve the existing resources across multiple infrastructure domains in multi-agent cloud robotics, which is commonly known as resource discovery. After discovering, resource scaling through aggregation, amalgamation and shifting is required to make the pool of resources compatible for hosting robotic applications. However, for creating any resource pool, two major research questions are required to be addressed. They are discussed below.
\begin{itemize}
\item{\textit{How secured the resources are?:} While selecting resources from local, edge and cloud infrastructure to create resource pool, the security of the resources needs to be ensured. The security issues can be classified with multiple aspects. Firstly, it is essential to protect the physical resources from malicious attacks. While sharing data with other robots, edge resources and cloud data centre, unauthorized access to data needs to be prevented by strong integrity and confidentiality protection \cite{jain2020ecc}. It is necessary to ensure that the robots only interact with legitimate servers and reliable devices. Reversely, with the help of authentication scheme edge or cloud data centre will only establish communication link with authorized robots and reject illegal communication requests. Privacy preservation is another fundamental requirement for data transmission among robots, robot-to-edge/cloud, and edge-to-cloud. Robust encryption mechanism is a viable solution to maintain data privacy. In multi-agent cloud robotics, for service provisioning, wireless access services from heterogeneous networks from multiple service providers may be required. Consequently, communication and transmission process become riskier for their inherent security vulnerabilities. Therefore, while creating the resource pool, trust-worthy resource selection is mandatory for efficient service provisioning. In addition, efficient security measures are required to prevent vulnerabilities both at resource level and communication level. Different mechanisms such as trust establishment, reputation-based trust, trust measurement techniques\cite{noor2015cloudarmor,zhu2018trust,Robotic_Cooperation,survey_cloudRObotics_architecture,kubiatowicz2018secure}  have been well studied in literature to provide trust-worthy resources. Recently, blockchain technology has become popular to address the security and privacy concerns in multi-agent cloud robotics \cite{Blockchain_FRUCT,zheng2018blockchain,yang2019integrated,lopes2018overview,khan2019blockchain}.} 
\item{\textit{How efficiently resources can interact?:} While creating the resource pool  in multi-agent cloud robotics, the selection of communication networks and protocols play an important role for efficient data exchange among resources and to maintain the QoS of the applications. Existing networking protocols have been widely used for wired or wireless communications in multi-agent cloud robotic systems \cite{survey_cloudRObotics_architecture}. Nevertheless, the efficiency of such communication depends on the working environment and application scenario. Consequently, depending on the contexts, researchers have adopted miscellaneous communication protocols to reduce data transmission delay and offer better QoS. Robots can communicate among themselves and with cloud using middleware like ROS\cite{ROS,crick2017rosbridge,bozcuouglu2018exchange,benavidez2015cloud,toffetti2017cloud,
PPAAS,beksi_object_recognition,miratabzadeh2016cloud,wen2016towards,hong2018cloud,
demarinis2019scanning}.
However, this type of middleware cannot always support secured communication. Another key challenge for communication in a dynamic environment is to optimize route discovery and route maintenance with minimum computation time and resource requirement. Gossip protocol is recommended in \cite{survey_cloudRObotics_architecture} for robot-to-robot and robot-to-cloud communications that are particularly suited for mobile robots. Data stream network (DSN) and controller area network (CAN) are adopted for real-time communication in \cite{beigi_smartcity_cloudRobotics} and \cite{vick2015robot}, respectively. Micro Internet protocol (uIP) is adopted in \cite{turnbull2013cloud} to support simplified communication between robot and cloud for data transmission. TCP or IP socket based communication also benefits robots for faster communicate with cloud \cite{tian2017cloud,vick2016model}. However, for delay-sensitive applications, it is always time consuming to send data to cloud rather than communicating with edge resources. Using short-range wireless communications technologies such as Zigbee, Bluetooth or Wi-Fi direct, robots get services from the edge resources \cite{li2016toward,antevski2018enhancing}. A heartbeat protocol is implemented in \cite{tian2019fog} to deal with network latency of multi-agent cloud robotics. Network bandwidth usage can be substantially preserved by processing the data on edge resources \cite{gudi2018fog,tanwani2019fog}. Packet delivery failure and communication outage are inherent drawbacks of wireless communications \cite{survey_cloudRObotics_architecture}. Therefore, backup mechanisms are required, and the system needs to be robust enough to recover from unwanted events.}
\end{itemize} 
\subsubsection{Resource Discovery}
Resource discovery indicates the arrangement of the resources for task processing and storing data. The amount of resource required for a robotic system is estimated according to the resource availability and the characteristics of the system in multi-agent cloud robotics. Service providers apply different resource discovery policies to meet the dynamic and static demands of robotic systems. Depending on the characteristics of the demands two basic type of resource arrangements such as on-demand and reserved are made to find the available resource pool from the cloud, edge, or local resources. 
\begin{itemize}
\item{{\em On-demand arrangement}:} On-demand resource discovery allows the robotic systems to access available resources on-the-fly when they are needed. This arrangement is selected for the execution of robotic applications where the workload varies or changes very frequently. To deal with the dynamic workload, on-demand approach is exploited in \cite{chen2016smart,sun2017cost,liu2018reinforcement,pandey2015dynamic,el2016environment,HierarchicalAuction}.  Moreover, a robotic system is billed on pay-per-use basis to utilize edge or cloud resources by the service provider. The idle or unused resources can also be provisioned for the robotic systems as spot-instances. The price of these instances changes periodically, depending on the supply and demand of the spot instances among multiple robotic systems. To get access to the spot-instances, a particular system is required to bid the current price. When multiple robotic systems target for the same service or the same resources from cloud data centre or edge server, spot-instances help a particular system to execute their tasks immediately. However, spot-instance discovery is yet to be explored for multi-agent cloud robotic systems. 
\item{{\em Reserved arrangement}:} Robotic systems arrange resources prior to executing any application rather than sending resource request on-demand to the service provider. The robotic systems harness the resources for a fixed contract period with a static price. In this stable resource arrangement, robotic system negotiates with the service provider for a particular service and the provider provisions resources in advance for that service. Basically, reserved provisioning is applicable for the robotic applications that have predictable and unchanging workload. In literature, considering the static workload, \cite{afrin2019multi,afrin2018energy,chen2018qos,lwowski2017task,li2018latency,li2017subtask,wang2012game,wang2016pricing} prefer reserved arrangements of resources.
\end{itemize}
There are pros and cons in each resource discovery mechanism. Although on-demand arrangement helps instant access to the cloud resources in dynamic environment, the cost of resources with this plan is higher than the others. Resource usage expenses get reduced through reservation plan. This plan sometimes causes under and over provisioning problem due to the uncertainty of workloads. Conversely, the spot-instance plan can deal with the unpredictable workload of multiple systems and set the resource price as per the intensity of competition. Because of the diversified features of resources in multi-agent cloud robotics, it is important for robotic systems to select the best solution for resource discovery according to the constraints and expectations to make the system cost efficient. Additionally, efficient resource discovery helps in enhancing the competency of edge or cloud based robotic services. 
\subsubsection{Resource Scaling}In multi-agent cloud robotics, the scattered and isolated geo-distribution of discovered resources makes the creation of resource pool challenging. In this case, resource scaling is required to enhance the suitability of the resources in provisioning the services. Different techniques for resource scaling including resource aggregation, resource amalgamation, and resource shift are applicable in multi-agent cloud robotics.
\begin{itemize}
\item{{\em Resource aggregation}:} It means combining the same type of resources to create resource pool with increased power \cite{katz2014mobile}. It is a commonly used technique to prepare resource for service provisioning in multi-agent cloud robotics. For example, memory from robot, edge and cloud storage can be aggregated together to store large amount of data while dealing with big data in CPSs. Similarly, for computation intensive tasks, CPUs from multiple resources are augmented to create a more powerful processor. However, resource aggregation only supports the same type of resource augmentation, which is not always capable of meeting the resource demand.
\item{{\em Resource amalgamation}:} It aims at combining individual capabilities of resources to offer greater dimension of resources \cite{katz2014mobile}. For example, while monitoring a field, the images from the on-board cameras of different robots can be combined to create an image with higher resolution and using edge or cloud resources with higher processing capabilities 3-D effects can be added to get more clearer view. During resource amalgamation, as multiple devices participate in sharing resources, the management of their heterogeneity and cross platform operations adds further overhead to the system.
\item{{\em Resource shift}:} It refers to moving resources from one device to another device \cite{katz2014mobile}, e.g., migrating VM from one cloud server to another server for executing robotic applications. Even the connectivity can also be shared by making additional access links available to any poorly connected robot or edge resource. However, while shifting resources, the communication time, and the compatibility of the destination device to host the robotics applications, also need to be taken into account.
\end{itemize} 
\begin{table*}[!htb]
\centering
\caption{Comparison among computation offloading schemes in multi-agent cloud robotics}\label{Tab:summary_of_offloading_techniques}
\scriptsize
\begin{tabular}{|p{2.7cm}|p{1.2cm}|p{4cm}|p{4cm}|p{4cm}|}
\hline
Criteria & Scheme & Pros & Cons & Open Issues \\ \hline
Time of decision making & Re-active & $\bullet$ Offers urgent action \newline $\bullet$ Deal with uncertain events & $\bullet$ Effective solution for shorter period \newline $\bullet$ Stringent deadline to make decision& $\bullet$ Balance between pro-active and re-active decision making \newline $\bullet$ Selection of appropriate algorithm \\
\cline{2-4}
& Pro-active & $\bullet$ Provides long term solution \newline $\bullet$ Supports predictive analytics & $\bullet$ Computation overhead \newline $\bullet$ Requires validation prior to setting into practice & \\\hline
Offloading type & Partial & $\bullet$ Supports parallel processing with load distribution \newline $\bullet$ Reduced waiting time & $\bullet$ Higher data dependency \newline $\bullet$ Time consuming for task partition and result aggregation
 & $\bullet$ Trade-off between computation and communication \newline $\bullet$ Energy-delay optimization
 \\
\cline{2-4}
& Full & $\bullet$ Easier decision-making\newline $\bullet$ Less operational latency
 & $\bullet$ Higher bandwidth requirement \newline $\bullet$ Burden on single resource&   \\\hline
Number of decision maker & Single & $\bullet$ Less communication overhead
 \newline $\bullet$ Simple decision making & $\bullet$ Lack of system overview
 \newline $\bullet$ Vulnerability on single point of failure & $\bullet$ Optimal selection of robots \newline $\bullet$ Synchronized decision making \\
\cline{2-4}
& Multiple & $\bullet$ Prolong network lifetime \newline $\bullet$ Better energy optimization& $\bullet$ Synchronization of decisions \newline $\bullet$ Security vulnerability & \\\hline
\end{tabular}
\end{table*}
\par To summarize, efficient resource pooling is a basic requirement in multi-agent cloud robotics. The security, accessibility, availability, and compatibility of the resources need to be assessed while creating the resource pool as it sets the foundation for further service provisioning operation including computation offloading and task scheduling.
\subsection{Computation Offloading}
As local robots in multi-agent cloud robotic systems have limitations in computational power, storage, and energy, it is often required to move the compute-intensive tasks to resource enriched cloud or edge instances for successful robotic application execution. The idea of computation offloading from local robots to cloud resources is derived from mobile cloud computing \cite{cuervo2010maui}. However, the service orchestration, application requirements and resource orientation differ from mobile cloud computing to multi-agent cloud robotics. More specifically, the key distinction between two paradigms is the robot's unique ability to move on-demand, which allows them to actively access better communication links for offloading. Therefore, it is infeasible to directly apply the mobile cloud-based offloading techniques to a multi-agent cloud robotic system. Additionally, a robot determines its position in respect of other robots which facilitates in local offloading as well as edge or cloud-based offloading. In multi-agent cloud robotic system, local robot itself can actively participate as a resource provider to support compute-intensive tasks execution offloaded by other robots. Thus, it is important to investigate the role of mobility and communication while offloading a task in multi-agent cloud robotics. On the other hand, it is difficult to make optimized offloading decisions due to the delay constraint, mobility of robots, and additional data transfer and computation cost. A trade-off among these parameters ensure the improved performance of robotic systems. For efficient computation offloading in multi-agent cloud robotics, two important questions need to be addressed. They are discussed below. 
\subsubsection{How offloading decision should be made?}
The offloading performance largely depends on how the offloading decision is made. This decision can be made either reactive or proactive, based on the time when the decision is made. 
\begin{itemize}
\item{{\em Re-active scheme}}: In this case, the offloading decision is taken after a situation has arisen where offloading is necessary. In such a situation, a quick or immediate decision making is required based on system demand. 
\item {{\em Pro-active scheme}}: This is a predictive scheme which anticipates computation requirements and takes a rational decision before any event has occurred. Using predictive analytics, the system context, tasks, and resource requirements are envisaged before actual offloading request is generated.
\end{itemize}
\par With re-active decision making, the deadline to make the decision can be very stringent, which can obstruct to make an effective decision. On the other hand, despite the consumption of additional time for decision making, with the help of predictive analytics, pro-active offloading gives a long-term solution to the system at the cost of additional computation overhead. Therefore, a balance between pro-active and re-active decision making is necessary. Selection of appropriate algorithm to implement this decision making is required for better performance.
\subsubsection{How actual offloading should be conducted?}
The performance of offloading techniques is also affected by the number of offloaded tasks. It is the responsibility of robots to conduct either full or partial offloading based on the assessment of network profile (bandwidth, access point), device profile (battery status, local storage), and system objectives (minimize cost, distance, time, energy).
\begin{itemize}
\item{{\em Partial offloading}:} In this offloading approach, rather than forwarding the whole application, a part of the application is transferred from local robot to remote instances for execution. The remaining parts are either managed by local robots or sent to other computing infrastructure like edge.  
\item{{\em Full offloading}:} This is a conventional approach for offloading in multi-agent cloud robotics, where the application is transferred completely to the remote cloud for execution and after execution the results are sent back to local robots. It increases the burden on communication link by sending a large volume of data along with the application to the remote destination.     
\end{itemize} 
\par Partial offloading supports parallel processing and consequently allows load distribution among the resources. The waiting time to process the data on a single resource is also reduced. However, data dependency among the resources becomes higher as tasks are partitioned among multiple resources. It consumes additional time for task partition and result aggregation from different resources. On the contrary, in full offloading, since the entire computation is offloaded, the burden of task partitioning diminishes and the decision making becomes easier. Nevertheless, it incurs higher bandwidth to send the full task and consumes higher energy of single resource. Therefore, a trade-off exists between computation and communication in offloading decision making. In addition, energy-delay optimization is also required for making efficient offloading decision.
\par Another crucial criterion of offloading is to distribute the offloaded tasks among the robots. It basically depends on the available local resources (robots) and their orientation. Regarding this feature, the offloading decision-making factors are classified as follows: 
\begin{itemize}
\item{{\em Single robot}:} For a single robot, the offloading decision depends on the location and the appropriate offloading approach that ensures optimal transfer of information between robot and the computing instances.
\item{{\em Multiple robot}:} In this approach, multiple robots work collaboratively to complete the offloading operation. It requires to make a balance between robot-to-robot coordination and robot-to-cloud or edge communication.
\end{itemize}
\begin{table*}[!htb]
\centering 
\caption{A Summary of works concentrating on computation offloading in multi-agent cloud robotics}\label{Tab:summary_of_offloading} 
\scriptsize
\begin{tabular}{|p{0.7 cm}|p{1cm}|p{.9cm}|p{2cm}|p{2cm}|p{1.8cm}|p{4cm}|p{2.4cm}|}
\hline
\textbf{Work} &  \textbf{Offloading Type}& \textbf{Decision Maker} & \textbf{Methodology} & \textbf{Objective} &  \textbf{Evaluation Method} &  \textbf{Results}&  \textbf{Use Case} \\\hline  
\cite{xu2016multi}&Partial&Single & Multi-level decision using heuristic & Minimize time and energy& Testbed& Saves 0.052s and 19.63J than local execution. &Object recognition and motion control.\\\hline
\cite{PPAAS}&Full& Multiple& Greedy approach & Minimize path cost&Testbed&Find the shortest path faster than local robot.&Path planning.\\\hline 
\cite{beksi_object_recognition}&Partial&Multiple& Cloud based task execution  & Minimize network latency&Prototype& Only 1\% false positive result.&Object recognition and grasping. \\\hline
\cite{berenson_robot-path-planning}&Partial& Multiple&  Path planning using heuristic & Minimize time&Simulation&Offers 90\% better result.&Path planning in smart home and healthcare. \\\hline
\cite{salmeron2015tradeoff}&Full& Single&  Dynamic parallel algorithm &	Minimize communication cost&Prototype&Performs better than only robotic execution.& Vision based navigation assistant. \\\hline
\cite{guo2018energy} &Partial/Full& Single & Genetic algorithm & Balance energy and network lifetime&Theoretical analysis and simulation & Lifetime of the network is prolonged&Generalized robotic applications. \\\hline
\cite{wan_Context-aware-cloud-robotics}&Full&Multiple & Context-aware greedy approach & Minimize energy and cost&Simulation.&Context awareness performs better in material handling &Smart factory. \\\hline 
\cite{Data_Retrieval}&Full&Multiple& Market-based management strategy &Minimize time&Testbed, Simulation&Significant improvement in bandwidth usage and load balancing.& Generalized real-time robotic tasks. \\\hline 
\cite{hong2019qos}&Partial&Multiple&QoS-aware game theory &Minimize latency and energy&Simulation&Stable performance gain with increasing number of robots.& Generalized robotic applications. \\\hline
\cite{Ak_TII}&Partial/Full&Single&Genetic algorithm & Minimize energy, time and distance&Simulation&Communication and mobility-aware offloading provide better performance.&Smart factory. \\\hline   
\cite{rahman2019energy}&Full&Multiple&Genetic algorithm & Minimize overall system energy&Simulation&Reduce energy consumption.&Smart factory. \\\hline
\cite{chinchali2019network}&Full&Multiple&Deep RL based offloading &	Minimize communication cost&Simulation&Supports offloading for mobile robots.&Generalized robotic applications. \\\hline
\cite{li2019partial}&Partial&Multiple&Hierarchical iterative approach& Minimize energy, computation time&Theoretical analysis and simulation&Effectively reduce the energy consumption.&SLAM. \\\hline 
\end{tabular}  
\end{table*} 
\par Using a single robot, the offloading decision becomes simple and requires less communication overhead. However, for a single robot, it is not always possible to get the complete overview of the system and make efficient decision to maintain the QoS. In addition, the failure of a single robot can affect the full system. On the other side, offloading decision with multiple robots offers prolonged network lifetime and better energy optimization. Yet, the synchronization of the decisions from multiple robots becomes challenging and the system becomes vulnerable to security issues due to higher exposure of data. Therefore, optimal selection of robots is essential in making offloading decision.
\par To sum up, the main challenge in computation offloading is to trade-off between the local and the remote computation and communication cost in multi-agent cloud robotics. It is also challenging to efficiently offload a task and receive the response, since the robots can move on-demand to multiple locations during the application execution period. Since the offloading performance depends on the network resources, it is required for the robots to make a balance between mobility and bandwidth utilization. Finally, the identification of an appropriate approach to perform the offloading based on the arbitrary system context requires further research. The pros and cons of different offloading schemes are summarized in Table \ref{Tab:summary_of_offloading_techniques}.
%
%
\par In the context of multi-agent cloud robotics, computation offloading is considered as a trending topic. Therefore, over the last few years, plenteous research has been conducted in this area. Initial studies such as \cite{DAvinCi,osunmakinde2014development,bogue2017cloud,Rapyuta} give particular attention to the infrastructure and framework for offloading in multi-agent cloud robotics. Platforms to support offloading the compute-intensive tasks from the robotic network to cloud and edge resources are also furnished in \cite{beksi_object_recognition,Cloudroid}. Subsequently, other studies including \cite{li2017subtask,song2017scheduling} exploit the optimal approach for efficient task offloading between local robot and remote computing instances. Additionally, task partitioning has been studied and established in \cite{li2017multi,lomenie2004generic}, where the applications are decomposed in tasks and distributed among the robots and cloud resources. Furthermore, based on each decision-making criterion, the offloading performance can vary significantly. To deal with such cases, different offloading approaches according to the resource architecture, application profile and performance parameters are also accentuated in the literature. Table \ref{Tab:summary_of_offloading} presents the summary of some notable works in computation offloading for multi-agent cloud robotics that refer the aforementioned attributes in detail. As seen in Table \ref{Tab:summary_of_offloading}, most of the existing approaches perform full application offloading to cloud, whereas very few including \cite{beksi_object_recognition,xu2016multi,hong2019qos,li2019partial} apply partial offloading. Another variation is seen in the number of engaged robots for offloading operations. For example, in \cite{Survey_RobotCloud}, the authors consider single robot methods.  
\par In addition, most of the offloading algorithms follow heuristic approach (\cite{chen2018qos,xu2016multi}) and evolutionary method (\cite{rahman2019energy,guo2018energy,Ak_TII}), whereas others adapt greedy algorithms (\cite{PPAAS,wan_Context-aware-cloud-robotics}), game theory (\cite{Data_Retrieval,chinchali2019network}) and dynamic programming (\cite{salmeron2015tradeoff,berenson2012robot}) to make offloading decisions. There also exist differences among the existing works in selecting the offloading objectives. Minimization of energy (\cite{rahman2019energy,pandey2015dynamic,xu2016multi,li2019partial}), latency (\cite{chen2018qos,Data_Retrieval}), distance or movement cost (\cite{PPAAS,Ak_TII}) and  bandwidth or communication cost (\cite{beksi_object_recognition,salmeron2015tradeoff,chinchali2019network}) are considered as the potential offloading objectives.
\par With the help of offloading, the management of resources become distributed among the local robots, edge and cloud. Based on the features of offloaded tasks, resources are usually allocated from the available resource pool. If the tasks to be offloaded are selected appropriately by the robots, task scheduling and service provisioning can be conducted efficiently at the edge or cloud. Nevertheless, if the resource of each cloud or edge server is limited, then offloading becomes inefficient. It turns into worst when all robots select the same server as the destination. Hence, a detailed study is required so that the offloading techniques in multi-agent cloud robotics can deal with such cases deliberately and improve the application QoS. 
\subsection{Task Scheduling}
Multi-agent cloud robotics enables the local robots to outsource their computing tasks and storage requirements to the edge or cloud and ensures enhanced capacities and higher performance for the robotic systems. One of the important research issues is to determine how resource providers can efficiently handle the overwhelming requests from different robotic systems when the amount of resource is limited. With the help of resource discovery and resource scaling, the resource pool is identified for the robotic systems while scheduling determines in which order the robotic tasks will arrive and leave the resources. A task scheduling approach becomes efficient when it avoids longer scheduling delay and waiting time for the requested services. It is also liable for arranging incoming service requests in a certain manner so that the available resources are properly utilized. To ensure the highest utilization of available resources, the tasks should be dispatched to the resources rationally. Usually, task scheduling is conducted after a fixed period or dynamically according to task arrival rate, availability of resources and workload on the resources.  
\subsubsection{Static scheduling} In this approach, tasks are scheduled after a specific time interval, where communication cost and computation cost of each task are estimated prior to its assignment on the resources. In the literature, \cite{afrin2019multi,afrin2018energy,chen2018qos,lwowski2017task,li2018latency,li2017subtask,wang2012game} follow this task scheduling strategy.
\subsubsection{Dynamic scheduling} According to this approach, task scheduling is conducted randomly without any prior time or cost estimation. In literature, \cite{chen2016smart,sun2017cost,liu2018reinforcement,pandey2015dynamic,wang2016pricing,el2016environment,HierarchicalAuction}  schedule the tasks dynamically.
\par The efficient resource utilization of a robotic system depends more on the scheduling and load balancing methodologies than the random allocation of resources. For instance, to share resources among cloud-aided industrial robots in a smart factory, dynamic scheduling of the tasks and computational load balancing are equally important \cite{wan2018artificial,Smart_Manufacturing}. In literature, very few works are concerned about the scheduling policies in multi-agent cloud robotics. Moreover, due to lack of holistic monitoring of the resources, scheduling techniques are tedious to implement in a multi-agent cloud robotic system. In future, appropriate selection of task scheduling strategies for successful service provisioning to the robotic systems should be explored extensively. In the following sections, the gaps of existing literature are explicitly identified from the lessons learned and multidisciplinary future research directions are discussed for further improvement of this domain.
\section{Lessons Learned}\label{lessons}
In this paper, multi-agent cloud robotics is reviewed from resource allocation and service provisioning perspectives and the lessons learned by this review are summarized in this section.
\par \textit{1)} It is clear that multi-agent cloud robotics enhance the performance of associate CPSs by enabling the execution of robotic applications in different infrastructure levels. These robotic applications are composed of various tasks. On the other hand, the computing resources within a multi-agent cloud robotic system are not equally competent to execute all types of tasks due to their heterogeneity. For example, the latency-sensitive tasks are preferable to execute at the robots or edge computing infrastructure whereas the compute and storage-intensive tasks require forwarding to cloud infrastructure. In such cases, the partitioning of robotic applications based on the characteristics of the tasks can be a viable solution, although it incurs additional management overhead in aggregating the outcome of the tasks at the receiver end.
\par \textit{2)} A significant amount of energy is required to run the robots and edge resources. Moreover, the arrangement of grid-based energy is costly when the robotic system works in remote places. To deal with such constraints different policies have already been developed to optimize the energy consumption in multi-agent cloud robotics. However, such reactive approaches are not feasible when the computing environment and corresponding resources depend on the renewable energy. Since the availability of renewable energy is subjected to the context of physical environment, the proactive resource allocation and service provisioning based on such uncertain parameter also becomes very complicated.  
\par \textit{3)} Mobility is one of the definitive feature of multi-agent cloud robotics. However, the mobility pattern of robots varies from one robot to another. For example, an UAV changes its location more frequently than a ground robot. In such cases, if the mobility-driven service provisioning requests are initiated from the robots, it will add a significant communication and data management overhead to the system which will be quite tedious to deal with at the run-time. A centralized approach to move the computation as per the location of robots can be an alternative solution. However, it requires a complete understanding of the network. Moreover, the monitoring of each robot in an individual manner can be challenging task when the mobility is uneven. Therefore, dynamic switching between these two solutions is highly preferable to ensure efficient mobility driven resource allocation and service provisioning.
\par \textit{4)} As noted, a multi-agent cloud robotic system is a combination of robotic systems, edge and cloud computing. Most likely, these infrastructures have their own resource manager that controls the internal functions. The co-existence of multiple resource manager incurs further synchronization problem. Through deliberate negotiation, this problem can be solved to a great extent. However, it limits the scope of making independent decisions for autonomic resource management. Therefore, a fair and adjustable distribution of resource management responsibilities in multi-agent cloud robotics is highly recommended.
\par \textit{5)} The computing infrastructures and execution platforms in multi-agent cloud robotics are highly exposed at different communication layers including edge-network layer, core-network layer and cloud communication layer. Consequently, it broadens the possibility of various security threats including data leakage and tempering. Since multi-agent cloud robotics deal with real-time use cases, heavy weight security features often slow down the interactions among different entities within the computing environment. Therefore, it is preferable to develop lightweight security features. However, it has a limitation in guaranteeing scalable solutions which is subjected to the extensive research.
\par \textit{6)} Only a handful of initiatives have been taken to realize the economic potential of multi agent cloud robotics. Since a universally accredited business model is yet to be developed, it is very difficult to make a comprehensive profit in multi agent cloud robotics satisfying the interest of all participating service providers. On different note, the rigid intention of profit enhancement sometimes urges to relax the important QoS parameters that degrades the trust between robotic system users and service providers significantly. Therefore, in most of cases, QoS requirements are given higher priority while provisioning services and allocating resources. It not only reduces the SLA violations but also resists the relinquish rate. Consequently, it improves the economic benefits of the service providers in multi-agent cloud robotics.
\par \textit{7)} There exist different frameworks to solve the resource allocation and service provisioning problem in multi-agent cloud robotics. Most of the frameworks conduct limited evaluation and hardly ensure low latency data-flow between robot and cloud. Furthermore, these frameworks are application-specific and their in-built software systems are not always adaptable to decentralized resource architecture and they often fail to deal with the real-world environmental constraints during robot to cloud interaction. Additionally, most of the existing works focus on computation offloading to the cloud as a part of the service provisioning. The edge resources are not set as the destination for offloading the computation. Besides, the importance of resource provisioning plan and determination of task scheduling period are barely investigated in these works.
\par The aforementioned lessons learned from the literature review helps in identifying the research gaps in multi-agent cloud robotics. In the following section a holistic framework for resource allocation and service provisioning along with some future research directions are provided that can address these research gaps to a great extent. 
\begin{figure*}[t]
\centering
\includegraphics[width=5in]{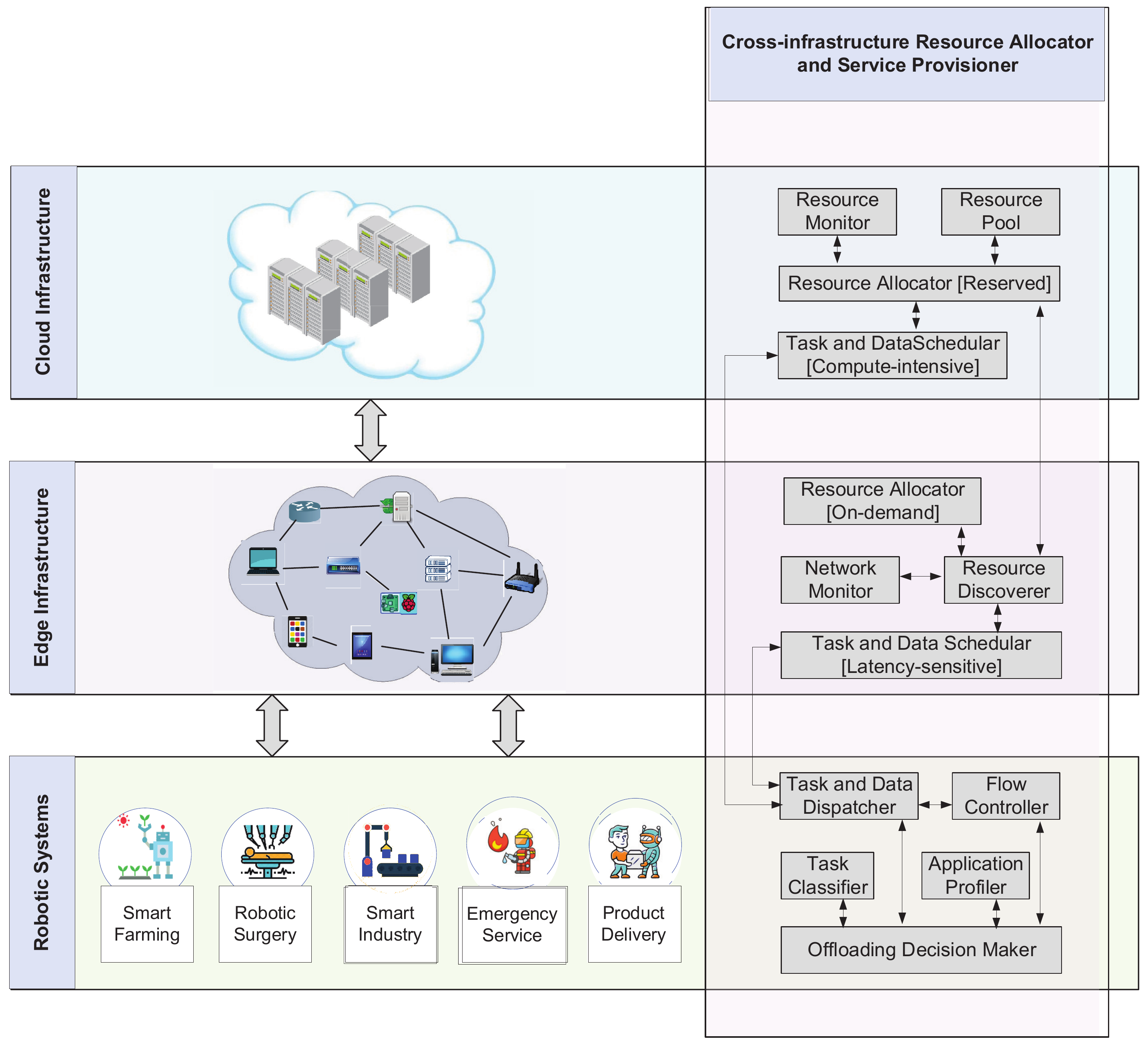} 
\caption{A holistic framework for multi-agent cloud robotics.}
\label{fig_framework}
\end{figure*}
\section{A Holistic Framework and Future Research Directions}\label{future}
A holistic framework encapsulates different hardware and software components that simplify the end-to-end interaction among the associated entities and facilitate the integration of various resource and service management policies in a scalable manner. Fig. \ref{fig_framework} depicts such a framework for resource allocation and service provisioning in a multi-agent cloud robotic system. As already mentioned, in this environment, the computing platform for robotic applications is extended to multiple infrastructure levels. This framework exploits the edge infrastructure for latency-sensitive tasks as the edge computing nodes are in the proximity of robotic systems and their dynamic augmentation for on-demand resource allocation is quite easier through ad-hoc networking. Moreover, the framework prefers to schedule compute-intensive tasks to cloud infrastructure as the datacentres host powerful computing servers and the resources can be reserved for a certain period. However, the main entity of the proposed holistic framework is a Cross-infrastructure Resource Allocator and Service Provisioner (CRASP). CRASP is also composed of several components which are deployed in distributed manner across different infrastructure levels. CRASP is provided with a robust and reliable communication link that interconnects its components logically. The brief discussion of these components in different infrastructure level is discussed below.
\subsection{Robotic System Level} In this infrastructure level, the \textit{Offloading Decision Maker} of CRASP resides that analyses the possibility and determines the benefits of task offloading in multi-agent cloud robotics. To perform these operations, the \textit{Application Profiler} assist the Offloading Decision Maker by providing meta-data regarding the application architecture and programming model and the \textit{Task Classifier} checks whether that task is latency-sensitive or compute-intensive. However, prior to start task offloading, the \textit{Dispatcher and Negotiator} connects the edge and cloud infrastructure and perceives the state of the computing platform. Later, based on the offloading decisions, the Dispatcher and Negotiator forwards the tasks and data to the corresponding infrastructure. Moreover, while the data are processing remotely, the Dispatcher and Negotiator also guides the \textit{Flow Controller} to tune the input data transmission rate as per the context of the processing destinations.   
\subsection{Edge Infrastructure Level} In this level, a specialized \textit{Scheduler} component of CRASP is placed that operates the execution of latency-sensitive tasks on edge computing nodes. Usually, these nodes interact with each other by forming ad-hoc clusters. The \textit{Resource Discoverer} helps to identify a suitable node from such cluster to execute a task. The \textit{Network Monitor} support this operation by updating the network status periodically. Based on the outcome of Resource Discoverer, the \textit{Resource Allocator} assign the task to the selected edge node. However, in most of the cases, latency-sensitive tasks originate from even-driven physical actions. Therefore, CRASP consider the allocation of resources at the edge infrastructure as an on-demand operation. Additionally, during uncertain scenarios such as node failure, power outage and resource shortage, the scheduled tasks to the edge infrastructure are required to be forwarded to the cloud. In such contexts, the Resource Discover directly communicates with the cloud-based resource allocator of CRASP and solve the issue.
\subsection{Cloud Infrastructure Level} In this level, CRASP focuses on scheduling compute-intensive tasks. Since the compute-intensive tasks are expected to have a longer period of execution time, the \textit{Resource Allocator} of CRASP at this level prefers to reserve the resources rather than dynamically provisioning them. However, to perform this operation, the Resource Allocator interacts with the \textit{Resource Pool} that contains the references of all computing resources within the cloud infrastructure. After the allocation, the Resource Allocator grasp the status of task execution time-to-time with the help of \textit{Resource Monitor}.   
\par However, there exist extensive research scopes to improve this framework which are discussed in the following subsections.
\subsubsection{Event-driven resource allocation and service provisioning} The execution of an application for a robotic system can trigger the execution of another application. For example, a robotic livestock monitoring application in a smart farm can trigger a robotic emergency management application. In this case, both the applications should be executed simultaneously. However, such arrangement often gets disrupted due to the resource constraints of multi-agent cloud robotics. Therefore, efficient resource allocation and service provisioning policies are required to deal with the event-driven requirements of robotic systems.
\subsubsection{Energy-efficient resource consolidation and scaling} Energy is one of the major concerns for any computing paradigms, especially when it is accumulated from renewable sources. Since the availability of renewable energy is subjected to the environmental contexts, the computing infrastructures are required to be adaptable to their sudden changes. In multi-agent cloud robotics, it can be attained by consolidating the resources when the supply of renewable energy is poor and scaling up the resources when the opposite happens. However, it is not such straightforward. Dynamic consolidation and scaling of resources in multi-agent cloud robotics alter the network topology and incurs additional resource management overhead. Therefore, the resource allocation and service provisioning policies require to observe these issues deliberately which demand extensive research.
\subsubsection{Balance between pro-active and re-active offloading decision}
In multi-agent cloud robotics, mobility of robots is one of the key factors that need to be considered for making the computation offloading decision. For static mobility pattern of the robots, pro-active offloading decision provides feasible solution as it helps anticipating system behaviour more accurately. However, in most of the cases, it is not obvious that robots will maintain a static pattern. To deal with such cases, re-active offloading decision making is required, despite of their effectiveness for shorter period. To deal with the dynamics of multi-agent cloud robotics, a balance between pro-active and re-active decision making is required so that with the help of predictive analytics the system can react in uncertain events with higher accuracy for a longer period.
\subsubsection{Allocation of network slices for 5G enabled multi-agent cloud robotics} 5G cellular communication has already created a significant buzz in both industry and academia. Unlike traditional cellular communications, the physical network in a 5G system is virtualized in multiple slices. These slices are used to transmit data traffic for different applications. The amount of network bandwidth to be allocated to each network slice depends on the priority of the applications. However, in multi-agent cloud robotics, the level of necessity for an application can change very frequently. For example, in a smart factory when an anomaly happens, the robotic application investigating the location of the fault runs in high priority. Soon after identifying the fault location, the emergency management application gets the higher priority of execution. Therefore, the resource allocation and service provisioning policies for multi-agent cloud robotics should be intelligent to make and tune prioritization of applications in run-time. Since such prioritization depends multiple parameters including the QoS requirements of the applications and user expectations, the policies require detailed exploitation of these parameters to ensure the air distribution of bandwidth on network slices. In future, this concept can be extended for 5G and beyond environment to support mission critical applications using Tactile Internet.
\subsubsection{Mobility-as-a-Service} In a geographically large-scale robotic system, the intensity of network connectivity is not uniform at all the locations. For example, in a smart farm, the robots working far from the access points receive poor network signal strength compared to others. At the same time, the inconsiderate deployment of access points will increase the mutual interference. Therefore, it is required to orchestrate the access points on ad-hoc basis. To resolve this issue, the mobility of robots can also be used as a service. The robots especially the UAVs can act as portable access points or signal booster for the robots receiving poor networking signal. However, the existing works in the literature consider mobility as a constraint of the multi-agent cloud robotics and to enable the mobility-as-a-service in this domain, significant research effort is required. 
\subsubsection{Real-time resource augmentation} As noted, multi-agent cloud robotics incorporate resources from different service providers. Since there exists a black box interface between the resource management policies of these service providers, a significant number of administrative operations is required to perform during resource provisioning in multi-agent cloud robotics. These operations are often time consuming that resist the real-time interactions among the corresponding entities. Interoperable resource allocation and service provisioning policies can solve this issue to a great extent. However, such policies need to observe the individual interests of each providers which is subjected to detail research. 
\subsubsection{Pricing model for resource consumption} Since there exists no accredited business model for the consuming the resources in multi-agent cloud robotic systems, it is very difficult to boost the revenue of service providers and facilitate the incentives for the users. Additionally, it obstructs the scope of providing compensation for SLA violations. Therefore, a detailed pricing model for multi-agent cloud robotic systems is required. However, such pricing model is difficult to developed as the computing components within a multi-agent cloud robotic system are highly heterogeneous and their operational cost in per unit time very significantly. Extensive research towards this direction can be a significant contribution to the existing literature. 
\subsubsection{{Lightweight security measures during data management}} There are some robotic use cases where sensitive data are exchanged. For example, in robotic medical assistance, the electronic health report contains a significant amount of private information. Sometimes this information requires to be accessed by different professional and organization bodies including insurance, pharmacist, and doctors. In such cases, an easy access to data is necessary for making real-time decisions. At the same time, the data access should be made secured so that unauthorized modification can be prevented. However, existing security measures such as blockchain and 128-bit asymmetric key cryptography are highly computation-intensive that slows down the real-time interactions to a great extent while identifying the authorized data access and modification \cite{yang2019integrated,lopes2018overview}. Therefore, lightweight security measures are needed for multi-agent robotic systems that not only ensure secured data management but also support real-time interactions.
%
\section{Conclusion}\label{conclusion}
We have presented a survey and the research outlook on resource allocation and service provisioning in multi-agent cloud robotics. The recent development and research on multi-agent cloud robotics both in academia and industry have been reviewed. As a prerequisite of efficient resource allocation and service provisioning, the concepts of resource pooling, computation offloading, and task scheduling have been discussed separately along with their challenges. In addition, the lessons learned from the survey have been summarized and a holistic framework for resource allocation and service provisioning in multi-agent cloud robotics has been presented along with several potential research directions. We believe that this comprehensive survey will serve as a useful reference and provide guidelines for further investigation and advancement of multi-agent cloud robotics.   
\section*{Acknowledgment}
This work is partially supported by Australian Research Council Discovery Project Grant DP190102828. The authors would also like to thank Data61, CSIRO, Australia for funding and supporting this work. In addition, the work of E. Hossain was supported by a Discovery Grant from the Natural Sciences and Engineering Research Council of Canada (NSERC).
%
\ifCLASSOPTIONcaptionsoff
  \newpage
\fi
\bibliographystyle{IEEEtran} 
\bibliography{references}

\begin{thebibliography}{100}
\providecommand{\url}[1]{#1}
\csname url@samestyle\endcsname
\providecommand{\newblock}{\relax}
\providecommand{\bibinfo}[2]{#2}
\providecommand{\BIBentrySTDinterwordspacing}{\spaceskip=0pt\relax}
\providecommand{\BIBentryALTinterwordstretchfactor}{4}
\providecommand{\BIBentryALTinterwordspacing}{\spaceskip=\fontdimen2\font plus
\BIBentryALTinterwordstretchfactor\fontdimen3\font minus
  \fontdimen4\font\relax}
\providecommand{\BIBforeignlanguage}[2]{{%
\expandafter\ifx\csname l@#1\endcsname\relax
\typeout{** WARNING: IEEEtran.bst: No hyphenation pattern has been}%
\typeout{** loaded for the language `#1'. Using the pattern for}%
\typeout{** the default language instead.}%
\else
\language=\csname l@#1\endcsname
\fi
#2}}
\providecommand{\BIBdecl}{\relax}
\BIBdecl

\bibitem{rajkumar2010cyber}
R.~Rajkumar, I.~Lee, L.~Sha, and J.~Stankovic, ``Cyber-physical systems: the
  next computing revolution,'' in \emph{Design Automation Conference}.\hskip
  1em plus 0.5em minus 0.4em\relax IEEE, 2010, pp. 731--736.

\bibitem{survey_cloudRObotics_architecture}
G.~Hu, W.~P. Tay, and Y.~Wen, ``Cloud robotics: architecture, challenges and
  applications,'' \emph{IEEE Network}, vol.~26, no.~3, pp. 21--28, 2012.

\bibitem{Survey_curretSatus_CloudRobotics}
J.~Wan, S.~Tang, H.~Yan, D.~Li, S.~Wang, and A.~V. Vasilakos, ``Cloud robotics:
  Current status and open issues,'' \emph{IEEE Access}, vol.~4, pp. 2797--2807,
  2016.

\bibitem{rahman2019energy}
A.~Rahman, J.~Jin, A.~Rahman, A.~Cricenti, M.~Afrin, and Y.-N. Dong,
  ``Energy-efficient optimal task offloading in cloud networked multi-robot
  systems,'' \emph{Computer Networks}, vol. 160, pp. 11 -- 32, 2019.

\bibitem{SPE_QCASH}
M.~R. Mahmud, M.~Afrin, M.~A. Razzaque, M.~M. Hassan, A.~Alelaiwi, and
  M.~Alrubaian, ``Maximizing quality of experience through context-aware mobile
  application scheduling in cloudlet infrastructure,'' \emph{Software: Practice
  and Experience}, vol.~46, no.~11, pp. 1525--1545, 2016.

\bibitem{mouradian2017comprehensive}
C.~Mouradian, D.~Naboulsi, S.~Yangui, R.~H. Glitho, M.~J. Morrow, and P.~A.
  Polakos, ``A comprehensive survey on fog computing: State-of-the-art and
  research challenges,'' \emph{IEEE Communications Surveys \& Tutorials},
  vol.~20, no.~1, pp. 416--464, 2017.

\bibitem{mahmud2019quality}
R.~Mahmud, S.~N. Srirama, K.~Ramamohanarao, and R.~Buyya, ``Quality of
  {E}xperience ({Q}o{E})-aware placement of applications in {F}og computing
  environments,'' \emph{Journal of Parallel and Distributed Computing}, vol.
  132, pp. 190--203, 2019.

\bibitem{Mahmud_Acm_fog_2020}
R.~Mahmud, K.~Ramamohanarao, and R.~Buyya, ``Application management in fog
  computing environments: A taxonomy, review and future directions,'' \emph{ACM
  Computing Survey}, vol.~53, no.~4, 2020.

\bibitem{edge_computing}
W.~Shi, J.~Cao, Q.~Zhang, Y.~Li, and L.~Xu, ``Edge computing: vision and
  challenges,'' \emph{IEEE Internet of Things Journal}, vol.~3, no.~5, pp.
  637--646, 2016.

\bibitem{afrin2019multi}
M.~Afrin, J.~Jin, A.~Rahman, Y.-C. Tian, and A.~Kulkarni, ``Multi-objective
  resource allocation for edge cloud based robotic workflow in smart factory,''
  \emph{Future Generation Computer Systems}, vol.~97, pp. 119--130, 2019.

\bibitem{ma2015robot}
Y.~Ma, Y.~Zhang, J.~Wan, D.~Zhang, and N.~Pan, ``Robot and cloud-assisted
  multi-modal healthcare system,'' \emph{Cluster Computing}, vol.~18, no.~3,
  pp. 1295--1306, 2015.

\bibitem{musat2018advanced}
G.-A. Musat, M.~Colezea, F.~Pop, C.~Negru, M.~Mocanu, C.~Esposito, and
  A.~Castiglione, ``Advanced services for efficient management of smart
  farms,'' \emph{Journal of Parallel and Distributed Computing}, vol. 116, pp.
  3--17, 2018.

\bibitem{smartfactory}
B.~Chen, J.~Wan, L.~Shu, P.~Li, M.~Mukherjee, and B.~Yin, ``Smart factory of
  industry 4.0: Key technologies, application case, and challenges,''
  \emph{IEEE Access}, vol.~6, pp. 6505--6519, 2017.

\bibitem{shaik2020enabling}
M.~S. Shaik, V.~Struh{\'a}r, Z.~Bakhshi, V.-L. Dao, N.~Desai, A.~V.
  Papadopoulos, T.~Nolte, V.~Karagiannis, S.~Schulte, A.~Venito \emph{et~al.},
  ``Enabling fog-based industrial robotics systems,'' in \emph{2020 25th IEEE
  International Conference on Emerging Technologies and Factory Automation
  (ETFA)}, vol.~1.\hskip 1em plus 0.5em minus 0.4em\relax IEEE, 2020, pp.
  61--68.

\bibitem{shiraz2012review}
M.~Shiraz, A.~Gani, R.~H. Khokhar, and R.~Buyya, ``A review on distributed
  application processing frameworks in smart mobile devices for mobile cloud
  computing,'' \emph{IEEE Communications Surveys \& Tutorials}, vol.~15, no.~3,
  pp. 1294--1313, 2012.

\bibitem{taleb2017multi}
T.~Taleb, K.~Samdanis, B.~Mada, H.~Flinck, S.~Dutta, and D.~Sabella, ``On
  multi-access edge computing: A survey of the emerging 5g network edge cloud
  architecture and orchestration,'' \emph{IEEE Communications Surveys \&
  Tutorials}, vol.~19, no.~3, pp. 1657--1681, 2017.

\bibitem{othman2013survey}
M.~Othman, S.~A. Madani, S.~U. Khan \emph{et~al.}, ``A survey of mobile cloud
  computing application models,'' \emph{IEEE Communications Surveys \&
  Tutorials}, vol.~16, no.~1, pp. 393--413, 2013.

\bibitem{dinh2013survey}
H.~T. Dinh, C.~Lee, D.~Niyato, and P.~Wang, ``A survey of mobile cloud
  computing: architecture, applications, and approaches,'' \emph{Wireless
  Communications and Mobile Computing}, vol.~13, no.~18, pp. 1587--1611, 2013.

\bibitem{porambage2018survey}
P.~Porambage, J.~Okwuibe, M.~Liyanage, M.~Ylianttila, and T.~Taleb, ``Survey on
  multi-access edge computing for internet of things realization,'' \emph{IEEE
  Communications Surveys \& Tutorials}, vol.~20, no.~4, pp. 2961--2991, 2018.

\bibitem{markakis2017efficient}
E.~K. Markakis, I.~Politis, A.~Lykourgiotis, Y.~Rebahi, G.~Mastorakis, C.~X.
  Mavromoustakis, and E.~Pallis, ``Efficient next generation emergency
  communications over multi-access edge computing,'' \emph{IEEE Communications
  Magazine}, vol.~55, no.~11, pp. 92--97, 2017.

\bibitem{sarker2019offloading}
V.~Sarker, J.~P. Queralta, T.~Gia, H.~Tenhunen, and T.~Westerlund, ``Offloading
  slam for indoor mobile robots with edge-fog-cloud computing,'' in \emph{2019
  1st International Conference on Advances in Science, Engineering and Robotics
  Technology (ICASERT)}.\hskip 1em plus 0.5em minus 0.4em\relax IEEE, 2019, pp.
  1--6.

\bibitem{beigi_smartcity_cloudRobotics}
N.~K. {Beigi}, B.~{Partov}, and S.~{Farokhi}, ``Real-time cloud robotics in
  practical smart city applications,'' in \emph{2017 IEEE 28th Annual
  International Symposium on Personal, Indoor, and Mobile Radio Communications
  (PIMRC)}, 2017, pp. 1--5.

\bibitem{birk2009networking}
A.~Birk, S.~Schwertfeger, and K.~Pathak, ``A networking framework for
  teleoperation in safety, security, and rescue robotics,'' \emph{IEEE Wireless
  Communications}, vol.~16, no.~1, pp. 6--13, 2009.

\bibitem{AK_MotionandConnectivity}
A.~Rahman, J.~Jin, A.~Cricenti, A.~Rahman, and M.~Panda, ``Motion and
  connectivity aware offloading in cloud robotics via genetic algorithm,'' in
  \emph{2017 IEEE Global Communications Conference}.\hskip 1em plus 0.5em minus
  0.4em\relax IEEE, 2017, pp. 1--6.

\bibitem{AK_SmartCity}
A.~Rahman, J.~Jin, A.~Cricenti, A.~Rahman, and D.~Yuan, ``A cloud robotics
  framework of optimal task offloading for smart city applications,'' in
  \emph{2016 IEEE Global Communications Conference (GLOBECOM)}.\hskip 1em plus
  0.5em minus 0.4em\relax IEEE, 2016, pp. 1--7.

\bibitem{wee2015industry}
D.~Wee, R.~Kelly, J.~Cattel, and M.~Breunig, ``Industry 4.0-how to navigate
  digitization of the manufacturing sector,'' \emph{McKinsey \& Company},
  vol.~58, 2015.

\bibitem{industry4.0_survey}
Y.~Lu, ``Industry 4.0: A survey on technologies, applications and open research
  issues,'' \emph{Journal of Industrial Information Integration}, vol.~6, pp.
  1--10, 2017.

\bibitem{Aissam2019}
M.~Aissam, M.~Benbrahim, and M.~N. Kabbaj, \emph{Cloud robotic: {O}pening a new
  road to the industry 4.0}.\hskip 1em plus 0.5em minus 0.4em\relax Springer
  Singapore, 2019, pp. 1--20.

\bibitem{wang2016ubiquitous}
W.~Wang, X.~Zhu, L.~Wang, Q.~Qiu, and Q.~Cao, ``Ubiquitous robotic technology
  for smart manufacturing system,'' \emph{Computational Intelligence and
  Neuroscience}, vol. 2016, 2016.

\bibitem{wang2016towards_smartfactory}
S.~Wang, J.~Wan, D.~Zhang, D.~Li, and C.~Zhang, ``Towards smart factory for
  industry 4.0: a self-organized multi-agent system with big data based
  feedback and coordination,'' \emph{Computer Networks}, vol. 101, pp.
  158--168, 2016.

\bibitem{wang2016implementing}
S.~Wang, J.~Wan, D.~Li, and C.~Zhang, ``Implementing smart factory of industrie
  4.0: an outlook,'' \emph{International Journal of Distributed Sensor
  Networks}, vol.~12, no.~1, p. 3159805, 2016.

\bibitem{wan2019reconfigurable}
J.~Wan, S.~Tang, D.~Li, M.~Imran, C.~Zhang, C.~Liu, and Z.~Pang,
  ``Reconfigurable smart factory for drug packing in healthcare industry 4.0,''
  \emph{IEEE Transactions on Industrial Informatics}, vol.~15, no.~1, pp.
  507--516, 2019.

\bibitem{peake2015cloud}
I.~D. Peake, A.~Vuyyuru, J.~O. Blech, N.~Vergnaud, and L.~Fernando,
  ``Cloud-based analysis and control for robots in industrial automation,'' in
  \emph{2015 IEEE 21st International Conference on Parallel and Distributed
  Systems (ICPADS)}.\hskip 1em plus 0.5em minus 0.4em\relax IEEE, 2015, pp.
  837--840.

\bibitem{cardarelli2015cloud}
E.~Cardarelli, L.~Sabattini, C.~Secchi, and C.~Fantuzzi, ``Cloud robotics
  paradigm for enhanced navigation of autonomous vehicles in real world
  industrial applications,'' in \emph{2015 IEEE/RSJ International Conference on
  Intelligent Robots and Systems (IROS)}.\hskip 1em plus 0.5em minus
  0.4em\relax IEEE, 2015, pp. 4518--4523.

\bibitem{cardarelli2017cooperative}
E.~Cardarelli, V.~Digani, L.~Sabattini, C.~Secchi, and C.~Fantuzzi,
  ``Cooperative cloud robotics architecture for the coordination of multi-{AGV}
  systems in industrial warehouses,'' \emph{Mechatronics}, vol.~45, pp. 1--13,
  2017.

\bibitem{wang2017cloud}
S.~Wang, C.~Zhang, C.~Liu, D.~Li, and H.~Tang, ``Cloud-assisted interaction and
  negotiation of industrial robots for the smart factory,'' \emph{Computers \&
  Electrical Engineering}, vol.~63, pp. 66--78, 2017.

\bibitem{wang2017integrated}
S.~Wang, J.~Ouyang, D.~Li, and C.~Liu, ``An integrated industrial ethernet
  solution for the implementation of smart factory,'' \emph{IEEE Access},
  vol.~5, pp. 25\,455--25\,462, 2017.

\bibitem{wang2017ubiquitous}
X.~V. Wang, L.~Wang, A.~Mohammed, and M.~Givehchi, ``Ubiquitous manufacturing
  system based on {C}loud: {A} robotics application,'' \emph{Robotics and
  Computer-Integrated Manufacturing}, vol.~45, pp. 116--125, 2017.

\bibitem{wang2018cloud}
S.~Wang, J.~Wan, M.~Imran, D.~Li, and C.~Zhang, ``Cloud-based smart
  manufacturing for personalized candy packing application,'' \emph{The Journal
  of Supercomputing}, vol.~74, no.~9, pp. 4339--4357, 2018.

\bibitem{wan_Context-aware-cloud-robotics}
J.~{Wan}, S.~{Tang}, Q.~{Hua}, D.~{Li}, C.~{Liu}, and J.~{Lloret},
  ``Context-aware cloud robotics for material handling in cognitive industrial
  {I}nternet of {T}hings,'' \emph{IEEE Internet of Things Journal}, vol.~5,
  no.~4, pp. 2272--2281, 2018.

\bibitem{wan2018artificial}
J.~Wan, J.~Yang, Z.~Wang, and Q.~Hua, ``Artificial intelligence for
  cloud-assisted smart factory,'' \emph{IEEE Access}, vol.~6, pp.
  55\,419--55\,430, 2018.

\bibitem{Smart_Manufacturing}
H.~Yan, Q.~Hua, Y.~Wang, W.~Wei, and M.~Imran, ``Cloud robotics in smart
  manufacturing environments: Challenges and countermeasures,'' \emph{Computers
  \& Electrical Engineering}, vol.~63, pp. 56--65, 2017.

\bibitem{bonaccorsi2015design}
M.~Bonaccorsi, L.~Fiorini, F.~Cavallo, R.~Esposito, and P.~Dario, ``Design of
  cloud robotic services for senior citizens to improve independent living and
  personal health management,'' in \emph{Ambient Assisted Living}.\hskip 1em
  plus 0.5em minus 0.4em\relax Springer, 2015, pp. 465--475.

\bibitem{bonaccorsi2016cloud}
M.~Bonaccorsi, L.~Fiorini, F.~Cavallo, A.~Saffiotti, and P.~Dario, ``A cloud
  robotics solution to improve social assistive robots for active and healthy
  aging,'' \emph{International Journal of Social Robotics}, vol.~8, no.~3, pp.
  393--408, 2016.

\bibitem{fiorini2017enabling}
L.~Fiorini, R.~Esposito, M.~Bonaccorsi, C.~Petrazzuolo, F.~Saponara,
  R.~Giannantonio, G.~De~Petris, P.~Dario, and F.~Cavallo, ``Enabling
  personalised medical support for chronic disease management through a hybrid
  robot-cloud approach,'' \emph{Autonomous Robots}, vol.~41, no.~5, pp.
  1263--1276, 2017.

\bibitem{kamei2012cloud}
K.~Kamei, S.~Nishio, N.~Hagita, and M.~Sato, ``Cloud networked robotics,''
  \emph{IEEE Network}, vol.~26, no.~3, pp. 28--34, 2012.

\bibitem{kamei2017cloud}
K.~Kamei, F.~Zanlungo, T.~Kanda, Y.~Horikawa, T.~Miyashita, and N.~Hagita,
  ``Cloud networked robotics for social robotic services extending robotic
  functional service standards to support autonomous mobility system in social
  environments,'' in \emph{2017 14th International Conference on Ubiquitous
  Robots and Ambient Intelligence (URAI)}.\hskip 1em plus 0.5em minus
  0.4em\relax IEEE, 2017, pp. 897--902.

\bibitem{ng2015cloud}
M.~K. Ng, S.~Primatesta, L.~Giuliano, M.~L. Lupetti, L.~O. Russo, G.~A.
  Farulla, M.~Indaco, S.~Rosa, C.~Germak, and B.~Bona, ``A cloud robotics
  system for telepresence enabling mobility impaired people to enjoy the whole
  museum experience,'' in \emph{2015 10th International Conference on Design \&
  Technology of Integrated Systems in Nanoscale Era (DTIS)}.\hskip 1em plus
  0.5em minus 0.4em\relax IEEE, 2015, pp. 1--6.

\bibitem{manzi2017design}
A.~Manzi, L.~Fiorini, R.~Esposito, M.~Bonaccorsi, I.~Mannari, P.~Dario, and
  F.~Cavallo, ``Design of a cloud robotic system to support senior citizens:
  The kubo experience,'' \emph{Autonomous Robots}, vol.~41, no.~3, pp.
  699--709, 2017.

\bibitem{zhang2015health}
Y.~Zhang, M.~Qiu, C.-W. Tsai, M.~M. Hassan, and A.~Alamri, ``Health-cps:
  Healthcare cyber-physical system assisted by cloud and big data,'' \emph{IEEE
  Systems Journal}, vol.~11, no.~1, pp. 88--95, 2015.

\bibitem{pham2018delivering}
M.~Pham, Y.~Mengistu, H.~Do, and W.~Sheng, ``Delivering home healthcare through
  a cloud-based smart home environment (coshe),'' \emph{Future Generation
  Computer Systems}, vol.~81, pp. 129--140, 2018.

\bibitem{chung2019chatbot}
K.~Chung and R.~C. Park, ``Chatbot-based heathcare service with a knowledge
  base for cloud computing,'' \emph{Cluster Computing}, vol.~22, no.~1, pp.
  1925--1937, 2019.

\bibitem{wan2020cognitive}
S.~Wan, Z.~Gu, and Q.~Ni, ``Cognitive computing and wireless communications on
  the edge for healthcare service robots,'' \emph{Computer Communications},
  vol. 149, pp. 99--106, 2020.

\bibitem{bauer2018design}
J.~Bauer and N.~Aschenbruck, ``Design and implementation of an agricultural
  monitoring system for smart farming,'' in \emph{2018 IoT Vertical and Topical
  Summit on Agriculture-Tuscany (IoT Tuscany)}.\hskip 1em plus 0.5em minus
  0.4em\relax IEEE, 2018, pp. 1--6.

\bibitem{duckett2018agricultural}
T.~Duckett, S.~Pearson, S.~Blackmore, and B.~Grieve, ``Agricultural robotics:
  the future of robotic agriculture,'' \emph{arXiv preprint arXiv:1806.06762},
  2018.

\bibitem{pivoto2018scientific}
D.~Pivoto, P.~D. Waquil, E.~Talamini, C.~P.~S. Finocchio, V.~F. Dalla~Corte,
  and G.~de~Vargas~Mores, ``Scientific development of smart farming
  technologies and their application in {B}razil,'' \emph{Information
  Processing in Agriculture}, vol.~5, no.~1, pp. 21--32, 2018.

\bibitem{kulbacki2018survey}
M.~Kulbacki, J.~Segen, W.~Knie{\'c}, R.~Klempous, K.~Kluwak, J.~Nikodem,
  J.~Kulbacka, and A.~Serester, ``Survey of drones for agriculture automation
  from planting to harvest,'' in \emph{2018 IEEE 22nd International Conference
  on Intelligent Engineering Systems (INES)}.\hskip 1em plus 0.5em minus
  0.4em\relax IEEE, 2018, pp. 353--358.

\bibitem{valecce2019interplay}
G.~Valecce, S.~Strazzella, and L.~A. Grieco, ``On the interplay between 5g,
  mobile edge computing and robotics in smart agriculture scenarios,'' in
  \emph{International Conference on Ad-Hoc Networks and Wireless}.\hskip 1em
  plus 0.5em minus 0.4em\relax Springer, 2019, pp. 549--559.

\bibitem{o2019edge}
M.~O'Grady, D.~Langton, and G.~O'Hare, ``Edge computing: A tractable model for
  smart agriculture?'' \emph{Artificial Intelligence in Agriculture}, vol.~3,
  pp. 42--51, 2019.

\bibitem{danton2020development}
A.~Danton, J.-C. Roux, B.~Dance, C.~Cariou, and R.~Lenain, ``Development of a
  spraying robot for precision agriculture: An edge following approach,'' in
  \emph{2020 IEEE Conference on Control Technology and Applications
  (CCTA)}.\hskip 1em plus 0.5em minus 0.4em\relax IEEE, 2020, pp. 267--272.

\bibitem{cho2012agricultural}
Y.~Cho, K.~Cho, C.~Shin, J.~Park, and E.-S. Lee, ``An agricultural expert cloud
  for a smart farm,'' in \emph{Future Information Technology, Application, and
  Service}.\hskip 1em plus 0.5em minus 0.4em\relax Springer, 2012, pp.
  657--662.

\bibitem{apostol2015towards}
E.~Apostol, C.~Leordeanu, M.~Mocanu, and V.~Cristea, ``Towards a hybrid
  local-cloud framework for smart farms,'' in \emph{2015 20th International
  Conference on Control Systems and Computer Science}.\hskip 1em plus 0.5em
  minus 0.4em\relax IEEE, 2015, pp. 820--824.

\bibitem{vasisht2017farmbeats}
D.~Vasisht, Z.~Kapetanovic, J.~Won, X.~Jin, R.~Chandra, S.~Sinha, A.~Kapoor,
  M.~Sudarshan, and S.~Stratman, ``Farmbeats: An {I}o{T} platform for
  data-driven agriculture,'' in \emph{14th USENIX Symposium on Networked
  Systems Design and Implementation (NSDI 17)}.\hskip 1em plus 0.5em minus
  0.4em\relax USENIX Association, 2017, pp. 515--529.

\bibitem{kim2018iot}
S.~Kim, M.~Lee, and C.~Shin, ``Io{T}-based strawberry disease prediction system
  for smart farming,'' \emph{Sensors}, vol.~18, no.~11, p. 4051, 2018.

\bibitem{nintanavongsa2017impact}
P.~Nintanavongsa and I.~Pitimon, ``Impact of sensor mobility on {UAV}-based
  smart farm communications,'' in \emph{2017 International Electrical
  Engineering Congress (IEECON)}.\hskip 1em plus 0.5em minus 0.4em\relax IEEE,
  2017, pp. 1--4.

\bibitem{lottes2017uav}
P.~Lottes, R.~Khanna, J.~Pfeifer, R.~Siegwart, and C.~Stachniss, ``{UAV}-based
  crop and weed classification for smart farming,'' in \emph{2017 IEEE
  International Conference on Robotics and Automation (ICRA)}.\hskip 1em plus
  0.5em minus 0.4em\relax IEEE, 2017, pp. 3024--3031.

\bibitem{kim2019unmanned}
J.~Kim, S.~Kim, C.~Ju, and H.~I. Son, ``Unmanned aerial vehicles in
  agriculture: A review of perspective of platform, control, and
  applications,'' \emph{IEEE Access}, vol.~7, pp. 105\,100--105\,115, 2019.

\bibitem{kamilaris2016agri}
A.~Kamilaris, F.~Gao, F.~X. Prenafeta-Bold{\'u}, and M.~I. Ali, ``Agri-{I}o{T}:
  A semantic framework for {I}nternet of {T}hings-enabled smart farming
  applications,'' in \emph{2016 IEEE 3rd World Forum on Internet of Things
  (WF-IoT)}.\hskip 1em plus 0.5em minus 0.4em\relax IEEE, 2016, pp. 442--447.

\bibitem{grieve2019challenges}
B.~D. Grieve, T.~Duckett, M.~Collison, L.~Boyd, J.~West, H.~Yin, F.~Arvin, and
  S.~Pearson, ``The challenges posed by global broadacre crops in delivering
  smart agri-robotic solutions: {A} fundamental rethink is required,''
  \emph{Global Food Security}, vol.~23, pp. 116--124, 2019.

\bibitem{ermacora2013cloud}
G.~Ermacora, A.~Toma, B.~Bona, M.~Chiaberge, M.~Silvagni, M.~Gaspardone, and
  R.~Antonini, ``A cloud robotics architecture for an emergency management and
  monitoring service in a smart city environment,'' \emph{Polytech. Univ.
  Turin, Turin, Italy, Tech. Rep}, 2013.

\bibitem{gregory2016application}
J.~Gregory, J.~Fink, E.~Stump, J.~Twigg, J.~Rogers, D.~Baran, N.~Fung, and
  S.~Young, ``Application of multi-robot systems to disaster-relief scenarios
  with limited communication,'' in \emph{Field and Service Robotics}.\hskip 1em
  plus 0.5em minus 0.4em\relax Springer, 2016, pp. 639--653.

\bibitem{jangid2016cloud}
N.~Jangid and B.~Sharma, ``Cloud computing and robotics for disaster
  management,'' in \emph{2016 7th International Conference on Intelligent
  Systems, Modelling and Simulation (ISMS)}.\hskip 1em plus 0.5em minus
  0.4em\relax IEEE, 2016, pp. 20--24.

\bibitem{botta2017networking}
A.~Botta, J.~Cacace, V.~Lippiello, B.~Siciliano, and G.~Ventre, ``Networking
  for cloud robotics: A case study based on the {SHERPA} project,'' in
  \emph{Proceedings of the 2017 International Conference on Cloud and Robotics
  (ICCR), Saint Quentin, France}, 2017, pp. 22--23.

\bibitem{Robots-as-a-Service-Search-and-Rescue}
C.~Mouradian, S.~Yangui, and R.~H. Glitho, ``Robots as-a-service in cloud
  computing: search and rescue in large-scale disasters case study,'' in
  \emph{2018 15th IEEE Annual Consumer Communications \& Networking Conference
  (CCNC)}.\hskip 1em plus 0.5em minus 0.4em\relax IEEE, 2018, pp. 1--7.

\bibitem{marconi2012sherpa}
L.~Marconi, C.~Melchiorri, M.~Beetz, D.~Pangercic, R.~Siegwart, S.~Leutenegger,
  R.~Carloni, S.~Stramigioli, H.~Bruyninckx, P.~Doherty \emph{et~al.}, ``The
  {SHERPA} project: Smart collaboration between humans and ground-aerial robots
  for improving rescuing activities in alpine environments,'' in \emph{2012
  IEEE International Symposium on Safety, Security, and Rescue Robotics
  (SSRR)}.\hskip 1em plus 0.5em minus 0.4em\relax IEEE, 2012, pp. 1--4.

\bibitem{afrin2018energy}
M.~Afrin, J.~Jin, and A.~Rahman, ``Energy-delay co-optimization of resource
  allocation for robotic services in cloudlet infrastructure,'' in
  \emph{International Conference on Service-Oriented Computing}.\hskip 1em plus
  0.5em minus 0.4em\relax Springer, 2018, pp. 295--303.

\bibitem{yousafzai2017cloud}
A.~Yousafzai, A.~Gani, R.~M. Noor, M.~Sookhak, H.~Talebian, M.~Shiraz, and
  M.~K. Khan, ``Cloud resource allocation schemes: review, taxonomy, and
  opportunities,'' \emph{Knowledge and Information Systems}, vol.~50, no.~2,
  pp. 347--381, 2017.

\bibitem{manvi2014resource}
S.~S. Manvi and G.~K. Shyam, ``Resource management for infrastructure as a
  service (iaas) in cloud computing: A survey,'' \emph{Journal of Network and
  Computer Applications}, vol.~41, pp. 424--440, 2014.

\bibitem{toosi2014interconnected}
A.~N. Toosi, R.~N. Calheiros, and R.~Buyya, ``Interconnected cloud computing
  environments: Challenges, taxonomy, and survey,'' \emph{ACM Computing
  Surveys}, vol.~47, no.~1, p.~7, 2014.

\bibitem{luong2017resource}
N.~C. Luong, P.~Wang, D.~Niyato, Y.~Wen, and Z.~Han, ``Resource management in
  cloud networking using economic analysis and pricing models: A survey,''
  \emph{IEEE Communications Surveys \& Tutorials}, vol.~19, no.~2, pp.
  954--1001, 2017.

\bibitem{mann2015allocation}
Z.~{\'A}. Mann, ``Allocation of virtual machines in cloud data centers—a
  survey of problem models and optimization algorithms,'' \emph{Acm Computing
  Surveys}, vol.~48, no.~1, p.~11, 2015.

\bibitem{singh2016survey}
S.~Singh and I.~Chana, ``A survey on resource scheduling in cloud computing:
  Issues and challenges,'' \emph{Journal of grid computing}, vol.~14, no.~2,
  pp. 217--264, 2016.

\bibitem{huang2013survey}
L.~Huang, H.-s. Chen, and T.-t. Hu, ``Survey on resource allocation policy and
  job scheduling algorithms of cloud computing,'' \emph{Journal of Software},
  vol.~8, no.~2, p. 481, 2013.

\bibitem{mastelic2015cloud}
T.~Mastelic, A.~Oleksiak, H.~Claussen, I.~Brandic, J.-M. Pierson, and A.~V.
  Vasilakos, ``Cloud computing: Survey on energy efficiency,'' \emph{ACM
  Computing Surveys}, vol.~47, no.~2, p.~33, 2015.

\bibitem{zhan2015cloud}
Z.-H. Zhan, X.-F. Liu, Y.-J. Gong, J.~Zhang, H.~S.-H. Chung, and Y.~Li, ``Cloud
  computing resource scheduling and a survey of its evolutionary approaches,''
  \emph{ACM Computing Surveys}, vol.~47, no.~4, p.~63, 2015.

\bibitem{zhang2012auction}
Y.~Zhang, C.~Lee, D.~Niyato, and P.~Wang, ``Auction approaches for resource
  allocation in wireless systems: A survey,'' \emph{IEEE Communications Surveys
  \& Tutorials}, vol.~15, no.~3, pp. 1020--1041, 2012.

\bibitem{budzisz2014dynamic}
{\L}.~Budzisz, F.~Ganji, G.~Rizzo, M.~A. Marsan, M.~Meo, Y.~Zhang, G.~Koutitas,
  L.~Tassiulas, S.~Lambert, B.~Lannoo \emph{et~al.}, ``Dynamic resource
  provisioning for energy efficiency in wireless access networks: A survey and
  an outlook,'' \emph{IEEE Communications Surveys \& Tutorials}, vol.~16,
  no.~4, pp. 2259--2285, 2014.

\bibitem{maallawi2014comprehensive}
R.~Maallawi, N.~Agoulmine, B.~Radier, and T.~B. Meriem, ``A comprehensive
  survey on offload techniques and management in wireless access and core
  networks,'' \emph{IEEE Communications Surveys \& Tutorials}, vol.~17, no.~3,
  pp. 1582--1604, 2014.

\bibitem{rebecchi2014data}
F.~Rebecchi, M.~D. De~Amorim, V.~Conan, A.~Passarella, R.~Bruno, and M.~Conti,
  ``Data offloading techniques in cellular networks: A survey,'' \emph{IEEE
  Communications Surveys \& Tutorials}, vol.~17, no.~2, pp. 580--603, 2014.

\bibitem{xu2013survey}
Y.~Xu and S.~Mao, ``A survey of mobile cloud computing for rich media
  applications,'' \emph{IEEE Wireless Communications}, vol.~20, no.~3, pp.
  46--53, 2013.

\bibitem{abolfazli2013cloud}
S.~Abolfazli, Z.~Sanaei, E.~Ahmed, A.~Gani, and R.~Buyya, ``Cloud-based
  augmentation for mobile devices: motivation, taxonomies, and open
  challenges,'' \emph{IEEE Communications Surveys \& Tutorials}, vol.~16,
  no.~1, pp. 337--368, 2013.

\bibitem{sanaei2013heterogeneity}
Z.~Sanaei, S.~Abolfazli, A.~Gani, and R.~Buyya, ``Heterogeneity in mobile cloud
  computing: taxonomy and open challenges,'' \emph{IEEE Communications Surveys
  \& Tutorials}, vol.~16, no.~1, pp. 369--392, 2013.

\bibitem{mach2017mobile}
P.~Mach and Z.~Becvar, ``Mobile edge computing: A survey on architecture and
  computation offloading,'' \emph{IEEE Communications Surveys \& Tutorials},
  vol.~19, no.~3, pp. 1628--1656, 2017.

\bibitem{mao2017survey}
Y.~Mao, C.~You, J.~Zhang, K.~Huang, and K.~B. Letaief, ``A survey on mobile
  edge computing: The communication perspective,'' \emph{IEEE Communications
  Surveys \& Tutorials}, vol.~19, no.~4, pp. 2322--2358, 2017.

\bibitem{wang2017survey}
S.~Wang, X.~Zhang, Y.~Zhang, L.~Wang, J.~Yang, and W.~Wang, ``A survey on
  mobile edge networks: Convergence of computing, caching and communications,''
  \emph{IEEE Access}, vol.~5, pp. 6757--6779, 2017.

\bibitem{khamis2015multi}
A.~Khamis, A.~Hussein, and A.~Elmogy, ``Multi-robot task allocation: A review
  of the state-of-the-art,'' in \emph{Cooperative Robots and Sensor Networks
  2015}.\hskip 1em plus 0.5em minus 0.4em\relax Springer, 2015, pp. 31--51.

\bibitem{mosteo2010survey}
A.~R. Mosteo and L.~Montano, ``A survey of multi-robot task allocation,''
  \emph{Instituto de Investigaci{\'o}n en Ingenier{\i}a de Arag{\'o}n,
  University of Zaragoza, Zaragoza, Spain, Technical Report No.
  AMI-009-10-TEC}, 2010.

\bibitem{survey_CloudRoboticsandAutomation}
B.~Kehoe, S.~Patil, P.~Abbeel, and K.~Goldberg, ``A survey of research on cloud
  robotics and automation,'' \emph{IEEE Transactions on Automation Science and
  Engineering}, vol.~12, no.~2, pp. 398--409, 2015.

\bibitem{saha_CloudRobotics_Survey}
O.~Saha and P.~Dasgupta, ``A comprehensive survey of recent trends in cloud
  robotics architectures and applications,'' \emph{Robotics}, vol.~7, no.~3,
  p.~47, 2018.

\bibitem{Robotic_Cooperation}
W.~{Chen}, Y.~{Yaguchi}, K.~{Naruse}, Y.~{Watanobe}, K.~{Nakamura}, and
  J.~{Ogawa}, ``A study of robotic cooperation in cloud robotics: architecture
  and challenges,'' \emph{IEEE Access}, vol.~6, pp. 36\,662--36\,682, 2018.

\bibitem{huiyong2013building}
W.~Huiyong, W.~Jingyang, and H.~Min, ``Building a smart home system with wsn
  and service robot,'' in \emph{2013 Fifth International Conference on
  Measuring Technology and Mechatronics Automation}.\hskip 1em plus 0.5em minus
  0.4em\relax IEEE, 2013, pp. 353--356.

\bibitem{li2015ehopes}
J.~Li, J.~Jin, D.~Yuan, M.~Palaniswami, and K.~Moessner, ``Ehopes:
  Data-centered fog platform for smart living,'' in \emph{2015 International
  Telecommunication Networks and Applications Conference (ITNAC)}.\hskip 1em
  plus 0.5em minus 0.4em\relax IEEE, 2015, pp. 308--313.

\bibitem{pan2016homecloud}
J.~Pan, L.~Ma, R.~Ravindran, and P.~TalebiFard, ``Homecloud: An edge cloud
  framework and testbed for new application delivery,'' in \emph{2016 23rd
  International Conference on Telecommunications (ICT)}.\hskip 1em plus 0.5em
  minus 0.4em\relax IEEE, 2016, pp. 1--6.

\bibitem{do2018rish}
H.~M. Do, M.~Pham, W.~Sheng, D.~Yang, and M.~Liu, ``Rish: A robot-integrated
  smart home for elderly care,'' \emph{Robotics and Autonomous Systems}, vol.
  101, pp. 74--92, 2018.

\bibitem{yang2020ai}
J.~Yang, R.~Wang, X.~Guan, M.~M. Hassan, A.~Almogren, and A.~Alsanad,
  ``Ai-enabled emotion-aware robot: The fusion of smart clothing, edge clouds
  and robotics,'' \emph{Future Generation Computer Systems}, vol. 102, pp.
  701--709, 2020.

\bibitem{liu2019summary}
Y.~Liu and Y.~Xu, ``Summary of cloud robot research,'' in \emph{2019 25th
  International Conference on Automation and Computing (ICAC)}.\hskip 1em plus
  0.5em minus 0.4em\relax IEEE, 2019, pp. 1--5.

\bibitem{Forbes}
\BIBentryALTinterwordspacing
Forbes. (2019) {Robots As A Service: A Technology Trend Every Business Must
  Consider}. [Online]. Available:
  \url{https://www.forbes.com/sites/bernardmarr/2019/08/05/robots-as-a-service-a-technology-trend-every-business-must-consider}
\BIBentrySTDinterwordspacing

\bibitem{CloudMinds}
\BIBentryALTinterwordspacing
CloudMinds. (2019) {CloudMinds}. [Online]. Available:
  \url{https://www.en.cloudminds.com/}
\BIBentrySTDinterwordspacing

\bibitem{rodrigues2019machine}
T.~K. Rodrigues, K.~Suto, H.~Nishiyama, J.~Liu, and N.~Kato, ``Machine learning
  meets computation and communication control in evolving edge and cloud:
  Challenges and future perspective,'' \emph{IEEE Communications Surveys \&
  Tutorials}, vol.~22, no.~1, pp. 38--67, 2020.

\bibitem{sun2017cost}
Y.~Sun, X.-s. Zhou, and G.~Yang, ``Cost aware offloading selection and resource
  allocation for cloud based multi-robot systems,'' \emph{IEICE Transactions on
  Information and Systems}, vol. 100, no.~12, pp. 3022--3026, 2017.

\bibitem{chen2018qos}
W.~Chen, Y.~Yaguchi, K.~Naruse, Y.~Watanobe, and K.~Nakamura, ``{QoS}-aware
  robotic streaming workflow allocation in cloud robotics systems,'' \emph{IEEE
  Transactions on Services Computing}, pp. 1--1, 2018.

\bibitem{liu2018reinforcement}
H.~Liu, S.~Liu, and K.~Zheng, ``A reinforcement learning-based resource
  allocation scheme for cloud robotics,'' \emph{IEEE Access}, vol.~6, pp.
  17\,215--17\,222, 2018.

\bibitem{Rapyuta}
G.~Mohanarajah, D.~Hunziker, R.~D'Andrea, and M.~Waibel, ``Rapyuta: A cloud
  robotics platform,'' \emph{IEEE Transactions on Automation Science and
  Engineering}, vol.~12, no.~2, pp. 481--493, 2014.

\bibitem{beksi_object_recognition}
W.~J. Beksi, J.~Spruth, and N.~Papanikolopoulos, ``{CORE}: A cloud-based object
  recognition engine for robotics,'' in \emph{2015 IEEE/RSJ International
  Conference on Intelligent Robots and Systems ({IROS})}.\hskip 1em plus 0.5em
  minus 0.4em\relax IEEE, 2015, pp. 4512--4517.

\bibitem{infrastructure-for-robotic-applications}
C.~Mouradian, F.~Z. Errounda, F.~Belqasmi, and R.~Glitho, ``An infrastructure
  for robotic applications as cloud computing services,'' in \emph{2014 IEEE
  World Forum on Internet of Things (WF-IoT)}.\hskip 1em plus 0.5em minus
  0.4em\relax IEEE, 2014, pp. 377--382.

\bibitem{ramharuk2014cloud}
V.~Ramharuk and I.~Osunmakinde, ``Cloud robotics: {A} framework towards
  cloud-enabled multi-robotics survivability,'' in \emph{Proceedings of the
  Southern African Institute for Computer Scientist and Information
  Technologists Annual Conference 2014 on SAICSIT 2014 Empowered by
  Technology}.\hskip 1em plus 0.5em minus 0.4em\relax ACM, 2014, p.~82.

\bibitem{Robot-As-a-Service}
Y.~Chen, Z.~Du, and M.~Garc{\'\i}a-Acosta, ``Robot as a service in cloud
  computing,'' in \emph{2010 Fifth IEEE International Symposium on Service
  Oriented System Engineering}.\hskip 1em plus 0.5em minus 0.4em\relax IEEE,
  2010, pp. 151--158.

\bibitem{doriya2017development}
R.~Doriya, ``Development of a cloud-based {RTAB}-map service for robots,'' in
  \emph{2017 IEEE International Conference on Real-time Computing and Robotics
  (RCAR)}.\hskip 1em plus 0.5em minus 0.4em\relax IEEE, 2017, pp. 598--605.

\bibitem{Mohanarajah_2}
G.~Mohanarajah, V.~Usenko, M.~Singh, R.~D'Andrea, and M.~Waibel, ``Cloud-based
  collaborative 3d mapping in real-time with low-cost robots,'' \emph{IEEE
  Transactions on Automation Science and Engineering}, vol.~12, no.~2, pp.
  423--431, 2015.

\bibitem{toffetti2017cloud}
G.~Toffetti, T.~L{\"o}tscher, S.~Kenzhegulov, J.~Spillner, and T.~M. Bohnert,
  ``Cloud robotics: {SLAM} and autonomous exploration on {P}aa{S},'' in
  \emph{Companion Proceedings of the 10th International Conference on Utility
  and Cloud Computing}.\hskip 1em plus 0.5em minus 0.4em\relax ACM, 2017, pp.
  65--70.

\bibitem{yun2017towards}
P.~Yun, J.~Jiao, and M.~Liu, ``Towards a cloud robotics platform for
  distributed visual slam,'' in \emph{International Conference on Computer
  Vision Systems}.\hskip 1em plus 0.5em minus 0.4em\relax Springer, 2017, pp.
  3--15.

\bibitem{ali2018fastslam}
S.~S. Ali, A.~Hammad, and A.~S.~T. Eldien, ``{FastSLAM} 2.0 tracking and
  mapping as a cloud robotics service,'' \emph{Computers \& Electrical
  Engineering}, vol.~69, pp. 412--421, 2018.

\bibitem{PPAAS}
M.-L. Lam and K.-Y. Lam, ``Path planning as a service {PP}aa{S}: Cloud-based
  robotic path planning,'' in \emph{2014 IEEE International Conference on
  Robotics and Biomimetics (ROBIO 2014)}.\hskip 1em plus 0.5em minus
  0.4em\relax IEEE, 2014, pp. 1839--1844.

\bibitem{limosani2016enabling}
R.~Limosani, A.~Manzi, L.~Fiorini, F.~Cavallo, and P.~Dario, ``Enabling global
  robot navigation based on a cloud robotics approach,'' \emph{International
  Journal of Social Robotics}, vol.~8, no.~3, pp. 371--380, 2016.

\bibitem{tian2017cloud}
N.~Tian, M.~Matl, J.~Mahler, Y.~X. Zhou, S.~Staszak, C.~Correa, S.~Zheng,
  Q.~Li, R.~Zhang, and K.~Goldberg, ``A cloud robot system using the dexterity
  network and berkeley robotics and automation as a service ({B}rass),'' in
  \emph{2017 IEEE International Conference on Robotics and Automation
  (ICRA)}.\hskip 1em plus 0.5em minus 0.4em\relax IEEE, 2017, pp. 1615--1622.

\bibitem{vick2016model}
A.~Vick, J.~Guhl, and J.~Kr{\"u}ger, ``Model predictive control as a
  service-concept and architecture for use in cloud-based robot control,'' in
  \emph{2016 21st International Conference on Methods and Models in Automation
  and Robotics (MMAR)}.\hskip 1em plus 0.5em minus 0.4em\relax IEEE, 2016, pp.
  607--612.

\bibitem{merle2017mobile}
P.~Merle, C.~Gourdin, and N.~Mitton, ``Mobile cloud robotics as a service with
  {OCCI}ware,'' in \emph{2017 IEEE International Congress on Internet of Things
  (ICIOT)}.\hskip 1em plus 0.5em minus 0.4em\relax IEEE, 2017, pp. 50--57.

\bibitem{wen2016towards}
S.~Wen, B.~Ding, H.~Wang, B.~Hu, H.~Liu, and P.~Shi, ``Towards migrating
  resource-consuming robotic software packages to cloud,'' in \emph{2016 IEEE
  International Conference on Real-time Computing and Robotics (RCAR)}.\hskip
  1em plus 0.5em minus 0.4em\relax IEEE, 2016, pp. 283--288.

\bibitem{koubaa2014service}
A.~Koubaa, ``A service-oriented architecture for virtualizing robots in
  robot-as-a-service clouds,'' in \emph{International Conference on
  Architecture of Computing Systems}.\hskip 1em plus 0.5em minus 0.4em\relax
  Springer, 2014, pp. 196--208.

\bibitem{C2TAM}
L.~Riazuelo, J.~Civera, and J.~M. Montiel, ``{C2TAM}: A cloud framework for
  cooperative tracking and mapping,'' \emph{Robotics and Autonomous Systems},
  vol.~62, no.~4, pp. 401--413, 2014.

\bibitem{turnbull2013cloud}
L.~Turnbull and B.~Samanta, ``Cloud robotics: {F}ormation control of a multi
  robot system utilizing cloud infrastructure,'' in \emph{2013 Proceedings of
  IEEE Southeastcon}.\hskip 1em plus 0.5em minus 0.4em\relax IEEE, 2013, pp.
  1--4.

\bibitem{vick2015robot}
A.~Vick, V.~Von{\'a}sek, R.~P{\v{e}}ni{\v{c}}ka, and J.~Kr{\"u}ger, ``Robot
  control as a service-towards cloud-based motion planning and control for
  industrial robots,'' in \emph{2015 10th International Workshop on Robot
  Motion and Control (RoMoCo)}.\hskip 1em plus 0.5em minus 0.4em\relax IEEE,
  2015, pp. 33--39.

\bibitem{hong2018cloud}
C.~Hong and D.~Shi, ``A cloud-based control system architecture for
  multi-{UAV},'' in \emph{Proceedings of the 3rd International Conference on
  Robotics, Control and Automation}.\hskip 1em plus 0.5em minus 0.4em\relax
  ACM, 2018, pp. 25--30.

\bibitem{li2016toward}
Y.~Li, H.~Wang, B.~Ding, P.~Shi, and X.~Liu, ``Toward {QoS}-aware cloud robotic
  applications: A hybrid architecture and its implementation,'' in \emph{2016
  Intl IEEE Conferences on Ubiquitous Intelligence \& Computing, Advanced and
  Trusted Computing, Scalable Computing and Communications, Cloud and Big Data
  Computing, Internet of People, and Smart World Congress
  (UIC/ATC/ScalCom/CBDCom/IoP/SmartWorld)}.\hskip 1em plus 0.5em minus
  0.4em\relax IEEE, 2016, pp. 33--40.

\bibitem{DAvinCi}
R.~Arumugam, V.~R. Enti, L.~Bingbing, W.~Xiaojun, K.~Baskaran, F.~F. Kong,
  A.~S. Kumar, K.~D. Meng, and G.~W. Kit, ``Davinci: A cloud computing
  framework for service robots,'' in \emph{2010 IEEE international conference
  on robotics and automation}.\hskip 1em plus 0.5em minus 0.4em\relax IEEE,
  2010, pp. 3084--3089.

\bibitem{Cloudroid}
B.~{Hu}, H.~{Wang}, P.~{Zhang}, B.~{Ding}, and H.~{Che}, ``Cloudroid: {A} cloud
  framework for transparent and {QoS}-aware robotic computation outsourcing,''
  in \emph{2017 IEEE 10th International Conference on Cloud Computing (CLOUD)},
  2017, pp. 114--121.

\bibitem{furrer2012unr}
J.~Furrer, K.~Kamei, C.~Sharma, T.~Miyashita, and N.~Hagita, ``{UNR-PF}: An
  open-source platform for cloud networked robotic services,'' in \emph{2012
  IEEE/SICE International Symposium on System Integration (SII)}.\hskip 1em
  plus 0.5em minus 0.4em\relax IEEE, 2012, pp. 945--950.

\bibitem{miratabzadeh2016cloud}
S.~A. Miratabzadeh, N.~Gallardo, N.~Gamez, K.~Haradi, A.~R. Puthussery, P.~Rad,
  and M.~Jamshidi, ``Cloud robotics: {A} software architecture: for
  heterogeneous large-scale autonomous robots,'' in \emph{2016 World Automation
  Congress (WAC)}.\hskip 1em plus 0.5em minus 0.4em\relax IEEE, 2016, pp. 1--6.

\bibitem{chen2016hybrid}
H.~Chen, G.~Tian, F.~Lu, and G.~Liu, ``A hybrid cloud robot framework based on
  intelligent space,'' in \emph{2016 12th World Congress on Intelligent Control
  and Automation (WCICA)}.\hskip 1em plus 0.5em minus 0.4em\relax IEEE, 2016,
  pp. 2996--3001.

\bibitem{lwowski2017task}
J.~Lwowski, P.~Benavidez, J.~J. Prevost, and M.~Jamshidi, ``Task allocation
  using parallelized clustering and auctioning algorithms for heterogeneous
  robotic swarms operating on a cloud network,'' in \emph{Autonomy and
  Artificial Intelligence: A Threat or Savior?}\hskip 1em plus 0.5em minus
  0.4em\relax Springer, 2017, pp. 47--69.

\bibitem{pandey2015dynamic}
P.~Pandey, D.~Pompili, and J.~Yi, ``Dynamic collaboration between networked
  robots and clouds in resource-constrained environments,'' \emph{IEEE
  Transactions on Automation Science and Engineering}, vol.~12, no.~2, pp.
  471--480, 2015.

\bibitem{li2018latency}
S.~Li, Z.~Zheng, W.~Chen, Z.~Zheng, and J.~Wang, ``Latency-aware task
  assignment and scheduling in collaborative cloud robotic systems,'' in
  \emph{2018 IEEE 11th International Conference on Cloud Computing
  (CLOUD)}.\hskip 1em plus 0.5em minus 0.4em\relax IEEE, 2018, pp. 65--72.

\bibitem{li2017subtask}
W.~Li, C.~Zhu, L.~T. Yang, L.~Shu, E.~C.-H. Ngai, and Y.~Ma, ``Subtask
  scheduling for distributed robots in cloud manufacturing,'' \emph{IEEE
  Systems Journal}, vol.~11, no.~2, pp. 941--950, 2017.

\bibitem{xie2019loosely}
Y.~{Xie}, Y.~{Guo}, Z.~{Mi}, Y.~{Yang}, and M.~S. {Obaidat}, ``Loosely coupled
  cloud robotic framework for {QoS}-driven resource allocation-based web
  service composition,'' \emph{IEEE Systems Journal}, pp. 1--12, 2019.

\bibitem{antevski2018enhancing}
K.~Antevski, M.~Groshev, L.~Cominardi, C.~J. Bernardos, A.~Mourad, and
  R.~Gazda, ``Enhancing edge robotics through the use of context information,''
  in \emph{Proceedings of the Workshop on Experimentation and Measurements in
  5G}, ser. EM-5G'18.\hskip 1em plus 0.5em minus 0.4em\relax ACM, 2018, pp.
  7--12.

\bibitem{mahmud2018fog}
R.~Mahmud, R.~Kotagiri, and R.~Buyya, ``Fog {C}omputing: A taxonomy, survey and
  future directions,'' in \emph{Internet of Everything}.\hskip 1em plus 0.5em
  minus 0.4em\relax Springer, 2018, pp. 103--130.

\bibitem{8894519}
R.~{Mahmud}, A.~N. {Toosi}, K.~{Ramamohanarao}, and R.~{Buyya}, ``Context-aware
  placement of industry 4.0 applications in fog computing environments,''
  \emph{IEEE Transactions on Industrial Informatics}, vol.~16, no.~11, pp.
  7004--7013, 2020.

\bibitem{mahmud2018latency}
R.~Mahmud, K.~Ramamohanarao, and R.~Buyya, ``Latency-aware application module
  management for fog computing environments,'' \emph{ACM Transactions Internet
  Technology}, vol.~19, no.~1, pp. 9:1--9:21, 2018.

\bibitem{mohamed2018fog}
N.~Mohamed, J.~Al-Jaroodi, and I.~Jawhar, ``Fog-enabled multi-robot systems,''
  in \emph{2018 IEEE 2nd International Conference on Fog and Edge Computing
  (ICFEC)}.\hskip 1em plus 0.5em minus 0.4em\relax IEEE, 2018, pp. 1--10.

\bibitem{fogRobotics}
S.~L. K.~C. Gudi, S.~Ojha, B.~Johnston, J.~Clark, and M.-A. Williams, ``Fog
  robotics for efficient, fluent and robust human-robot interaction,'' in
  \emph{2018 IEEE 17th International Symposium on Network Computing and
  Applications (NCA)}.\hskip 1em plus 0.5em minus 0.4em\relax IEEE, 2018, pp.
  1--5.

\bibitem{karjee2017distributed}
J.~Karjee, S.~Behera, H.~K. Rath, and A.~Simha, ``Distributed cooperative
  communication and link prediction in cloud robotics,'' in \emph{2017 IEEE
  International Conference on Sensing, Communication and Networking (SECON
  Workshops)}.\hskip 1em plus 0.5em minus 0.4em\relax IEEE, 2017, pp. 1--7.

\bibitem{botta2019cloud}
A.~Botta, L.~Gallo, and G.~Ventre, ``Cloud, fog, and dew robotics:
  architectures for next generation applications,'' in \emph{2019 7th IEEE
  International Conference on Mobile Cloud Computing, Services, and Engineering
  (MobileCloud)}.\hskip 1em plus 0.5em minus 0.4em\relax IEEE, 2019, pp.
  16--23.

\bibitem{ichnowski2020fog}
J.~Ichnowski, W.~Lee, V.~Murta, S.~Paradis, R.~Alterovitz, J.~E. Gonzalez,
  I.~Stoica, and K.~Goldberg, ``Fog robotics algorithms for distributed motion
  planning using lambda serverless computing,'' in \emph{2020 IEEE
  International Conference on Robotics and Automation (ICRA)}.\hskip 1em plus
  0.5em minus 0.4em\relax IEEE, 2020, pp. 4232--4238.

\bibitem{wang2012game}
L.~Wang and M.~Q.-H. Meng, ``A game theoretical bandwidth allocation mechanism
  for cloud robotics,'' in \emph{Proceedings of the 10th World Congress on
  Intelligent Control and Automation}.\hskip 1em plus 0.5em minus 0.4em\relax
  IEEE, 2012, pp. 3828--3833.

\bibitem{wang2016pricing}
L.~Wang, M.~Liu, and M.~Q.-H. Meng, \emph{A pricing mechanism for task oriented
  resource allocation in cloud robotics}.\hskip 1em plus 0.5em minus
  0.4em\relax Springer International Publishing, 2016, pp. 3--31.

\bibitem{el2016environment}
M.~El~Hariri, I.~H. Elhajj, C.~Mansour, E.~Shammas, and D.~Asmar,
  ``Environment-motivated real-time bandwidth allocation for collaborative
  robots teleoperation,'' in \emph{2016 18th Mediterranean Electrotechnical
  Conference (MELECON)}.\hskip 1em plus 0.5em minus 0.4em\relax IEEE, 2016, pp.
  1--6.

\bibitem{HierarchicalAuction}
L.~Wang, M.~Liu, and M.~Q.-H. Meng, ``A hierarchical auction-based mechanism
  for real-time resource allocation in cloud robotic systems,'' \emph{IEEE
  Transactions on Cybernetics}, vol.~47, no.~2, pp. 473--484, 2016.

\bibitem{CloudResourceAllocation_Survey_2}
S.~H.~H. Madni, M.~S.~A. Latiff, Y.~Coulibaly \emph{et~al.}, ``Recent
  advancements in resource allocation techniques for cloud computing
  environment: a systematic review,'' \emph{Cluster Computing}, vol.~20, no.~3,
  pp. 2489--2533, 2017.

\bibitem{west2019debris}
C.~West, F.~Arvin, W.~Cheah, A.~West, S.~Watson, M.~Giuliani, and B.~Lennox,
  ``A debris clearance robot for extreme environments,'' in \emph{Annual
  Conference Towards Autonomous Robotic Systems}.\hskip 1em plus 0.5em minus
  0.4em\relax Springer, 2019, pp. 148--159.

\bibitem{cheah_watson_lennox_2019}
W.~C. Cheah, S.~A. Watson, and B.~Lennox, ``Limitations of wireless power
  transfer technologies for mobile robots,'' \emph{Wireless Power Transfer},
  vol.~6, no.~2, p. 175–189, 2019.

\bibitem{shinohara2011power}
N.~Shinohara, ``Power without wires,'' \emph{IEEE Microwave magazine}, vol.~12,
  no.~7, pp. S64--S73, 2011.

\bibitem{chen2011contactless}
L.~J. Chen, W.~I.~S. Tong, B.~Meyer, A.~Abdolkhani, and A.~P. Hu, ``A
  contactless charging platform for swarm robots,'' in \emph{IECON 2011-37th
  Annual Conference of the IEEE Industrial Electronics Society}.\hskip 1em plus
  0.5em minus 0.4em\relax IEEE, 2011, pp. 4088--4093.

\bibitem{samdanis2016network}
K.~Samdanis, X.~Costa-Perez, and V.~Sciancalepore, ``From network sharing to
  multi-tenancy: The 5g network slice broker,'' \emph{IEEE Communications
  Magazine}, vol.~54, no.~7, pp. 32--39, 2016.

\bibitem{kalor2018network}
A.~E. Kal{\o}r, R.~Guillaume, J.~J. Nielsen, A.~Mueller, and P.~Popovski,
  ``Network slicing in industry 4.0 applications: Abstraction methods and
  end-to-end analysis,'' \emph{IEEE Transactions on Industrial Informatics},
  vol.~14, no.~12, pp. 5419--5427, 2018.

\bibitem{chen2016smart}
F.~Chen, L.~Wang, J.~Lu, F.~Ren, Y.~Wang, X.~Zhang, and C.~Xu, ``A smart cloud
  robotic system based on cloud computing services,'' in \emph{2016 7th
  International Conference on Cloud Computing and Big Data (CCBD)}.\hskip 1em
  plus 0.5em minus 0.4em\relax IEEE, 2016, pp. 316--321.

\bibitem{kattepur2016resource}
A.~Kattepur, H.~Dohare, V.~Mushunuri, H.~K. Rath, and A.~Simha, ``Resource
  constrained offloading in fog computing,'' in \emph{Proceedings of the 1st
  Workshop on Middleware for Edge Clouds \& Cloudlets}.\hskip 1em plus 0.5em
  minus 0.4em\relax ACM, 2016, p.~1.

\bibitem{gudi2017fog}
S.~Gudi, S.~Ojha, J.~Clark, B.~Johnston, and M.-A. Williams, ``Fog robotics: An
  introduction,'' in \emph{IEEE/RSJ International Conference on Intelligent
  Robots and Systems}, 2017.

\bibitem{kattepur2017priori}
A.~Kattepur, H.~K. Rath, and A.~Simha, ``A-priori estimation of computation
  times in fog networked robotics,'' in \emph{2017 IEEE International
  Conference on Edge Computing (EDGE)}.\hskip 1em plus 0.5em minus 0.4em\relax
  IEEE, 2017, pp. 9--16.

\bibitem{cran_my}
M.~Afrin, M.~Razzaque, I.~Anjum, M.~Hassan, and A.~Alamri, ``Tradeoff between
  user quality-of-experience and service provider profit in 5g cloud radio
  access network,'' \emph{Sustainability}, vol.~9, no.~11, p. 2127, 2017.

\bibitem{Fault_tolerance}
M.~Hasan and M.~S. Goraya, ``Fault tolerance in cloud computing environment: A
  systematic survey,'' \emph{Computers in Industry}, vol.~99, pp. 156--172,
  2018.

\bibitem{li2020energy}
M.~Li, N.~Cheng, J.~Gao, Y.~Wang, L.~Zhao, and X.~Shen, ``Energy-efficient
  uav-assisted mobile edge computing: Resource allocation and trajectory
  optimization,'' \emph{IEEE Transactions on Vehicular Technology}, vol.~69,
  no.~3, pp. 3424--3438, 2020.

\bibitem{iosup2010grid}
A.~Iosup and D.~Epema, ``Grid computing workloads,'' \emph{IEEE Internet
  Computing}, vol.~15, no.~2, pp. 19--26, 2010.

\bibitem{tanwani2019fog}
A.~K. Tanwani, N.~Mor, J.~Kubiatowicz, J.~E. Gonzalez, and K.~Goldberg, ``A fog
  robotics approach to deep robot learning: Application to object recognition
  and grasp planning in surface decluttering,'' in \emph{2019 International
  Conference on Robotics and Automation (ICRA)}.\hskip 1em plus 0.5em minus
  0.4em\relax IEEE, 2019, pp. 4559--4566.

\bibitem{tanwani2020rilaas}
A.~K. Tanwani, R.~Anand, J.~E. Gonzalez, and K.~Goldberg, ``Rilaas: Robot
  inference and learning as a service,'' \emph{IEEE Robotics and Automation
  Letters}, 2020.

\bibitem{kehoe2013cloud}
B.~Kehoe, A.~Matsukawa, S.~Candido, J.~Kuffner, and K.~Goldberg, ``Cloud-based
  robot grasping with the google object recognition engine,'' in \emph{2013
  IEEE International Conference on Robotics and Automation}.\hskip 1em plus
  0.5em minus 0.4em\relax IEEE, 2013, pp. 4263--4270.

\bibitem{beksi2014point}
W.~J. Beksi and N.~Papanikolopoulos, ``Point cloud culling for robot vision
  tasks under communication constraints,'' in \emph{2014 IEEE/RSJ International
  Conference on Intelligent Robots and Systems}.\hskip 1em plus 0.5em minus
  0.4em\relax IEEE, 2014, pp. 3747--3752.

\bibitem{Survey_RobotCloud}
Z.~Du, L.~He, Y.~Chen, Y.~Xiao, P.~Gao, and T.~Wang, ``Robot cloud: Bridging
  the power of robotics and cloud computing,'' \emph{Future Generation Computer
  Systems}, vol.~74, pp. 337--348, 2017.

\bibitem{nakashima2015fourth}
K.~Nakashima, Y.~Iwashita, P.~Yoonseok, A.~Takamine, and R.~Kurazume,
  ``Fourth-person sensing for a service robot,'' in \emph{2015 IEEE
  SENSORS}.\hskip 1em plus 0.5em minus 0.4em\relax IEEE, 2015, pp. 1--4.

\bibitem{chung2009door}
W.~Chung, C.~Rhee, Y.~Shim, H.~Lee, and S.~Park, ``Door-opening control of a
  service robot using the multifingered robot hand,'' \emph{IEEE Transactions
  on Industrial Electronics}, vol.~56, no.~10, pp. 3975--3984, 2009.

\bibitem{bistry2010cloud}
H.~Bistry and J.~Zhang, ``A cloud computing approach to complex robot vision
  tasks using smart camera systems,'' in \emph{2010 IEEE/RSJ International
  Conference on Intelligent Robots and Systems}.\hskip 1em plus 0.5em minus
  0.4em\relax IEEE, 2010, pp. 3195--3200.

\bibitem{desouza2002vision}
G.~N. DeSouza and A.~C. Kak, ``Vision for mobile robot navigation: A survey,''
  \emph{IEEE Transactions on Pattern Analysis and Machine Intelligence},
  vol.~24, no.~2, pp. 237--267, 2002.

\bibitem{mozaffari2016efficient}
M.~Mozaffari, W.~Saad, M.~Bennis, and M.~Debbah, ``Efficient deployment of
  multiple unmanned aerial vehicles for optimal wireless coverage,'' \emph{IEEE
  Communications Letters}, vol.~20, no.~8, pp. 1647--1650, 2016.

\bibitem{nasir2019uav}
A.~A. Nasir, H.~D. Tuan, T.~Q. Duong, and H.~V. Poor, ``Uav-enabled
  communication using noma,'' \emph{IEEE Transactions on Communications},
  vol.~67, no.~7, pp. 5126--5138, 2019.

\bibitem{8930577}
T.~{Shafique}, H.~{Tabassum}, and E.~{Hossain}, ``End-to-end energy-efficiency
  and reliability of uav-assisted wireless data ferrying,'' \emph{IEEE
  Transactions on Communications}, vol.~68, no.~3, pp. 1822--1837, 2020.

\bibitem{fu2020joint}
S.~Fu, Y.~Tang, N.~Zhang, L.~Zhao, S.~Wu, and X.~Jian, ``Joint unmanned aerial
  vehicle (uav) deployment and power control for internet of things networks,''
  \emph{IEEE Transactions on Vehicular Technology}, vol.~69, no.~4, pp.
  4367--4378, 2020.

\bibitem{motlagh2017uav}
N.~H. Motlagh, M.~Bagaa, and T.~Taleb, ``Uav-based iot platform: A crowd
  surveillance use case,'' \emph{IEEE Communications Magazine}, vol.~55, no.~2,
  pp. 128--134, 2017.

\bibitem{sekander2018multi}
S.~Sekander, H.~Tabassum, and E.~Hossain, ``Multi-tier drone architecture for
  5g/b5g cellular networks: Challenges, trends, and prospects,'' \emph{IEEE
  Communications Magazine}, vol.~56, no.~3, pp. 96--103, 2018.

\bibitem{pillajo2015implementation}
C.~Pillajo, R.~Hincapi{\'e}, and E.~Pilatasig, ``Implementation of a network
  control system for a robotic manipulator as cloud service,'' in \emph{2015
  CHILEAN Conference on Electrical, Electronics Engineering, Information and
  Communication Technologies (CHILECON)}.\hskip 1em plus 0.5em minus
  0.4em\relax IEEE, 2015, pp. 791--795.

\bibitem{pearl1984heuristics}
J.~Pearl, ``Heuristics: Intelligent search strategies for computer problem
  solving,'' 1984.

\bibitem{NSGAII}
K.~Deb, A.~Pratap, S.~Agarwal, and T.~Meyarivan, ``A fast and elitist
  multiobjective genetic algorithm: {NSGA-II},'' \emph{IEEE Transactions on
  Evolutionary Computation}, vol.~6, no.~2, pp. 182--197, 2002.

\bibitem{wei2010game}
G.~Wei, A.~V. Vasilakos, Y.~Zheng, and N.~Xiong, ``A game-theoretic method of
  fair resource allocation for cloud computing services,'' \emph{The Journal of
  Supercomputing}, vol.~54, no.~2, pp. 252--269, 2010.

\bibitem{jain2020ecc}
S.~Jain, C.~Nandhini, and R.~Doriya, ``{ECC}-based authentication scheme for
  cloud-based robots,'' \emph{Wireless Personal Communications}, pp. 1--20,
  2020.

\bibitem{noor2015cloudarmor}
T.~H. Noor, Q.~Z. Sheng, L.~Yao, S.~Dustdar, and A.~H. Ngu, ``Cloudarmor:
  Supporting reputation-based trust management for cloud services,'' \emph{IEEE
  Transactions on Parallel and Distributed Systems}, vol.~27, no.~2, pp.
  367--380, 2015.

\bibitem{zhu2018trust}
C.~Zhu, J.~J. Rodrigues, V.~C. Leung, L.~Shu, and L.~T. Yang, ``Trust-based
  communication for the industrial internet of things,'' \emph{IEEE
  Communications Magazine}, vol.~56, no.~2, pp. 16--22, 2018.

\bibitem{kubiatowicz2018secure}
J.~Kubiatowicz, K.~Lutz, K.~Goldberg, A.~Joseph, and J.~Gonzalaz, ``Secure fog
  robotics using the global data plane,'' \emph{NSF Proposal}, 2018.

\bibitem{Blockchain_FRUCT}
I.~Afanasyev, A.~Kolotov, R.~Rezin, K.~Danilov, A.~Kashevnik, and V.~Jotsov,
  ``Blockchain solutions for multi-agent robotic systems: Related work and open
  questions,'' in \emph{Proceedings of the 24th Conference of Open Innovations
  Association FRUCT}, ser. FRUCT'24.\hskip 1em plus 0.5em minus 0.4em\relax
  FRUCT Oy, 2019.

\bibitem{zheng2018blockchain}
Z.~Zheng, S.~Xie, H.-N. Dai, X.~Chen, and H.~Wang, ``Blockchain challenges and
  opportunities: A survey,'' \emph{International Journal of Web and Grid
  Services}, vol.~14, no.~4, pp. 352--375, 2018.

\bibitem{yang2019integrated}
R.~Yang, F.~R. Yu, P.~Si, Z.~Yang, and Y.~Zhang, ``Integrated blockchain and
  edge computing systems: A survey, some research issues and challenges,''
  \emph{IEEE Communications Surveys \& Tutorials}, vol.~21, no.~2, pp.
  1508--1532, 2019.

\bibitem{lopes2018overview}
V.~Lopes and L.~A. Alexandre, ``An overview of blockchain integration with
  robotics and artificial intelligence,'' \emph{arXiv preprint
  arXiv:1810.00329}, 2018.

\bibitem{khan2019blockchain}
A.~T. Khan, X.~Cao, S.~Li, and Z.~Milosevic, ``Blockchain technology with
  applications to distributed control and cooperative robotics: A survey,''
  \emph{International Journal of Robotics and Control}, vol.~2, no.~1, pp.
  36--48, 2019.

\bibitem{ROS}
M.~Quigley, K.~Conley, B.~Gerkey, J.~Faust, T.~Foote, J.~Leibs, R.~Wheeler, and
  A.~Y. Ng, ``{ROS}: an open-source robot operating system,'' in \emph{ICRA
  workshop on open source software}, vol.~3, no. 3.2.\hskip 1em plus 0.5em
  minus 0.4em\relax Kobe, Japan, 2009, p.~5.

\bibitem{crick2017rosbridge}
C.~Crick, G.~Jay, S.~Osentoski, B.~Pitzer, and O.~C. Jenkins,
  \emph{{ROS}bridge: {ROS} for non-{ROS} users}.\hskip 1em plus 0.5em minus
  0.4em\relax Springer International Publishing, 2017, pp. 493--504.

\bibitem{bozcuouglu2018exchange}
A.~K. Bozcuo{\u{g}}lu, G.~Kazhoyan, Y.~Furuta, S.~Stelter, M.~Beetz, K.~Okada,
  and M.~Inaba, ``The exchange of knowledge using cloud robotics,'' \emph{IEEE
  Robotics and Automation Letters}, vol.~3, no.~2, pp. 1072--1079, 2018.

\bibitem{benavidez2015cloud}
P.~Benavidez, M.~Muppidi, P.~Rad, J.~J. Prevost, M.~Jamshidi, and L.~Brown,
  ``Cloud-based realtime robotic visual {SLAM},'' in \emph{2015 Annual IEEE
  Systems Conference (SysCon) Proceedings}.\hskip 1em plus 0.5em minus
  0.4em\relax IEEE, 2015, pp. 773--777.

\bibitem{demarinis2019scanning}
N.~DeMarinis, S.~Tellex, V.~P. Kemerlis, G.~Konidaris, and R.~Fonseca,
  ``Scanning the internet for ros: A view of security in robotics research,''
  in \emph{2019 International Conference on Robotics and Automation
  (ICRA)}.\hskip 1em plus 0.5em minus 0.4em\relax IEEE, 2019, pp. 8514--8521.

\bibitem{tian2019fog}
N.~Tian, A.~K. Tanwani, J.~Chen, M.~Ma, R.~Zhang, B.~Huang, K.~Goldberg, and
  S.~Sojoudi, ``A fog robotic system for dynamic visual servoing,'' in
  \emph{2019 International Conference on Robotics and Automation (ICRA)}.\hskip
  1em plus 0.5em minus 0.4em\relax IEEE, 2019, pp. 1982--1988.

\bibitem{gudi2018fog}
S.~L. K.~C. Gudi, S.~Ojha, B.~Johnston, J.~Clark, and M.-A. Williams, ``Fog
  robotics for efficient, fluent and robust human-robot interaction,'' in
  \emph{2018 IEEE 17th International Symposium on Network Computing and
  Applications (NCA)}.\hskip 1em plus 0.5em minus 0.4em\relax IEEE, 2018, pp.
  1--5.

\bibitem{katz2014mobile}
M.~Katz, F.~H. Fitzek, D.~E. Lucani, and P.~Seeling, ``Mobile clouds as the
  building blocks of shareconomy: Sharing resources locally and widely,''
  \emph{IEEE Vehicular Technology Magazine}, vol.~9, no.~3, pp. 63--71, 2014.

\bibitem{cuervo2010maui}
E.~Cuervo, A.~Balasubramanian, D.~Cho, A.~Wolman, S.~Saroiu, R.~Chandra, and
  P.~Bahl, ``Maui: making smartphones last longer with code offload,'' in
  \emph{Proceedings of the 8th International Conference on Mobile Systems,
  Applications, and Services}.\hskip 1em plus 0.5em minus 0.4em\relax ACM,
  2010, pp. 49--62.

\bibitem{xu2016multi}
D.~Xu and X.~Peng, ``Multi-level offloading decision on efficient object
  tracking for humanoid robot,'' \emph{Advanced Robotics}, vol.~30, no.~15, pp.
  979--991, 2016.

\bibitem{berenson_robot-path-planning}
D.~{Berenson}, P.~{Abbeel}, and K.~{Goldberg}, ``A robot path planning
  framework that learns from experience,'' in \emph{2012 IEEE International
  Conference on Robotics and Automation}, 2012, pp. 3671--3678.

\bibitem{salmeron2015tradeoff}
J.~Salmer{\'o}n-Garc{\i}, P.~Inigo-Blasco, F.~D{\i}, D.~Cagigas-Mu{\~n}iz
  \emph{et~al.}, ``A tradeoff analysis of a cloud-based robot navigation
  assistant using stereo image processing,'' \emph{IEEE Transactions on
  Automation Science and Engineering}, vol.~12, no.~2, pp. 444--454, 2015.

\bibitem{guo2018energy}
Y.~Guo, Z.~Mi, Y.~Yang, and M.~S. Obaidat, ``An energy sensitive computation
  offloading strategy in cloud robotic network based on {GA},'' \emph{IEEE
  Systems Journal}, no.~99, pp. 1--11, 2018.

\bibitem{Data_Retrieval}
L.~Wang, M.~Liu, and M.~Q.-H. Meng, ``Real-time multisensor data retrieval for
  cloud robotic systems,'' \emph{IEEE Transactions on Automation Science and
  Engineering}, vol.~12, no.~2, pp. 507--518, 2015.

\bibitem{hong2019qos}
Z.~{Hong}, H.~{Huang}, S.~{Guo}, W.~{Chen}, and Z.~{Zheng}, ``{QoS}-aware
  cooperative computation offloading for robot swarms in cloud robotics,''
  \emph{IEEE Transactions on Vehicular Technology}, vol.~68, no.~4, pp.
  4027--4041, 2019.

\bibitem{Ak_TII}
A.~Rahman, J.~Jin, A.~Rahman, T.~Cricenti, and A.~Kulkarni,
  ``Communication-aware cloud robotic task offloading with on-demand mobility
  for smart factory maintenance,'' \emph{IEEE Transactions on Industrial
  Informatics}, vol.~15, no.~5, pp. 2500--2511, 2019.

\bibitem{chinchali2019network}
S.~Chinchali, A.~Sharma, J.~Harrison, A.~Elhafsi, D.~Kang, E.~Pergament,
  E.~Cidon, S.~Katti, and M.~Pavone, ``Network offloading policies for cloud
  robotics: a learning-based approach,'' \emph{arXiv preprint
  arXiv:1902.05703}, 2019.

\bibitem{li2019partial}
B.~Li, Z.~Mi, Y.~Guo, Y.~Yang, and M.~S. Obaidat, ``Partial computing
  offloading assisted cloud point registration in multi-robot slam,''
  \emph{arXiv preprint arXiv:1905.12973}, 2019.

\bibitem{osunmakinde2014development}
I.~Osunmakinde and R.~Vikash, ``Development of a survivable cloud multi-robot
  framework for heterogeneous environments,'' \emph{International Journal of
  Advanced Robotic Systems}, vol.~11, no.~10, p. 164, 2014.

\bibitem{bogue2017cloud}
R.~Bogue, ``Cloud robotics: a review of technologies, developments and
  applications,'' \emph{Industrial Robot: An International Journal}, vol.~44,
  no.~1, pp. 1--5, 2017.

\bibitem{song2017scheduling}
K.-T. Song, Y.-H. Chiu, S.-H. Song, and K.~Zinchenko, ``Scheduling and control
  of a cloud robot for reception and guidance,'' in \emph{2017 International
  Automatic Control Conference (CACS)}.\hskip 1em plus 0.5em minus 0.4em\relax
  IEEE, 2017, pp. 1--6.

\bibitem{li2017multi}
X.~Li, Z.~Liu, and F.~Tan, ``Multi-robot task allocation based on cloud ant
  colony algorithm,'' in \emph{International Conference on Neural Information
  Processing}.\hskip 1em plus 0.5em minus 0.4em\relax Springer, 2017, pp.
  3--10.

\bibitem{lomenie2004generic}
N.~Lom{\'e}nie, ``A generic methodology for partitioning unorganised 3d point
  clouds for robotic vision,'' in \emph{First Canadian Conference on Computer
  and Robot Vision, 2004. Proceedings.}\hskip 1em plus 0.5em minus 0.4em\relax
  IEEE, 2004, pp. 64--71.

\bibitem{berenson2012robot}
D.~Berenson, P.~Abbeel, and K.~Goldberg, ``A robot path planning framework that
  learns from experience,'' in \emph{2012 IEEE International Conference on
  Robotics and Automation}.\hskip 1em plus 0.5em minus 0.4em\relax IEEE, 2012,
  pp. 3671--3678.

\end{thebibliography}
%
\vskip -1\baselineskip plus -1.1fil
\begin{IEEEbiography}[\vspace{-8mm}
{\includegraphics[width=1in,height=1.25in,clip,keepaspectratio]{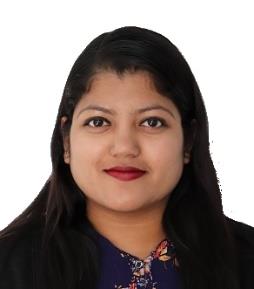}}]{Mahbuba Afrin} is currently pursuing her PhD degree at Swinburne University of Technology, Melbourne, Australia in collaboration with Commonwealth Scientific and Industrial Research Organisation (CSIRO). She received her M.Sc. (2017) and B.Sc. (2015) in Computer Science and Engineering from University of Dhaka, Bangladesh. Before starting PhD programme, she worked as a Lecturer at Department of Computer Science and Engineering in United International University, Bangladesh (2015-2017) and as a Research Assistant in Green Networking Research Group, University of Dhaka (2013-2015). Her research interests include Cloud robotics, multi-robot systems, edge computing, cloud computing, cloud radio access network, mobile cloud computing and cyber-physical systems. She is a Student Member of IEEE.
\end{IEEEbiography}
\vskip -1.5\baselineskip plus -1.3fil
\begin{IEEEbiography}[\vspace{-8mm}
{\includegraphics[width=1in,height=1.25in,clip,keepaspectratio]{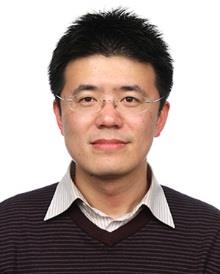}}]{Jiong Jin} (IEEE M’11) received the B.E. degree with First Class Honours in Computer Engineering from Nanyang Technological University, Singapore, in 2006, and the Ph.D. degree in Electrical and Electronic Engineering from the University of Melbourne, Australia, in 2011. From 2011 to 2013, he was a Research Fellow in the Department of Electrical and Electronic Engineering at the University of Melbourne. He is currently a Senior Lecturer in the School of Software and Electrical Engineering, Faculty of Science, Engineering and Technology, Swinburne University of Technology, Melbourne, Australia. His research interests include network design and optimization, edge computing and distributed systems, robotics and automation, and cyber-physical systems and Internet of Things as well as their applications in smart manufacturing, smart transportation, and smart cities.
\end{IEEEbiography}
\vskip -1.5\baselineskip plus -1fil
\begin{IEEEbiography}[\vspace{-8mm}
{\includegraphics[width=1in,height=1.25in,clip,keepaspectratio]{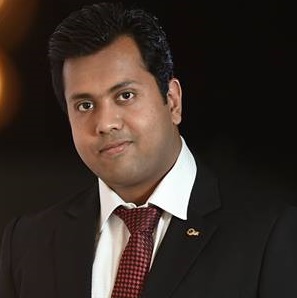}}]{Akhlaqur Rahman} received his PhD degree in Electrical and Electronic Engineering from Swinburne University of Technology, Melbourne, Australia in 2019. He received his B.Sc in Electrical and Electronic Engineering from American International University, Bangladesh in 2012. Before joining Swinburne, Akhlaqur was a Lecturer and Coordinator for the department of Electrical and Electronic Engineering (from 2013 to 2014) at Uttara University in Bangladesh. Akhlaqur is currently working as Lecturer at Engineering Institute of Technology (EIT), Australia. Akhlaqur's research focus is on cloud networked robotics, industrial automation, network optimization, smart manufacturing, and Internet of Things (IoT). He is a Member of IEEE.
\end{IEEEbiography}
\vskip -1.5\baselineskip plus -1fil
\begin{IEEEbiography}[\vspace{-8mm}
{\includegraphics[width=1in,height=1.25in,clip,keepaspectratio]{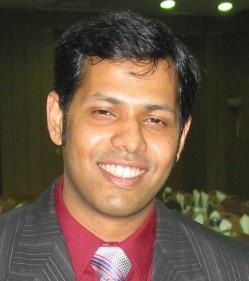}}]{Ashfaqur Rahman} is a Principal Research Scientist and Team Leader at Data61 division of CSIRO, Australia. He is a data scientist and working as a researcher for nearly 15 years. His key research areas are machine learning and Data mining. More specifically Ensemble learning and fusion, Evolutionary Algorithm based network optimization, Distributed Machine Learning for Big Data, Feature selection/weighting methods, Image segmentation and classification. Dr. Rahman received his Ph.D. degree in Information Technology from Monash University, Australia in 2008. He has published over 100 peer-reviewed journal articles, book chapters and conference papers. He supervises several PhD students in collaboration with Swinburne University of Technology and Federation University. He served as reviewer of prestigious journals and conferences. He is the associate editor of Elsevier journal 'Information Processing in Agriculture'. He is involved in organization of key events including IEEE DICTA [2018, 2013], MSLDA [2018, 2017, 2016, 2014]. He is a Senior Member of IEEE.
\end{IEEEbiography}
\vskip -1.5\baselineskip plus -1fil
\begin{IEEEbiography}[\vspace{-8mm}
{\includegraphics[width=1in,height=1.25in,clip,keepaspectratio]{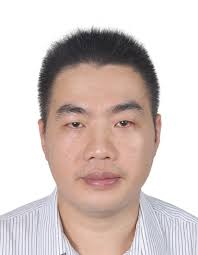}}]{Jiafu Wan} (M’11) has been a Professor with the School of Mechanical and Automotive Engineering, South China University of Technology (SCUT), since 2015. His research interests include cyber-physical systems, Industry 4.0, smart factory, Industrial Big Data, Industrial Robot, and the Internet of Vehicles. He has directed 20 research projects, including the National Key Research and Development Project, the National Natural Science Foundation of China, and the Joint Fund of the National Natural Science Foundation of China and Guangdong Province, the High-level Talent Project of Guangdong Province, and the Natural Science Foundation of Guangdong Province. So far, he has published more than 140 scientific papers, including over 100 SCI-indexed papers, over 40 IEEE Transactions/Journal papers, 19 ESI Highly Cited Papers, and four ESI Hot Papers. His research results have been published in some famous IEEE Journals and Magazines, such as the IEEE Transactions on Industrial Informatics, the IEEE/ASME Transactions on Mechatronics, the IEEE Communications Surveys and Tutorials, the IEEE Communications Magazine, the IEEE Transactions on Intelligent Transportation Systems, the IEEE Network, the IEEE Wireless Communications, the IEEE Systems Journal, the IEEE Sensors Journal, and the IEEE Internet of Things Journal. According to Google Scholar, his published work has been cited more than 7400 times (H-index = 43). His other SCI citations, the sum of times cited without self-citations reached 2000 times (H-index = 28), according to the Web of Science Core Collection. He is a Member of IEEE and Senior Member of the CMES and CCF. He is the General Chair for the IndustrialIoT 2016 and CloudComp 2016. He is an Associate Editor of the IEEE/ASME Transactions on Mechatronics, the Journal of Intelligent Manufacturing, the IEEE Access, the Computers and Electrical Engineering, and PLOS ONE. He is a Leading Guest Editor of several SCI-indexed journals, such as the IEEE Systems Journal, the IEEE Access, the Elsevier Computer Networks, Mobile Networks and Applications, the Computers and Electrical Engineering, and the Microprocessors and Microsystems.
\end{IEEEbiography}
\vskip -1.5\baselineskip plus -1fil
\begin{IEEEbiography}[{\includegraphics[width=1in,height=1.25in,clip,keepaspectratio]{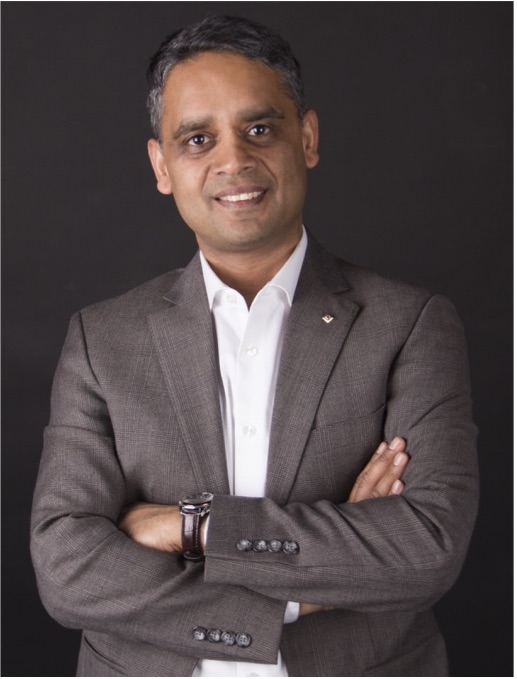}}]{Ekram Hossain} is a Professor and the Associate Head (Graduate Studies) in the Department of Electrical and Computer Engineering at University of Manitoba, Canada. He is a Member (Class of 2016) of the College of the Royal Society of Canada. Also, he is a Fellow of the Canadian Academy of Engineering. Dr. Hossain's current research interests include design, analysis, and optimization beyond 5G cellular wireless networks. He was elevated to an IEEE Fellow "for contributions to spectrum management and resource allocation in cognitive and cellular radio networks". He received the 2017 IEEE ComSoc TCGCC (Technical Committee on Green Communications \& Computing) Distinguished Technical Achievement Recognition Award "for outstanding technical leadership and achievement in green wireless communications and networking". Dr. Hossain has won several research awards including the 2017 IEEE Communications Society Best Survey Paper Award and the 2011 IEEE Communications Society Fred Ellersick Prize Paper Award. He was listed as a Clarivate Analytics Highly Cited Researcher in Computer Science in 2017, 2018, 2019, and 2020. Currently he serves as the Editor-in-Chief of IEEE Press and an Editor for the IEEE Transactions on Mobile Computing. Previously, he served as the Editor-in-Chief for the IEEE Communications Surveys and Tutorials (2012-2016). He is a Distinguished Lecturer of the IEEE Communications Society and the IEEE Vehicular Technology Society. He serves as the Director of Magazines for the IEEE Communications Society (2020-2021). Also, he is an elected member of the Board of Governors of the IEEE Communications Society for the term 2018-2020. 
\end{IEEEbiography}
\end{document}